\documentclass[10pt,journal,compsoc]{IEEEtran}

\usepackage[OT1]{fontenc} 
\usepackage[utf8]{inputenc}
\usepackage{amsmath,amsfonts,amssymb,amsthm}
\usepackage{graphicx}
\usepackage{color}
\usepackage{caption}
\usepackage{subcaption}
\usepackage{algorithm}
\usepackage[noend]{algpseudocode}
\usepackage{xspace}
\usepackage{ifthen}
\usepackage[nocompress]{cite}
\usepackage{array}
\usepackage{multirow}
\usepackage[pagebackref=true,breaklinks=true,letterpaper=true,colorlinks,bookmarks=false]{hyperref}
\usepackage{pifont}%
\newcommand{\cmark}{\ding{51}}%
\newcommand{\xmark}{\ding{55}}%
\usepackage{soul}

\newcommand\labelLastPage{%
  \edef\@currentlabel{\thepage}%
  \label{LastPage}%
}

\hypersetup{
	colorlinks = true,
	linkcolor = blue,
	anchorcolor = blue,
	citecolor = blue,
	filecolor = blue,
	urlcolor = blue
}

\graphicspath{ {gfx/} }

\newcommand{\bs}{\mathbf{s}}

\newcommand{\bp}{\mathbf{p}}

\newcommand{\bc}{\mathbf{c}}

\newcommand{\bpi}{\boldsymbol{\pi}}

\newcommand{\cB}{\mathcal{B}}
\newcommand{\cL}{\mathcal{L}}

\newcommand{\cP}{\mathcal{P}}

\newcommand{\cV}{\mathcal{V}}

\newcommand{\figref}[1]{\Fig~\ref{#1}}
\newcommand{\secref}[1]{Section~\ref{#1}}
\newcommand{\appref}[1]{Appendix~\ref*{#1}}

\renewcommand{\eqref}[1]{Eq.~\ref{#1}}
\newcommand{\tabref}[1]{Table~\ref{#1}}

\makeatletter
\DeclareRobustCommand\onedot{\futurelet\@let@token\@onedot}
\def\@onedot{\ifx\@let@token.\else.\null\fi\xspace}
\def\eg{e.g\onedot} 
\def\ie{i.e\onedot}

\def\wrt{wrt\onedot}

\def\etal{et~al\onedot} 
\def\Fig{Fig\onedot}   
\makeatother

\newcommand*\rot{\rotatebox{90}}
\newcommand{\boldparagraph}[1]{\vspace{0.2cm}\noindent{\bf #1:} }

\newcolumntype{P}[1]{>{\centering\arraybackslash}m{#1}}

\newcommand{\del}[1]{}

\newif\ifcomment
\commenttrue
\ifcomment
	\newcommand{\ag}[1]{ \noindent {\color{red} {\bf Andreas:} {#1}} }
	\newcommand{\jx}[1]{ \noindent {\color{blue} {\bf Jun:} {#1}} }
	\newcommand{\kc}[1]{ \noindent {\color{green} {\bf Kashyap:} {#1}} }
	\newcommand{\yl}[1]{ \noindent {\color{cyan} {\bf Yiyi:} {#1}} }
\else
	\newcommand{\ag}[1]{}
	\newcommand{\jx}[1]{}
	\newcommand{\kc}[1]{}
	\newcommand{\yl}[1]{}
\fi

\begin{document}
\title{KITTI-360: A Novel Dataset and Benchmarks \\ for Urban Scene Understanding in 2D and 3D}
\author{Yiyi~Liao \qquad Jun~Xie \qquad Andreas~Geiger%
\IEEEcompsocitemizethanks{\IEEEcompsocthanksitem Y. Liao is with the Autonomous Vision Group, University of T{\"u}bingen and Max Planck Institute for Intelligent Systems, T{\"u}bingen, Germany, and Zhejiang University. J. Xie is with Google Research.  A. Geiger is with the Autonomous Vision Group, University of T{\"u}bingen and Max Planck Institute for Intelligent Systems, T{\"u}bingen, Germany. E-mails: yiyi.liao@tue.mpg.de, junx@google.com, a.geiger@uni-tuebingen.de.}
}

\IEEEtitleabstractindextext{%
\parbox{0.918\textwidth}{
\begin{abstract}
For the last few decades, several major subfields of artificial intelligence including computer vision, graphics, and robotics have progressed largely independently from each other. Recently, however, the community has realized that progress towards robust intelligent systems such as self-driving cars requires a concerted effort across the different fields. This motivated us to develop KITTI-360, successor of the popular KITTI dataset. KITTI-360 is a suburban driving dataset which comprises richer input modalities, comprehensive semantic instance annotations and accurate localization to facilitate research at the intersection of vision, graphics and robotics. For efficient annotation, we created a tool to label 3D scenes with bounding primitives and developed a model that transfers this information into the 2D image domain, resulting in over  150k images and 1B 3D points with coherent semantic instance annotations across 2D and 3D. Moreover, we established benchmarks and baselines for several tasks relevant to mobile perception, encompassing problems from computer vision, graphics, and robotics on the same dataset, \eg, semantic scene understanding, novel view synthesis and semantic SLAM. KITTI-360 will enable progress at the intersection of these research areas and thus contribute towards solving one of today's grand challenges: the development of fully autonomous self-driving systems.
\end{abstract}
}

\begin{IEEEkeywords}
Point Cloud Labeling, Semantic Label Transfer, Scene Understanding, Self-Driving, Datasets, Performance Evaluation
\end{IEEEkeywords}}

\maketitle
\IEEEdisplaynontitleabstractindextext
\IEEEpeerreviewmaketitle

\IEEEraisesectionheading{\section{Introduction}\label{sec:introduction}}

\IEEEPARstart{O}{ne} of the pioneering works in \textit{computer vision} can be traced back to Larry Roberts' ``Blocks World'' in the 1960s~\cite{Roberts1963}, which aimed at identifying individual objects and inferring the 3D structure of simple shapes from 2D images. With the goal of understanding a scene from visual cues, computer vision was viewed as a comparably easy first step towards solving higher-level reasoning tasks in \textit{robotics} at that time (\eg, the MIT copy demo \cite{Winston1971}). Albeit being seemingly easy for humans, robustly perceiving geometry and semantics from images proved hard for machines due to the high complexity of real-world environments.
Thus, in the 1980s, computer vision and robotics evolved into their own, largely independent research fields. Only recently, the communities have realized that it is impossible to solve one without the other, e.g., in the context of self-driving~\cite{Janai2020FTCGV}. Similarly, computer vision's interaction with \textit{computer graphics} emerged in the 1990s~\cite{Seitz1999} and has gained traction over the last decade, in particular in areas such as neural and image-based rendering~\cite{Mildenhall2020ECCV}. These advances can in turn benefit robotics as simulation will be crucial for training and validating the next generation of robotic systems.

\begin{figure*}[t]

\def\mywidth{0.25}
\setlength\tabcolsep{0.0em}
\begin{tabular}{cccc}
	\multicolumn{4}{c}{\includegraphics[width=\linewidth]{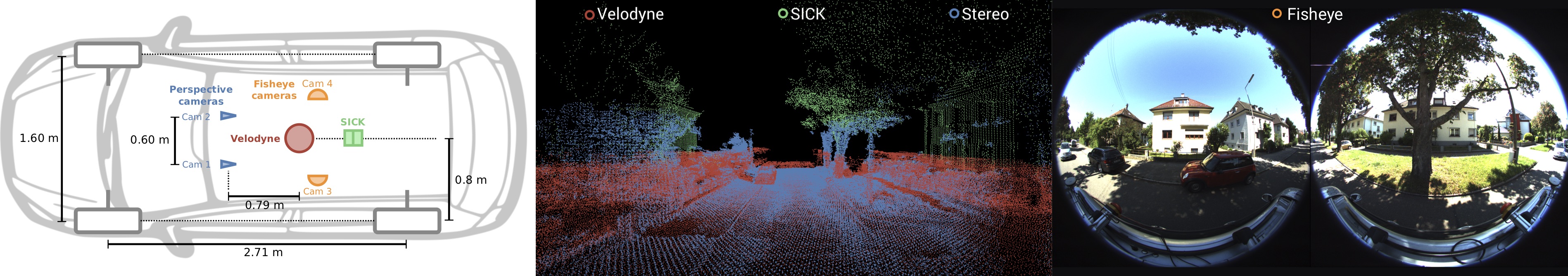}}\\
	\includegraphics[width=\mywidth\linewidth]{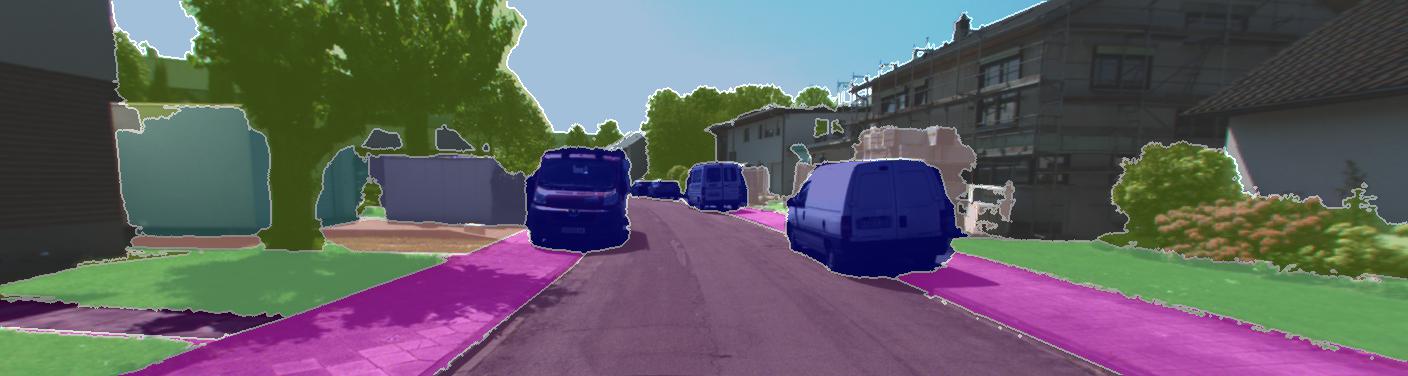}&
	\includegraphics[width=\mywidth\linewidth]{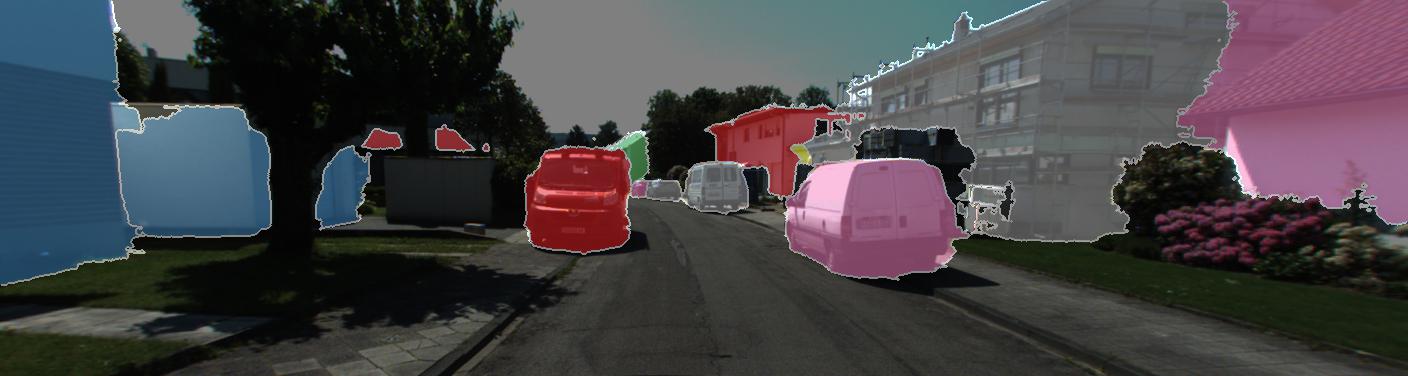}&
	\includegraphics[width=\mywidth\linewidth]{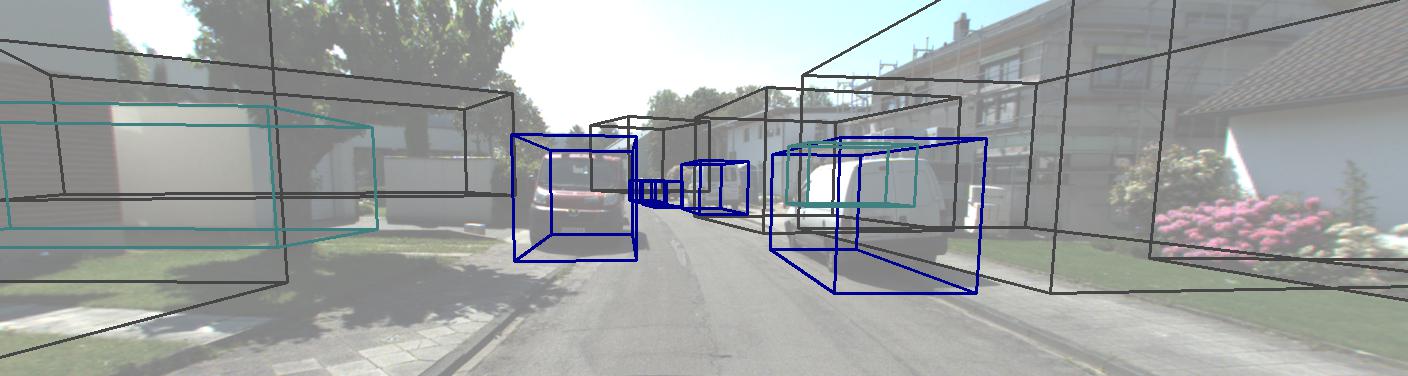}&
	\includegraphics[width=\mywidth\linewidth]{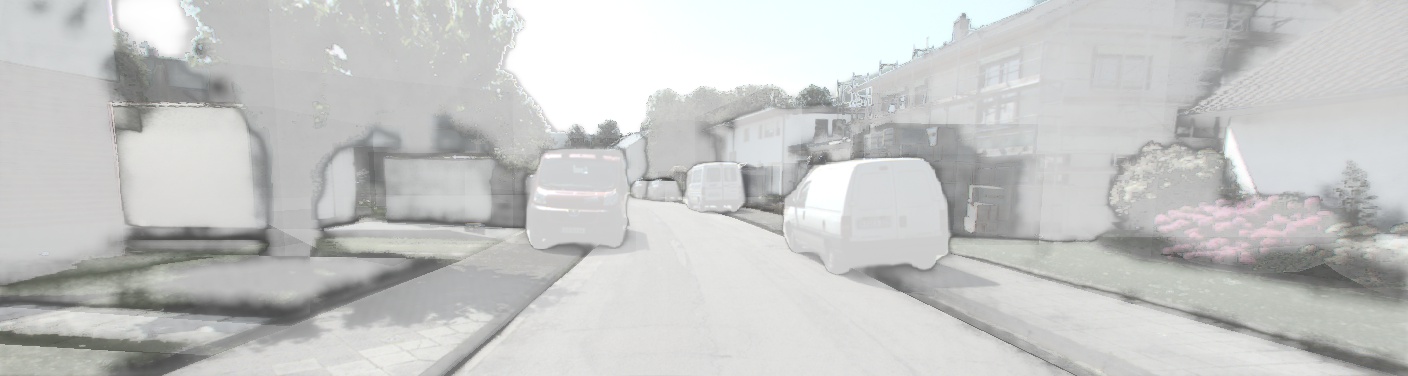}\\
	\includegraphics[width=\mywidth\linewidth]{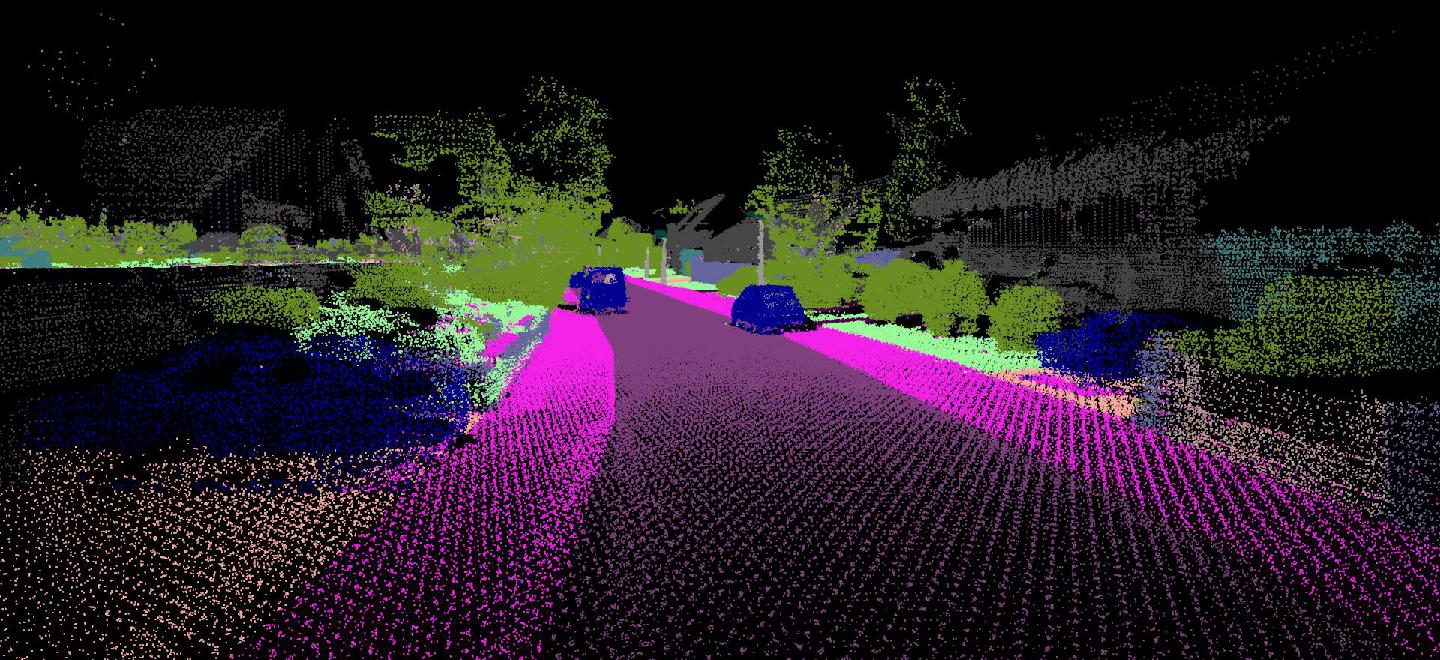}&
	\includegraphics[width=\mywidth\linewidth]{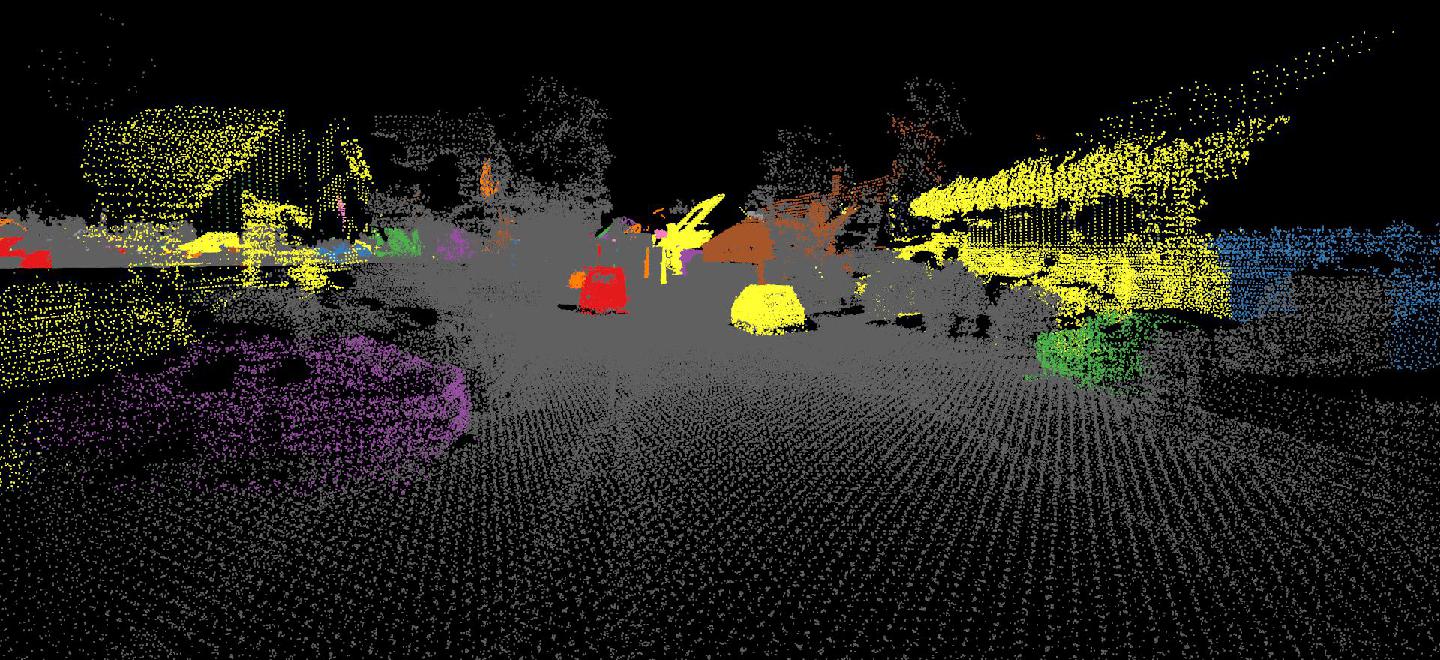}&
	\includegraphics[width=\mywidth\linewidth]{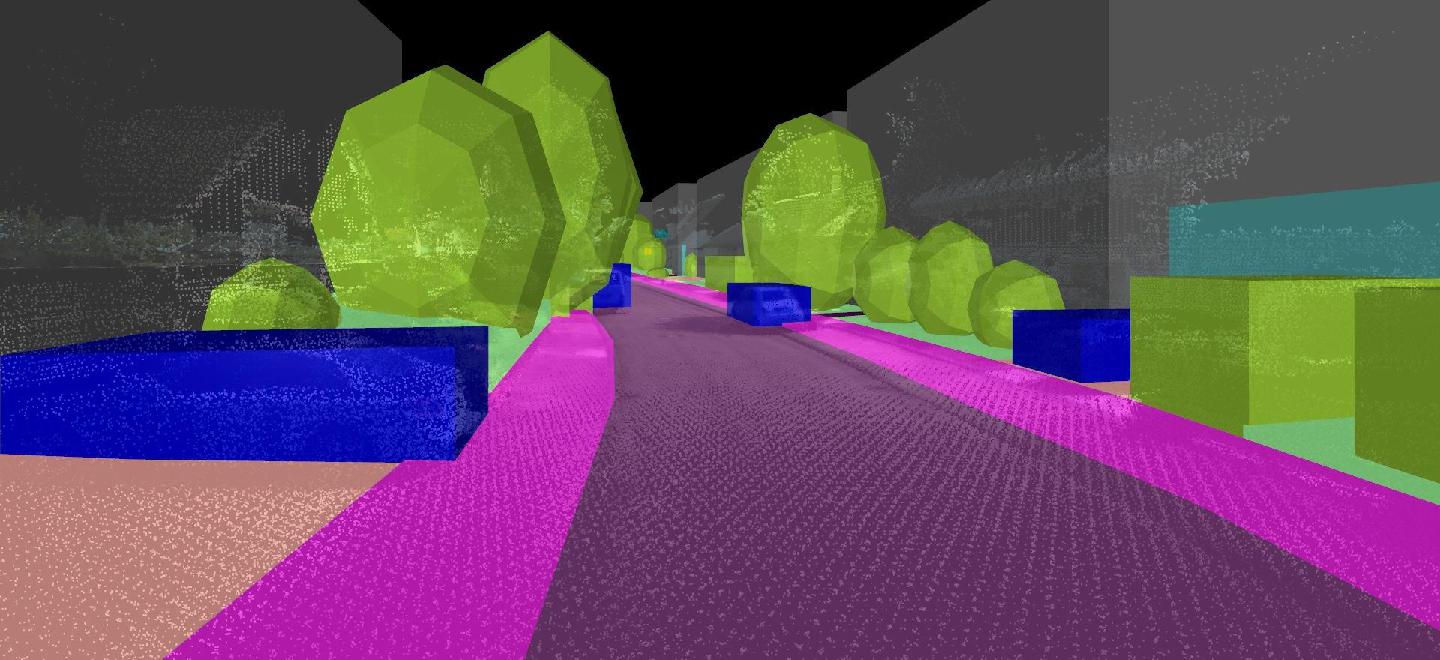}&
	\includegraphics[width=\mywidth\linewidth]{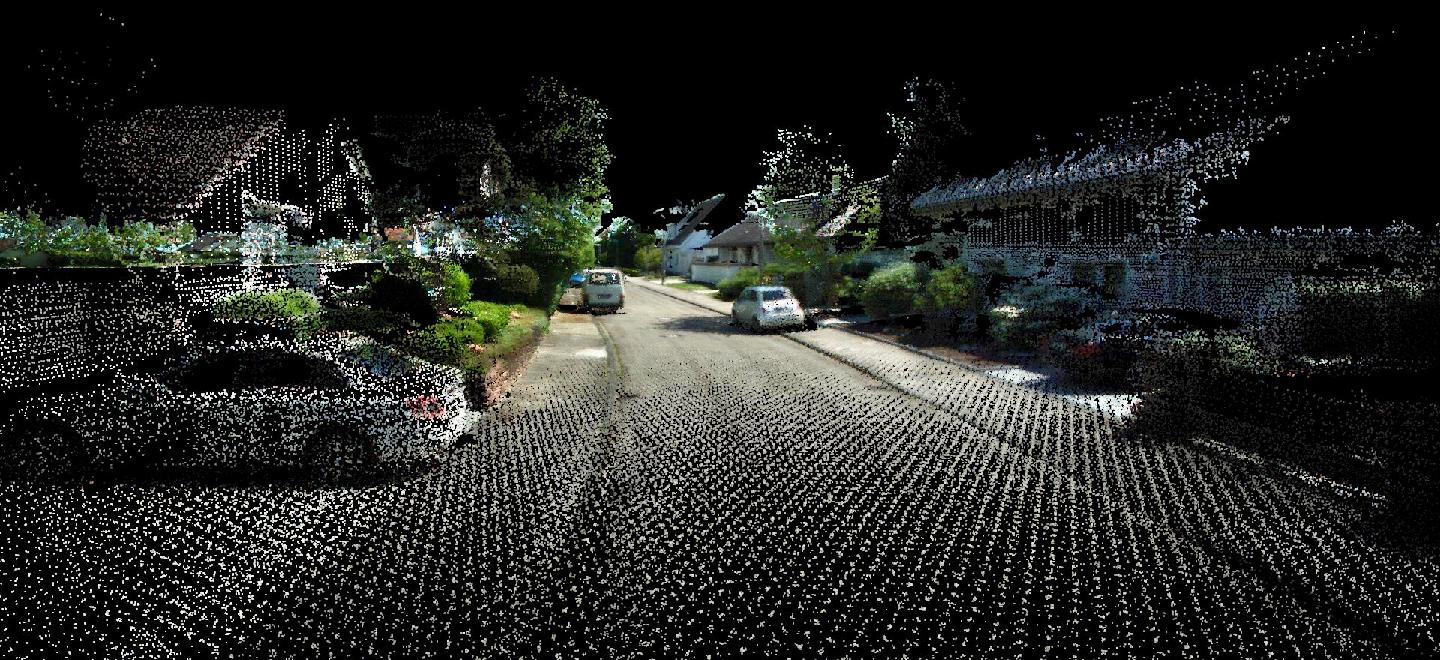}\\
	Semantic & Instance & Bounding Box & Confidence \& RGB \\
\end{tabular}
\vspace{-0.2cm}
\caption{{\bf KITTI-360.} Our dataset contains rich sensor modalities, including a perspective stereo camera, a pair of fisheye cameras, a Velodyne and a SICK laser scanning unit which together enable 360$^\circ$ scene perception. We release comprehensive annotations including consistent semantic and instance labels for every 2D image pixel and 3D point.}
\label{fig:kitti360}
\end{figure*}

The converging trend of vision, graphics, and robotics motivates us to create a new dataset, KITTI-360, that addresses tasks at the intersection of these fields with a focus on autonomous driving.
While the KITTI dataset~\cite{Geiger2012CVPR} has pushed the state-of-the-art in computer vision algorithms forward, it does not contain dense and complete semantic labels. 
Thus, many interesting interdisciplinary tasks, e.g., synthesizing novel view images jointly with semantics or reconstructing large-scale semantic maps,
cannot be evaluated on KITTI. 
Moreover, the captured perspective front images provide only a partial view of the scene and the 3D information provided by the LiDAR sensor is very sparse. The GPS localization of KITTI is reliable but does not reach sub-pixel accuracy when fusing multiple frames.
With KITTI-360 we address these shortcomings by providing a new dataset with more comprehensive semantic/instance labels in 2D and 3D, richer 360$^\circ$ sensory information (fisheye images and pushbroom laser scans), very accurate and geo-localized vehicle and camera poses, and a series of new challenging benchmarks, see Fig.~\ref{fig:kitti360} for an overview.

A key challenge towards building such a dataset is to obtain coherent dense and comprehensive semantics in 2D and 3D.
Many existing datasets are annotated in the 2D image domain where pixel-wise labeling requires up to $60$ minutes per image for a human annotator\cite{Badrinarayanan2010CVPR}. Other datasets~\cite{Zolanvari2019BMVC,Behley2019ICCV,Tan2020CVPRW} are annotated in 3D space while ignoring information in the 2D image domain. 
A few datasets~\cite{Caesar2020CVPR} offer labels in both 2D and 3D. However, annotation is conducted independently, thus duplicating the labeling effort. 

In this paper, we propose an alternative approach that leverages coarse 3D annotations to significantly simplify the dense annotation task and establish coherent labels in both 2D and 3D space. Moreover, this yields a unique \textit{instance} index for each object in the scene across all 2D video frames. 
Specifically, we build a WebGL-based annotation tool that allows for annotating both \textit{static} and \textit{dynamic} scene elements directly in 3D using simple primitives. 
This approach has several advantages over labeling in 2D: First, objects often project into several video frames, 
thus lowering annotation efforts considerably. Further, the obtained 2D instance annotations are temporally coherent as they are associated with a single physical 3D object. Finally, our 3D annotations covering the full 3D scene are useful on their own, \eg,  for reasoning in 3D \cite{Zhang2013ICCV,Geiger2015GCPR} or to enrich 2D annotations with approximate 3D geometry.

However, obtaining dense and accurate pixel-wise 2D labels and point-wise 3D labels from sparse, noisy point clouds and coarse 3D annotations is a challenging task. %
Towards solving this problem, we propose a non-local multi-field CRF model which reasons jointly about semantic and instance labels of all 3D points and 2D pixels. 
Our approach also leverages learning-based methods to provide dense semantic and instance priors in the 2D image domain. As evidenced by our experiments, our method outperforms label propagation methods operating purely in 2D as well as pure learning-based approaches.
Furthermore, the probabilistic nature of our model allows for estimating label uncertainties which can be used to increase label accuracy when only a subset of the pixels require a label.

From the annotated dataset, we derive several benchmarks and baselines with novel and challenging tasks %
at the intersection of vision, graphics and robotics which we believe are crucial for making progress towards the grand challenge of fully autonomous driving. Our \textit{semantic scene understanding} benchmark includes tasks for 2D/3D recognition and semantic scene completion.
The former requires predicting a semantic/instance label for the visible part of the scene, while the latter aims for joint geometric completion and semantic perception that can benefit higher-level reasoning, \eg, control and planning. In our \textit{novel view synthesis} benchmark,  we establish a challenging task that requires synthesis of both RGB appearance and semantic labels at a given novel viewpoint, aiming to foster research on building fully labeled simulation environments from real-world images. Lastly, our \textit{semantic SLAM} benchmark evaluates vehicle localization as well as geometric and semantic 3D reconstruction over long sequences. 

We summarize the contributions of this paper as follows:
\vspace{-0.5cm}
\begin{itemize}
\item We present a novel georegistered dataset of suburban scenes recorded by a moving platform. The dataset comprises over $300$k images and $80$k laser scans. 
\item We create and release a WebGL-based annotation tool that allows for labeling street scenes in 3D space. Exploiting our annotation tool, we obtain 3D annotations for all static and dynamic scene elements.
\item We propose a method which transfers these labels from 3D into 2D, yielding pixel-wise semantic instance annotations. We validate our approach in ablation studies and demonstrate its potential with respect to several 2D and 3D baselines.
\item Enabled by our dense and coherent semantic instance annotations in both 2D and 3D as well as accurate vehicle and camera poses, we establish an online benchmark with novel and challenging tasks at the intersection of computer vision, graphics and robotics. We believe that our dataset and benchmarks will complement existing datasets and foster novel research towards solving the grand goal of full autonomy.
\end{itemize}

This journal paper is an extension of a conference paper published at CVPR 2016~\cite{Xie2016CVPR}. In comparison to \cite{Xie2016CVPR}, we 1) extend our annotation tool and update our inference algorithm to support the annotation of dynamic objects; 2) provide a detailed description of the annotation tool and process; 3) establish new online benchmarks with held-out test data on a set of challenging tasks; 4) propose and evaluate several baselines to bootstrap the leaderboards and assess the difficulties of the tasks.
We make our dataset\footnote{\url{http://www.cvlibs.net/datasets/kitti-360}}, utility scripts\footnote{\url{https://github.com/autonomousvision/kitti360scripts}} and annotation tool\footnote{\url{https://github.com/autonomousvision/kitti360labeltool}} publicly available.

\begin{table*}[t!]
\setlength{\tabcolsep}{2.65pt}
\begin{center}
\begin{tabular}{|p{3.0cm}|>{\centering}p{1.4cm}>{\centering}p{1.3cm}>{\centering}p{0.8cm}|>{\centering}p{1.3cm}>{\centering}p{1.2cm}>{\centering}p{1.3cm}>{\centering}p{1.2cm}>{\centering}p{1.3cm}|>{\centering}p{1.0cm}>{\centering}p{1.0cm}|>{\centering}p{0.9cm}|}  \hline
 Dataset  & \multicolumn{3}{c|}{ 2D Annotations}   & \multicolumn{5}{c|}{ 3D Annotations}    &  \multicolumn{2}{c|}{ Coherency} &  Test \tabularnewline
          &  \#Smt. Img. &  \#Ins. Img. &  Dense &  \#Smt. Pts. &  \#Ins. Pts.  & FoV Azm. & FoV Plr.  & {\#3D Bbox}  & Temporal & 3D-2D & Server \tabularnewline \hline
{ \href{http://mi.eng.cam.ac.uk/research/projects/VideoRec/CamVid/}{CamVid}~\cite{Brostow2009PRL}} & 631 & --  & \cmark & -- & -- & -- & -- & -- & \cmark &  -- &  --  \tabularnewline
{ \href{http://www.6d-vision.com/scene-labeling}{DUS}~\cite{DUS}} & 1k & -- & \cmark & -- & -- & -- & -- & -- & \cmark &  --  & -- \tabularnewline 
{ \href{http://www.cityscapes-dataset.net/}{CityScape (fine)}~\cite{Cordts2016CVPR}}    
& 5k  & 5k   & \cmark &  --  &  --  & -- & -- & -- & -- & -- & \cmark   \tabularnewline
{ \href{http://www.cityscapes-dataset.net/}{CityScape (coarse)}~\cite{Cordts2016CVPR}}  
& 20k & 20k  & -- &  --  &  --  & -- & -- & -- & -- & -- & \cmark   \tabularnewline 
{ \href{https://www.mapillary.com/dataset/vistas}{Mapillary Vistas}~\cite{Neuhold2017ICCV}}   & 25k & 25k  & \cmark &  --  &  --  & -- & -- & -- & -- & -- & --   \tabularnewline
{ \href{https://github.com/mcahny/vps}{CityScape-VPS}}~\cite{Kim2020CVPR}   & 3k & 3k & \cmark  & --  &  --  & -- & -- & -- & \cmark & -- &  --   \tabularnewline
{ \href{http://www.cvlibs.net/datasets/kitti//eval_step.php}{KITTI-STEP}}~\cite{Weber2021NEURIPSDATA}   & 19k & 19k & \cmark  & --  &  --  & -- & -- & -- & \cmark & --  & \cmark  \tabularnewline
{ \href{https://woodscape.valeo.com/dataset}{WoodScape}~\cite{Yogamani2019ICCV
}}  & 10k  & 10k  & \cmark & --    & --  & -- & -- & -- & \cmark & -- & -- \tabularnewline 
\hline
{ \href{https://github.com/WeikaiTan/Toronto-3D}{Toronto-3D}~\cite{Tan2020CVPRW}}  & --  & --  & --  & 78.3M   & --  & \textbf{360$^\circ$} & 40$^\circ$ & -- & -- & -- & -- \tabularnewline
{ \href{https://npm3d.fr/paris-lille-3d}{Paris-Lille-3D}~\cite{Roynard2018CVPRW}} & --  & --  & --  & 143.1M  & --  & \textbf{360$^\circ$} & 40$^\circ$ & -- & -- & -- & \cmark \tabularnewline 
{ \href{https://v-sense.scss.tcd.ie/dublincity/}{DublinCity}~\cite{Zolanvari2019BMVC}}    & --  & --  & --  & 260M    & --  & - & - & -- & -- & --  & -- \tabularnewline 
{ \href{http://www.semantic3d.net/}{Semantic3D.net}~\cite{Hackel2017IV}}     & --  & --  & --  & 4.0B    & --  & \textbf{360$^\circ$} & \textbf{180$^\circ$} & -- & -- & -- & \cmark \tabularnewline
{ \href{http://www.semantic-kitti.org/}{SemanticKITTI}~\cite{Behley2019ICCV}}    & --  & --  & --  & \textbf{4.5B}    & --  & \textbf{360$^\circ$} & 26.8$^\circ$ & -- & -- & -- & \cmark \tabularnewline
{ \href{https://www.argoverse.org/data.html}{Argoverse}~\cite{Chang2019CVPR}}		  & --  & --  & --  & --      & --  & --    & --& 993k & -- & -- & -- \tabularnewline
{ \href{https://level-5.global/data/}{Lyft}~\cite{Kesten2019Lyft}} & --  & --  & --  & --     & --  & --    & -- & \textbf{1.3M} & -- & -- & -- \tabularnewline
{ \href{https://waymo.com/open/}{Waymo}~\cite{Sun2020CVPR}}        		  & --  & --  & --  & --      & --  & --    & --& 12M  & -- & -- & \cmark \tabularnewline
{ \href{https://github.com/I2RDL2/ASTAR-3D#Dataset}{A*3D}~\cite{Pham2020ICRA}}       		  & --  & --  & --  & --      & --  & --    & --& 230k & -- & -- & -- \tabularnewline
\hline
{ \href{http://www.cvlibs.net/datasets/kitti/}{KITTI}~\cite{Geiger2012CVPR}}   
& 200  & 200 & \cmark & -- & --  & -- & -- & 200k & -- & -- & \cmark \tabularnewline
{ \href{http://apolloscape.auto/}{ApolloScape}~\cite{Huang2020PAMI}}    & 144k & 90k & \cmark & -- & -- & --  & -- & 70k  & -- & -- & \cmark \tabularnewline
{ \href{https://www.nuscenes.org/}{nuScenes}~\cite{Caesar2020CVPR}}   & 93k & 93k  & -- & 1.2B & 78.9M & 
\textbf{360$^\circ$} & 40$^\circ$ & 1.2M  & --  & -- & \cmark \tabularnewline
{ \href{https://www.a2d2.audi/a2d2/en/dataset.html}{A2D2}~\cite{Geyer2020ARXIV}}          & 41k  & 41k & \cmark &  387.1M  &  23.8M  & 60$^\circ$ & 30$^\circ$ & 43k & -- &  \cmark & -- \tabularnewline
{ \href{https://github.com/joe-siyuan-qiao/ViP-DeepLab}{SemKITTI-DVPS}~\cite{Qiao2021CVPR}}  & 23k  & 23k  & -- & \textbf{4.5B}   & \textbf{400M}  & \textbf{360$^\circ$} & 26.8$^\circ$ & -- & \cmark & \cmark  & \cmark \tabularnewline
{ \href{http://www.cvlibs.net/datasets/kitti-360/}{KITTI-360} (Ours)}                 
& \textbf{2$\times$ 78k} & \textbf{2$\times$ 78k}  & \cmark &  1.0B & 172.4M & \textbf{360$^\circ$} &120$^\circ$ & 68k & \cmark & \cmark & \cmark \tabularnewline 
\hline 
\end{tabular}

\end{center}
\vspace{-0.3cm}
\caption{{\bf Overview of Publicly Available Datasets.}
For pixel-level 2D annotations, we compare the number of semantic maps (\#Smt. Img.), the number of instance maps (\#Ins. Img.) and the density of the 2D semantic maps (Density). Note that the proposed KITTI-360 dataset includes images from both left and right view of the stereo camera. For 3D annotation, we show the number of semantically labeled points (\#Smt. Pts.), the number of points with instance labels (\#Ins. Pts.), the field of view with 3D semantic annotations at both azimuthal and polar directions (FoV Azm. and FoV Plr.), and the number of 3D bounding boxes (\#3D Bbox). We further compare temporal consistency and 3D-to-2D consistency of the instance labels. The last row indicates whether the dataset hosts an online evaluation benchmark with held-out ground truth.}
\label{tab:dataset_statistics}
\end{table*}

\section{Related Work}

In this section, we first discuss existing datasets in the context of autonomous driving, followed by a review of current methods for efficient (semi-automatic) label annotation.

\subsection{Datasets}

\boldparagraph{Indoor Video Datasets}
Several datasets provide annotations for video sequences captured in indoor scenes \cite{Xiao2013ICCV,Song2015CVPR,Dai2017CVPR, Chang3DV2017}. The SUN \mbox{RGB-D} dataset \cite{Song2015CVPR} provides labeled 2D polygons as well as 3D cuboids for $10$k indoor \mbox{RGB-D} images. In a closely related work, ScanNet~\cite{Dai2017CVPR} is annotated in 3D with its 2D labels directly obtained from 3D-to-2D projection based on the dense depth from RGB-D sensors. In this work, we focus on outdoor street scenes where 3D observations are much more sparse, posing a challenging task for 3D-to-2D label transfer.

\boldparagraph{Outdoor Datasets}
A number of outdoor datasets of driving scenes have been released in the literature~\cite{Behley2012ICRA,Munoz2012ECCV,Riemenschneider2014ECCV,Valentin2013CVPR,
Brostow2009PRL,Neuhold2017ICCV,Cordts2016CVPR,Zolanvari2019BMVC,
Behley2019ICCV,Caesar2020CVPR,Tan2020CVPRW,Geiger2012CVPR,Huang2020PAMI,Geyer2020ARXIV,
Kim2020CVPR,Weber2021NEURIPSDATA,MadhavanEECS2017113,Larsson2019CVPR,Yogamani2019ICCV,Qiao2021CVPR}. We summarize the most related ones in \tabref{tab:dataset_statistics}, categorized by whether they offer labels in the 2D image domain or in 3D space. 

For datasets focusing on 2D labels, CamVid \cite{Brostow2009PRL} is the first dataset for semantic segmentation in the context of self-driving. However, CamVid does not provide instance labels and only a very limited number of frames.
Both Cityscapes~\cite{Cordts2016CVPR} and Mapillary Vistas~\cite{Neuhold2017ICCV} release thousands of manually annotated 2D images. However, they do not offer temporally coherent semantic instance annotations.
Recently, Cityscape-VPS~\cite{Kim2020CVPR} extends Cityscapes by providing semantic instance labels for every 5 frames. Furthermore, KITTI-STEP~\cite{Weber2021NEURIPSDATA} offers spatially and temporally dense semantic instance annotations for the KITTI tracking dataset~\cite{Geiger2012CVPR}. %
While aforementioned works focus on perspective images, WoodScape~\cite{Yogamani2019ICCV} releases semantic instance annotations of fisheye images.
Our dataset differs from the above in that we provide not only temporally coherent semantic instance annotations for perspective images, but also omnidirectional imagery, 3D laser scans, and 3D annotations which are useful for 3D reasoning. While \cite{Cordts2016CVPR} focuses on inner-city scenes, our dataset comprises mainly suburban areas, thus both datasets complement each other (we use the same label definition to facilitate research).

Another line of works provides labels in 3D space. Toronto-3D~\cite{Tan2020CVPRW}, Paris-Lille-3D~\cite{Roynard2018CVPRW} and DublinCity~\cite{Zolanvari2019BMVC} offer annotated point clouds collected from urban environments. %
Semantic3D.net~\cite{Hackel2017IV} presents a large-scale dataset with $4$ billion points, labeled with $8$ semantic categories. SemanticKITTI~\cite{Behley2019ICCV} provides semantic labels for raw laser scans in KITTI, resulting in $4.5$ billion labeled 3D points in $28$ classes. Instead of focusing on point cloud semantic classification, Argoverse~\cite{Chang2019CVPR}, Lyft~\cite{Kesten2019Lyft}, Waymo~\cite{Sun2020CVPR}, and A*3D~\cite{Pham2020ICRA} offer 2D/3D bounding boxes and establish benchmarks for 2D/3D detection and tracking\footnote{We refer to 2D annotations as pixel-level annotations in \tabref{tab:dataset_statistics}. 2D bounding boxes are not included.}. In contrast to KITTI-360, the aforementioned datasets either lack dense annotations in images or they do not have per-point 3D annotations of stuff classes. 

Our dataset provides labels for both 2D images and corresponding 3D points. Within this category, KITTI \cite{Geiger2012CVPR} provides dense semantic information on $200$ images and $200$k 3D bounding boxes. However, KITTI does not provide dense (per-point) 3D labels on the point cloud. 
Closely related to our work, ApolloScape~\cite{Huang2020PAMI} annotates static scene elements in the 3D space\footnote{The 3D annotation has not been released yet.} and projects them to the 2D image space, followed by manual annotation of dynamic objects in images. In this work, we annotate both static and dynamic objects in 3D, providing coherent annotations for dynamic objects both in 2D and 3D. 
More recently, nuScenes~\cite{Caesar2020CVPR} and A2D2~\cite{Geyer2020ARXIV} released labels in both 2D and 3D.
However, the labels of nuScenes are manually and independently annotated in 2D and 3D, and not every pixel is labeled in 2D. In contrast, we propose to leverage labels in the 3D space to infer dense labels in the image domain, thus providing consistent labels across 2D and 3D space.
A2D2~\cite{Geyer2020ARXIV} labels 2D images and maps 2D labels to 3D to obtain per-point 3D labels. Thus, the 3D labels are limited to a small FoV of the cameras in the azimuthal direction. While A2D2 also provides 3D bounding boxes, all of them are within the FoV of the forward-facing camera.  We instead offer per-point 3D labels and 3D bounding boxes within an azimuthal FoV of 360$^\circ$.  A concurrent work, SemKITTI-DVPS~\cite{Qiao2021CVPR} provides labels in both 2D and 3D by projecting the 3D labels of SemanticKITTI to images. Compared to the projected sparse 2D labels of SemKITTI-DVPS, KITTI-360 offers dense pixel-wise labels and additionally provides 3D bounding boxes. 

There also exist several synthetic urban datasets~\cite{Gaidon2016CVPR,Ros2016CVPR,Richter2016ECCV,Cabon2020ARXIV}. However, there still exists a significant perceptual gap between the virtual and real domains~\cite{HOFFMAN2016ARXIV,Tsai2018CVPR}, making synthetic-to-real generalization difficult. 

\boldparagraph{Benchmarks} Recently, evaluation benchmarks have been widely recognized by the community. Some of the previously mentioned datasets also provide online evaluation benchmarks and held-out test data for different tasks. For instance, Cityscapes~\cite{Cordts2016CVPR} offers a benchmark suite for pixel and instance-level semantic segmentation as well as 3D vehicle detection. SemanticKITTI~\cite{Behley2019ICCV,Aygun2021CVPR} hosts lidar segmentation challenges to predict the category of every point. For datasets including both 2D and 3D annotations, KITTI~\cite{Geiger2013IJRR}, nuScenes~\cite{Caesar2020CVPR}, and ApolloScape~\cite{Huang2020PAMI} provide benchmarks on a set of vision tasks including detection, stereo, localization, multi-object tracking, and segmentation in both 2D and 3D, etc. Moving beyond the established tasks, KITTI-360 provides novel benchmarks and will hold new challenges, \eg, on novel view semantic synthesis and semantic SLAM, to foster new progress towards full autonomy.

\subsection{Methods}
\boldparagraph{Efficient Annotation}
Many works have attempted to reduce the per pixel annotation time of individual images, including classical methods~\cite{Guillaumin2014IJCV,Liu2011PAMI} and learning based methods~\cite{Castrejon2017CVPR,Acuna2018CVPR,LingCVPR2019,Andriluka2018ACMMM}. While all of these methods focus on annotating images individually, we are interested in annotating 2D video sequences as well as 3D scenes.
There is also a growing interest in autolabeling 3D shapes or 3D bounding boxes~\cite{Qi2021CVPR,Zakharov2020CVPR}. These methods are only applicable to a specific class, \eg, vehicles. We instead annotate the full 3D scene and aim to obtain coherent per-pixel 2D annotations and per-point 3D annotations.

\boldparagraph{2D Label Propagation}
Compared to annotating individual images,
video sequences offer the advantage of temporal coherence between adjacent frames. 
Label propagation techniques exploit this fact by transferring labels from a sparse set of annotated keyframes to all unlabeled frames based on color and motion information. While in some works a single foreground object is assumed~\cite{Jain2014ECCV}, here we focus on methods that can handle multiple object categories.
Towards this goal, \cite{Badrinarayanan2010CVPR} and \cite{Budvytis2010BMVC} propose a coupled Bayesian network based on video epitomes and semantic regions to propagate label information between two annotated keyframes. \cite{Zhu2019CVPR} proposes a joint propagation strategy with synthesized training samples.
To better account for errors in label propagation, \cite{Nagaraja2012DAGM} proposes a hierarchy of local classifiers for this task and \cite{Badrinarayanan2014IJCV} leverages a mixture-of-tree model for temporal association. The work of \cite{Budvytis2017ICCV} leverages label propagation as a data augmentation scheme and demonstrate improved performance on semantic segmentation.
Optical flow is also commonly used for semantic video label transfer. \cite{Raghudeep2017ICCV} uses optical flow of adjacent frames to warp network representations across time and thus propagates labels from previous frames to the current one. \cite{Zhu2017CVPR} proposes to run a convolutional sub-network only on sparse keyframes and propagate the deep feature maps to other frames via flow fields. In the indoor scenario where dense geometry is available, \cite{Reza2017IROS} proposes a method on RGB-D video propagating labels on super-pixel.

In contrast to the aforementioned methods which propagate labels in 2D, in this paper we propose to annotate both semantic and instance labels directly in 3D and then project these annotations into the 2D domain. While this approach requires a source of 3D information (\eg, SfM, stereo, laser),
it is able to produce more accurate semantic and temporally consistent instance annotations for tracking purposes. Further, our experiments indicate that annotation in 3D is more time-efficient than labeling in 2D as scene elements can be separated more easily and often project into many images of the input video sequence while being only annotated once.

\boldparagraph{3D-to-2D Label Propagation}
There are a few existing works on 3D-to-2D label transfer. Chen \etal~\cite{Chen2014CVPRb} leverage annotations from KITTI \cite{Geiger2013IJRR} as well as 3D car models to infer separate figure-ground segmentation for all vehicles in the image. In comparison, our approach reasons jointly about all objects in the scene and also handles categories for which CAD models or 3D point measurements are unavailable (\eg, ``Tree'', ``Sky''). 
Huang \etal \cite{Huang2020PAMI} also applies 3D to 2D label transfer for generating the ApolloScape dataset. In this work, labels are transferred from 3D point clouds to images with simple splatting and projection. However, 3D points are too sparse compared with image pixels, thus, setting the splatting range is not trivial. Similarly, in \cite{Dai2017CVPR}, semantic labels annotated in the reconstructed scene are projected into each frame but not all 2D pixels are covered due to missing geometry.
In addition, the two aforementioned works are limited to static scenes.

In the context of street view image segmentation, 
\cite{Xiao2009ICCV,Mustafa2017CVPR,Bruls2018ICRA,Munoz2012ECCV,Namin2015WACV,Martinovic2015CVPR} exploit the interaction between image pixels and 3D points to improve classification performance or efficiency. In comparison, our goal is to transfer ambiguous 3D primitive labels to every pixel in the image.

\begin{figure}[t]
\centering
\includegraphics[width=\linewidth]{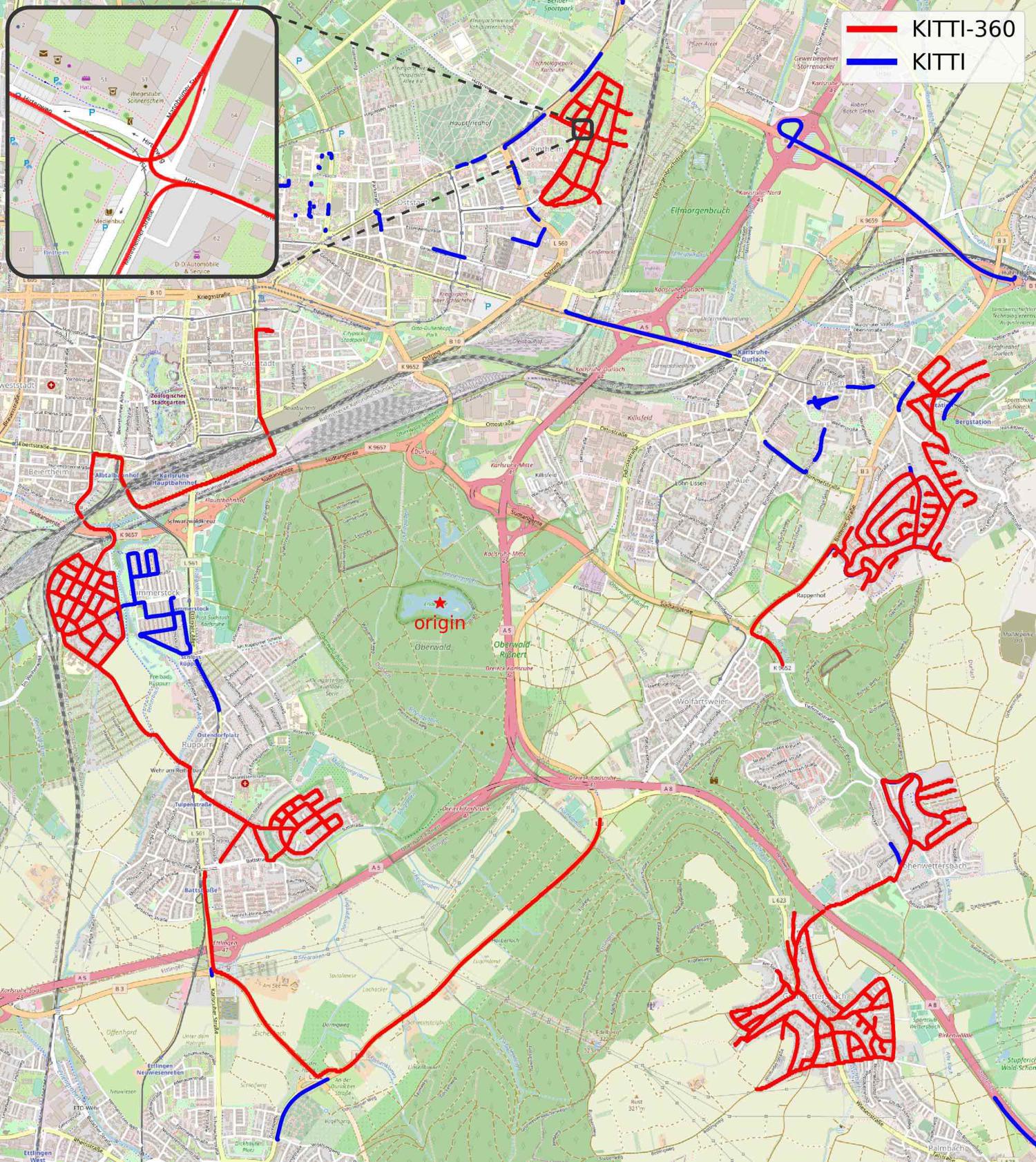}
\caption{{\bf Georegistered poses} overlaid on OpenStreetMap.}
\label{fig:osm}
\end{figure}

\section{Annotation}

In this section, we describe our data collection efforts, data preprocessing, the annotation tool, and annotation details.

\subsection{Data Collection}
\label{sec:data_collection}

For data collection, we equipped a station wagon with one $180^\circ$ fisheye camera to each side and a $90^\circ$ perspective stereo camera (baseline $60$ cm) to the front. Furthermore, we mounted a Velodyne HDL-64E and a SICK LMS 200 laser scanning unit in pushbroom configuration on top of the roof. This setup is similar to the one used in KITTI \cite{Geiger2012CVPR,Geiger2013IJRR}, except that we gain a wider field of view with the additional fisheye cameras and the pushbroom laser scanner while KITTI only provides perspective images and Velodyne laser scans with a $26.8^\circ$ vertical field of view. 
Compared to omnidirectional camera systems \cite{Schoenbein2014IROS,Schoenbein2014ICRA}, our setup benefits from increased resolution of the 3D reconstruction. Localization is provided by IMU and GPS which we fuse with visual features. \figref{fig:kitti360} (top left) illustrates our setup.

Using this setup, we recorded several suburbs of a mid-size city corresponding to over $300$k images and $80$k laser scans, covering a driving distance of 73.7km.
We estimate all vehicle and camera poses using structure-from-motion \cite{Heng2013IROS}.
More specifically, we minimize 3D reprojection errors based on all feature matches while regularizing against the GPS location. We further add loop-closures detected from LiDAR scans as regularization to complement image feature matching (which might fail on opposite-facing frames). This results in accurate georegistered camera poses. 
\figref{fig:osm} illustrates the camera poses overlaid on OpenStreetMap\footnote{\url{http://www.openstreetmap.org/}}. We also plot the camera poses of the KITTI dataset \cite{Geiger2012CVPR,Geiger2013IJRR} for reference. 
KITTI-360 follows KITTI's forward facing camera configuration, but has minimal overlap with KITTI in terms of trajectories. This allows us to split training and test data without conflicting with the KITTI dataset, \eg, avoiding the situation where a region is used for training in KITTI but testing in KITTI-360.
Following KITTI, we use Mercator projection~\cite{Osborne2008} to convert geographic coordinates to a local Euclidean coordinate frame in order to facilitate usage of the dataset. The origin of the coordinate frame is chosen as the center of the map as illustrated in \figref{fig:osm}.

\subsection{Annotation Interface}
\label{sec:data_gui}

To facilitate 3D annotation, we developed an online annotation tool based on WebGL. We release our annotation tool (see \figref{fig:annotation_gui}) as part of this project. It consists of three main components: %
a scene viewer (including 2D images and 3D scene), a semantic label selection panel, and controllers. Annotators are asked to insert 3D primitives with adjustable shapes and semantic labels into the 3D scene. 

\subsubsection{Scene}
To annotate the data while limiting transfer bandwidth, we split the collected data into batches according to the accumulated driving distances. Specifically, a single batch contains observations within a driving distance of about 200 meters (240 frames on average) and there is an overlap of 10 meters between two consecutive batches. Within one batch, we accumulate 3D points observed from the Velodyne and SICK laser scanning unit as well as the stereo camera. 

During annotation, the accumulated point clouds are downsampled to reduce data loading traffic and memory.
However, downsampling makes it hard to precisely perceive dynamic objects whose 3D observations are distributed along a moving trajectory. To allow for accurate labeling of dynamic objects, we apply a simple heuristic to detect dynamic objects, see \appref{app:dynamic_detection}. We then load all detected dynamic points at each frame into the annotation tool without down-sampling.
To help the annotators efficiently identifying dynamic objects, we highlight dynamic objects using white color as illustrated in \figref{fig:annotation_gui}.

As auxiliary visualization to the 3D point clouds, we provide fisheye and perspective images (see ``Side View'' and ``Front View'' in \figref{fig:annotation_gui}) in order to allow annotators to select and perceive the scene from different camera views. We also visualize the pose of each camera, enabling annotators to quickly select informative viewpoints.

\subsubsection{Semantic Label Panel and Controllers}
We show semantic labels with different colors in the label panel for users to choose from. 
To better assist annotators in placing the primitives accurately, we also offer easy-to-use controllers to interact with the 3D scene, including zoom, pan, rotation of the point cloud, switching data sources or camera views, and toggling annotations. We provide more details about the annotation interface in \appref{app:ann_interface}.

\subsection{Annotation Details}

We ask the annotators to annotate the 3D point clouds in the form of bounding primitives, \ie, place cuboids and ellipsoids to enclose objects in 3D and assign a semantic label to each of them. 
The 3D scene is annotated with 37 label classes, including 24 ``instance'' classes and 13 ``stuff'' classes. Labels are defined in accordance with the Cityscapes dataset \cite{Cordts2016CVPR} label definition. More details about the label definition can be found in \appref{app:label_definition}. The annotations are categorized into static and dynamic objects, which are treated differently by our annotation tool.

\begin{figure}[t]
\centering
\includegraphics[width=\linewidth]{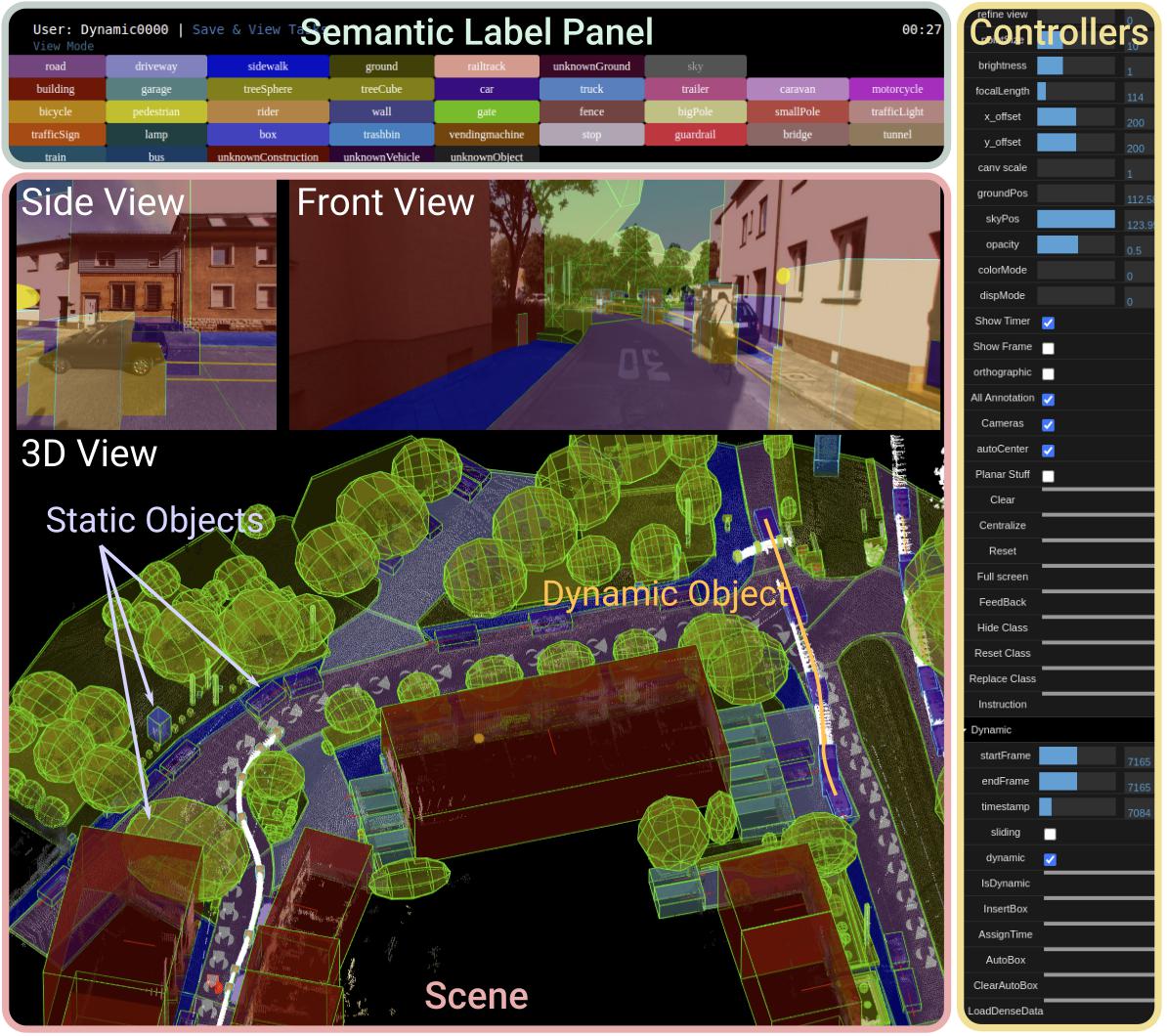}
\caption{{\bf Annotation Interface.} Our interface consists of three main components: scene view (perspective views and 3D view), semantic label panel, and controllers. }
\label{fig:annotation_gui}
\end{figure}

\subsubsection{Static Objects Annotation}
Static labels can be further classified into two categories: ``stuff'' and ``instance''. 
For \emph{instance} classes, each object is constrained to be associated with only one cuboid primitive, representing both semantic and instance labels of this object.
We ask the annotators to tightly enclose the point clouds with the bounding primitives. 
For \emph{stuff} classes, which usually have irregular shapes, annotators are allowed to use multiple cuboids or ellipsoids to roughly enclose the 3D points of the target objects.

We also provide a ``planar'' annotation option for stuff categories on the ground such as ``Road'' and ``Sidewalk''. Using this option, we allow annotators to draw a 2D polygon representing the ground object's boundary in bird's eye view. The interface then automatically estimates the height of the polygon based on the surrounding 3D geometry and extrudes the 2D polygon into 3D along the vertical direction to enclose corresponding 3D ground points. We provide more details in \appref{app:ann_ground}.

\subsubsection{Dynamic Objects Annotation}
Dynamic objects mainly comprise moving vehicle and pedestrian instances. In contrast to ApolloScape~\cite{Huang2020PAMI} which annotates static objects in 3D and dynamic objects in 2D respectively, we annotate both static and dynamic objects in 3D space. 
However, compared to static objects, annotating dynamic objects in 3D outdoor scenes is more challenging as individual dynamic objects in the 3D reconstruction are hard to perceive and distinguish. Moreover, we need to label not only where the moving instance is, but also ``when'' the instance appears, requiring the annotation of moving 3D bounding boxes over time.
A na\"ive solution is to place a 3D bounding box in every frame where the dynamic object is present. However, such an annotation process would be intractably slow. Thus, we instead implement a semi-automatic annotation scheme to reduce label time. 
Specifically, we minimize the effort required by annotators by making two assumptions: the size of the dynamic object is fixed over time and its trajectory is smooth. Under these assumptions, the required annotation is reduced to the size of a single 3D primitive and the pose of this primitive at several keyframes. Our annotation tool then automatically places the remaining primitives along the trajectory, see \appref{app:dynamic_annotation} for more details.

\subsection{Annotation Procedure}
We annotated 379 batches in total, assigning one batch to one annotator. To control the annotation quality, we train and evaluate the annotators based on multiple pilot tasks until they have proven
qualified for the full task. We also regularly verify their annotation quality and ask them for correction if necessary. We further identify a few annotators who consistently produced high-quality labels and ask them to cross-check other annotators' quality.
Our annotation interface simplifies the detection and correction of annotation errors compared to annotating image sequences, which requires corrections across multiple frames. \figref{fig:annotation_gui} shows parts of an annotated batch via our web interface.

\subsection{Annotation Time}
\label{sec:annotation_time}
On average, annotating one full batch ($\sim$ 240 frames) in 3D required about 3 hours. Thus, our annotators spend only 3 $\times$ 60 $/$ 240 = 0.75 minutes for ``annotating'' one image. 
In comparison, 7 minutes are required for coarse annotation of semantic instance labels in the image domain, and 1.5 hours for pixel-accurate annotations as discussed by the creators of the Cityscapes dataset~\cite{Cordts2016CVPR}.

\section{Label Transfer Method}\label{sec:method}

In this section, we first provide an overview of our method for transferring the 3D annotation to semantic instance annotations in 2D. Next, we formally introduce the model and discuss parameter learning and inference.
\subsection{Overview}
\label{sec:label_transfer_overview}
Given 3D annotations, we are interested in generating dense semantic instance annotations for all images and all 3D points.
To incorporate inductive biases about image formation and label smoothness, we explore a Conditional Random Field (CRF) model which reasons jointly about the labels of the \textit{3D points} and \textit{all pixels} in the image. 
In practice, we apply the CRF at every timestamp independently to keep inference tractable. Despite independent inference, we are able to obtain consistent results over multiple frames thanks to the shared 3D annotations. We also experimented with inference over multiple adjacent frames but did not observe measurable improvements.

Let $\cB_t=\{\{b^n\}, \{b_t^m\}\}$ denote all 3D annotations available at timestamp $t$. Here, $b^n$ and $b^m_t$ correspond to 3D bounding primitives of static and dynamic objects respectively, with $n$ and $m$ indexing each primitive. 
Note that a static primitive $b^n$ is used at all timestamps (if visible) whereas a dynamic primitive $b^m_t$ is only included in $\cB_t$ when it is labeled to appear at timestamp $t$. \figref{fig:model_dense_crf_projection} illustrates static and dynamic bounding primitives as well as their projection into the 2D image domain. With this design, our framework allows for annotating the same object using a unique instance ID across the entire sequence as well as across 2D and 3D.

Let $\cP_t$ denote the set of image pixels at timestamp $t$ and $\cL_t$ denote the visible 3D points at the same timestamp. The CRF model is defined over all elements in $\cP_t$ and $\cL_t$. To obtain more complete 3D information, $\cL_t$ fuses stereo and laser scans over multiple frames. 
\del{Towards this goal,} We first fuse points covering static parts of the scene, and then 
accumulate points of each dynamic object according to its bounding primitives and insert them into the static scene depending on the location of ${b^m_t}$. We provide more details regarding the accumulation of static and dynamic 3D points in \appref{app:pcd_accumulation}.

\subsection{Model}
\label{sec:label_transfer}

\begin{figure*}[t]
\center
\begin{minipage}[c]{0.503\textwidth}
\begin{subfigure}{\linewidth}
\includegraphics[width=\linewidth]{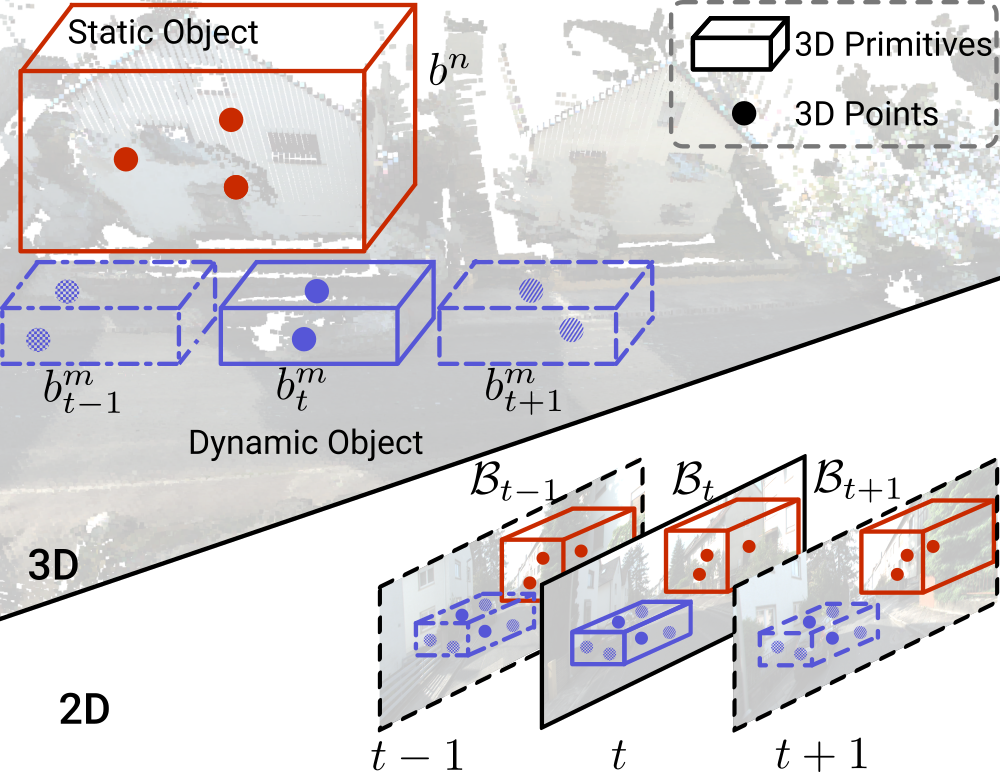}
\caption{Projection of 3D Annotations at Multiple Timestamps}
\label{fig:model_dense_crf_projection}
\end{subfigure}%
\end{minipage}
\hspace{1cm}
\begin{minipage}{0.3\textwidth}
\begin{subfigure}{\linewidth}
\includegraphics[width=\linewidth]{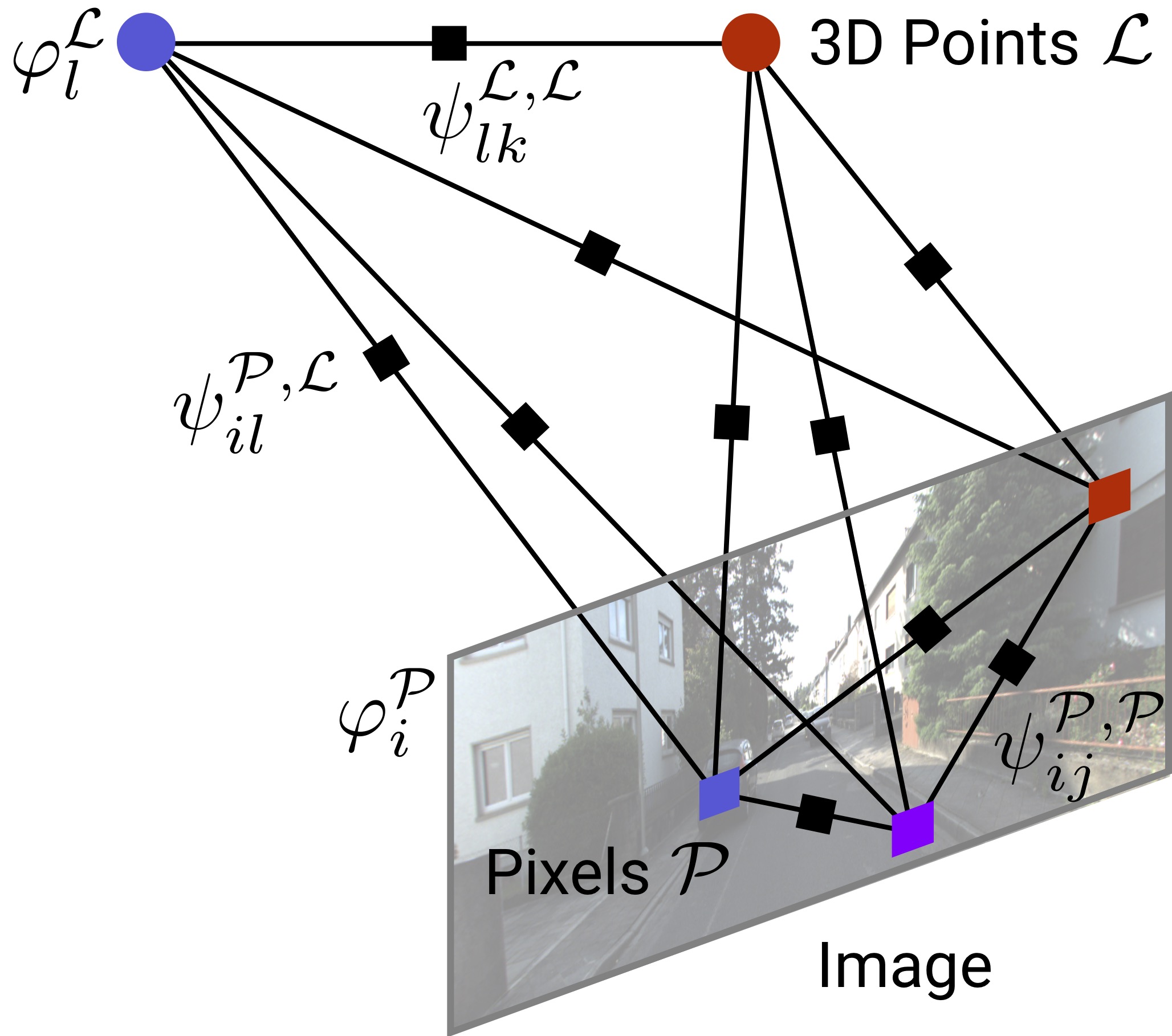}
\caption{Graphical Model at Timestamp $t$}
\label{fig:model_dense_crf_graph}
\end{subfigure}%
\vspace{0.2cm}
\begin{subfigure}{\linewidth}
\includegraphics[width=\linewidth]{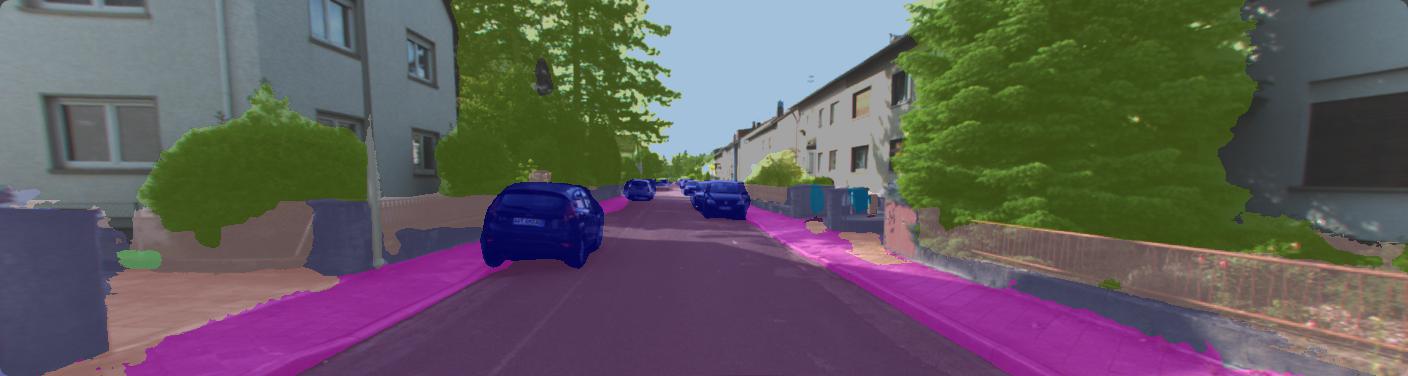}
\caption{Inference Result at Timestamp $t$}
\label{fig:model_dense_crf_result}
\end{subfigure}%
\end{minipage}
\caption{{\bf 3D-to-2D Label Transfer.} (\subref{fig:model_dense_crf_projection}) We illustrate the 3D-to-2D projection of static and dynamic object annotations. A static 3D primitive is projected to multiple frames while a dynamic 3D object is projected only into the corresponding frame.  (\subref{fig:model_dense_crf_graph}) Factor graph representation of our model. Note that the CRF model is defined over all pixels and visible 3D points at a single timestamp. (\subref{fig:model_dense_crf_result}) We show the semantic inference result at timestamp $t$.}
\label{fig:model_dense_crf}
\end{figure*}

We now formalize the CRF model applied at every frame as illustrated in \figref{fig:model_dense_crf_graph}.
Note that our 3D annotations are sparse and noisy, \ie, 3D points can carry none, one or multiple labels due to overlapping bounding primitives in 3D. The algorithm described in this section is designed to resolve these situations and infers marginal estimates for all 3D points and pixels in the image.

As the CRF model is applied at every frame independently, we drop the dependency on timestamp $t$ of $\cB, \cL$ and $\cP$ for simplicity.
For each pixel $i\in\cP$ and each 3D point $l\in\cL$, we specify random variables $s_i$ and $s_l$ taking values from the set of semantic (or instance) labels $\{1,\dots,S\}$, where $S$ denotes the number of classes.
For instance inference, we assign a unique ID to each object which projects into the image. Thus, semantic and instance inference can be treated equally under our model and we will refer to both as ``semantic labels'' in the following.
Note that there is no need to distinguish static or dynamic objects in the single frame-based CRF model. Still, we are able to retrieve whether a pixel or a 3D point belongs to a dynamic object or not according to its instance ID.

Let $\bs = \{s_i|i\in\cP\} \cup \{s_l|l\in\cL\}$ denote the set of semantic labels. Dropping all dependencies on the image and point cloud for clarity we specify our CRF in terms of the following Gibbs energy function:
\begin{align}
E(\bs) = & \sum_{i \in \cP} \varphi_{i}^{\cP}(s_i) 
+ \sum_{l \in \cL} \varphi_{l}^{\cL}(s_l) \label{eq:crf_energy} 
+ \hspace{-0.1cm} \sum_{i,j\in\cP} \hspace{-0.1cm}\psi_{ij}^{\cP,\cP}(s_i,s_j) \nonumber\\
&+ \hspace{-0.1cm} \sum_{l,k\in\cL} \hspace{-0.1cm} \psi_{lk}^{\cL,\cL}(s_l,s_k) 
+ \hspace{-0.3cm} \sum_{i \in \cP, l \in \cL} \hspace{-0.2cm} \psi_{il}^{\cP, \cL}(s_i,s_l) 
\end{align}
with unary potentials $\varphi(\cdot)$ and pairwise potentials $\psi(\cdot)$. For notational clarity, we omit all conditional dependencies on the input images, 3D points and 3D annotations. 

\noindent{\bf Pixel Unary Potentials:}
The pixel unary potentials $\varphi_i^{\cP}(s_i)$ encode the likelihood of pixel $i$ taking label $s_i$
\begin{equation}
\varphi_i^{\cP}(s_i) =  w^{\cP}_1(s_i) \, \xi^{\cP}_i(s_i) -\,w^{\cP}_2(s_i) \log p^{\cP}_i(s_i)
\end{equation}
where $w^\cP_1$ and $w^\cP_2$ denote learned feature weights.
Our first constraint $\xi^{\cP}_i(s_i)$ determines the set of admissible labels and is obtained by projecting all 3D bounding primitives $\cB$ (which are an upper bound on the objects' extent) into the image.
We formulate the constraint via a binary feature $\xi^{\cP}_i(s_i)\in\{0,1\}$ which takes $0$ for pixel $i$ if its ray passes through a primitive of class $s_i$, and $1$ otherwise.

In addition, we exploit a data-driven approach in order to obtain a per-pixel probability distribution over semantic labels $p^{\cP}_i(s_i)$. Specifically, we project all non-occluded and uniquely labeled sparse 3D points into the image plane, and use these sparse projections as supervision to train a semantic segmentation network (PSPNet~\cite{Zhao2017CVPR}) on the entire dataset. 
The output of the network's last layer is taken as the probability distribution.  We also augment the training dataset using Cityscape images and labels~\cite{Cordts2016CVPR} to enable the model to learn accurate object boundaries which is difficult to learn based on the projection of sparse and noisy LiDAR point clouds.
As the semantic segmentation model does not distinguish instances, we further adopt a state-of-the-art instance segmentation method~\cite{Xiong2019CVPR} to obtain instance hypotheses for ``car'', ``truck'', and ``pedestrian''. Thus, we effectively exploit the inductive biases of modern neural network architectures and co-training on related labeled datasets. As demonstrated in \appref{app:ablation_object_boudary}, this leads to a significant improvement at object boundaries.

\noindent{\bf 3D Point Unary Potentials:}
The 3D point unary potentials $\varphi_l^{\cL}(s_l)$ encode the likelihood of 3D point $l$ taking label $s_l$:
\begin{equation}
\varphi_{l}^{\cL}(s_l) = -w^{\cL}(s_l)  \, \xi^{\cL}_l(s_l)
\end{equation}
where $\xi^{\cL}_l(s_l)$ denotes a feature which takes $0$ if the 3D point $l$ lies within a 3D primitive of class $s_l$ within $\cB$, and $1$ otherwise. As the ``sky'' class can't be modeled with primitives, we set $\xi^{\cL}_l(s_l)$ to $0$ if $s_l$ takes the label ``sky''. Additionally, we create ``virtual sky points'' at infinity for all pixels whose ray doesn't intersect any 3D primitive. Note that these pixels must correspond to sky regions as we assume that the scene is densely annotated, hence each object is contained in one or several bounding 3D primitive(s).

\noindent{\bf Pixel Pairwise Potentials:}
Our dense pairwise term encourages semantic label coherence and connects all pixels in the image via Gaussian edge kernels following \cite{Kraehenbuehl2011NIPS}
\begin{eqnarray}
&&\psi_{ij}^{\cP,\cP}(s_i,s_j) ~=~
w_1^{\cP,\cP}(s_i,s_j)\exp\left\{-\frac{\lVert  \bp_{i} - \bp_{j} \rVert^2}
{2\,\theta_1^{\cP,\cP}}\right\} \nonumber\\
&&+\,
w_2^{\cP,\cP}(s_i,s_j)\exp\left\{-\frac{\lVert  \bp_{i} - \bp_{j} \rVert^2}
{2\,\theta_2^{\cP,\cP}}-\frac{\lVert  \bc_{i} - \bc_{j} \rVert^2}
{2\,\theta_3^{\cP,\cP}}\right\}\hspace{0.6cm}
\end{eqnarray}
where $\bp_i$ is the 2D location of pixel $i$ and $\bc_i$ denotes its color value.
Further, $w_1^{\cP,\cP}$ and $w_2^{\cP,\cP}$ are learned pairwise feature weights and $\theta^{\cP,\cP}$ parameterizes the kernel width.

\noindent{\bf 3D Pairwise Potentials:}
Similarly, we apply a Gaussian edge kernel to encourage label consistency between 3D points based on their 3D location and surface normals
\begin{eqnarray}
\psi_{lk}^{\cL, \cL}(s_l,s_k) &=& w^{\cL,\cL}(s_l,s_k)\\
&\times&\exp\left\{-\frac{\lVert \bp_{l}^{3d} - \bp_{k}^{3d} \rVert^2}
{2\,\theta_1^{\cL,\cL}}-\frac{{(n_{l} - n_{k})}^2}
{2\,\theta_2^{\cL,\cL}}\right\}\nonumber
\end{eqnarray}
where $\bp^{3d}_l$ is the 3D location of point $l$ and $n_l$ denotes the vertical (up) component of its normal. We use the normal's z-component as it is the most discriminative cue for label changes between horizontal (\eg, road, sidewalk) and vertical (\eg, side of car, wall) surfaces. We estimate the respective normals using principal component analysis in a local neighborhood around each 3D point.

\noindent{\bf 2D/3D Pairwise Potentials:}
Finally, we encourage coherence between all 3D points and the image pixels
\begin{equation}
\psi_{il}^{\cP, \cL}(s_i,s_l) =
w^{\cP, \cL}(s_i,s_l)\exp\left\{-\frac{\lVert \bp_{i} -\bpi_l \rVert^2}
{2\,\theta^{\cP, \cL}}\right\}
\end{equation}
where $\bpi_l$ denotes the projection of the 3D laser or stereo point $l$ onto the image plane. Importantly, we project only points into the image which are likely to be visible. We determine these points by meshing the 3D point cloud using the ball-pivoting method of Bernardini \etal \cite{Bernardini1999VCG}, and considering only 3D points in front of the mesh. We also experimented with multi-view reconstruction approaches \cite{Jancosek2011CVPR} for mesh generation, but obtained better results using this simpler approach.
As applying the meshing algorithm independently for every frame is time-consuming, we generate meshes on entire batches, processing the static part and dynamic objects independently. This allows us to reuse the mesh of the static part for all frames of a batch.

\subsection{Learning and Inference}
\label{sec:learning_inference}

This section describes inference and parameter estimation in our label transfer model.

\noindent{\bf Inference:}
At test time, we are interested in estimating the marginal distribution of each semantic or instance label in $\bs$ under our model, specified by the Gibbs distribution defined in \eqref{eq:crf_energy}. A likely configuration can then be estimated by variable-wise maximization of these marginals.
As our graphical model is loopy, exact inference in polynomial time is
intractable. We thus resort to variational inference and approximate the
probability distribution on $\bs$ by replacing it with a factorized mean
field distribution $Q(\bs) = \prod_{i\in \cP\cup \cL} Q_i(s_i)$.
This mean field approximation can be  computed efficiently using bilateral filtering \cite{Kraehenbuehl2011NIPS}.
As our model comprises three sets of densely connected variables (namely $\cP$, $\cL$ and $\cP\leftrightarrow\cL$), we exploit the algorithm of \cite{Kiefel2014ECCV,Vineet2013EMMCVPR} which generalizes \cite{Kraehenbuehl2011NIPS} to multiple fields. \figref{fig:model_dense_crf_result} illustrates the inference result for a single frame, overlaid on the corresponding input image. Moreover, we obtain an uncertainty estimate for each pixel/3D point by computing the entropy over the respective marginal distribution. We will use this estimate in \secref{sec:benchmark} to weigh the evaluation metrics according to the confidence of our label estimates.

\noindent{\bf Learning:}
We employ empirical risk minimization in order to learn the parameters in our model, considering the univariate logistic loss, defined as
$\Delta(s) = -\log \left(P(s)\right)$
where $P(\cdot)$ denotes the marginal distribution at the respective site. Let
us subsume all model parameters into $\Theta = \{w^\cP_1, w^\cP_2, w^\cL,w^{\cP,\cP}_1,w^{\cP,\cP}_2,w^{\cP,\cL},w^{\cL,\cL}\}$. We define our minimization objective $f(\Theta)$ as the regularized univariate logistic loss:
\begin{eqnarray}
f(\Theta) &=& \sum_{n=1}^N \, \sum_{i\in \cP} - \log \left(Q_{n,i}(s_{n,i}^*)\right) + \lambda \, C(\Theta)\hspace{0.5cm}
\end{eqnarray}
Here, $N$ is the number of training images, $s_{n,i}^*$ denotes the ground truth semantic label and $Q_{n,i}(\cdot)$ the approximate marginal at pixel $i$ in image $n$, calculated via mean field approximation. $C(\Theta)$ is a quadratic regularizer on the parameter vector $\Theta$. We whiten all features and use a single value $\lambda$ which we select via cross-validation on the training set.
For learning the instance segmentation parameters we exploit the same loss $f(\Theta)$ as for semantic segmentation, but assign unique labels to each individual object, e.g., different cars will be assigned different labels even if they occlude each other. In order to associate 2D ground truth instances with 3D instances we project all visible 3D points into the image and find a consensus via the majority vote which gave good results in practice. As the number of instances per semantic class varies between images, we learn intra- and inter-class pairwise potentials using parameter tying.
We optimize the objective function $f(\Theta)$ using stochastic gradient descent and obtain $\partial Q / \partial \Theta$ using auto differentiation. 
We make use of the ADADELTA algorithm \cite{Zeiler2013ARXIV} with decay parameter $0.95$ and $\epsilon=10^{-8}$, and randomly sample a batch of $16$ training images at each iteration for which all gradients can be computed in parallel.

\section{Label Transfer Evaluation}
\label{sec:results}

In this section, we first introduce the datasets we use for training and evaluating our label transfer method. Next, we evaluate our method in ablation studies and compare it against several label transfer baselines. 
Finally, we also show qualitative results of our method.

\subsection{Training and Evaluation Data}
We manually annotate a set of images with pixel-wise ground truth to train and evaluate our label transfer method. The \textit{training} set contains $125$ images selected from diverse scenarios such that a substantial amount of pixels are labeled within each class. These training images are different from those used in our conference version \cite{Xie2016CVPR}. We create this new training set following the label definition of CityScapes \cite{Cordts2016CVPR} as \cite{Xie2016CVPR} considers fewer classes.
To enable comparison to 2D label transfer methods which require images with large overlapping regions, we additionally annotate $240$ adjacent frames from $13$ different suburbs in equidistant steps of $5$ frames in the 2D image domain for \textit{evaluation}. The evaluation set has no spatial overlapping with the training set, allowing us to assess the generalization ability of our method.
We evaluate our label transfer method on static and dynamic objects separately.
Following \cite{Xie2016CVPR}, the performance of static objects is evaluated on $120$ densely labeled frames from $5$ suburbs containing the most frequently occurring $14$ classes. The remaining $120$ frames are sampled from $8$ different suburbs which contain dynamic objects. For these frames we label the dynamic objects while leaving the static region unannotated.
We consider $7$ common dynamic objects, see \appref{app:label_transfer_result_quantitative} for details.

\subsection{Quantitative Evaluation}
\label{sec:results_quantitative}

This section presents our quantitative evaluation on semantic and instance segmentation. We compare our method with several label transfer baselines and conduct ablation studies.

\subsubsection{\bf Semantic Segmentation Transfer}
For evaluating semantic segmentation transfer performance, we measure overall performance by the mean intersection over union (mIoU) 
and the average pixel accuracy (Acc). 
While \cite{Xie2016CVPR} evaluates the \textit{weighted} mean IoU which is biased by object occurrences, we follow Cityscapes \cite{Cordts2016CVPR} and measure the mean IoU without weighting. For all experiments, we provide results for individual classes in \appref{app:label_transfer_result_quantitative}.

\boldparagraph{Baselines}
We compare our method to several 2D to 2D label transfer methods on both static and dynamic objects in \tabref{tab:baseline}. Here, the task is to predict the center frame from two annotated images ($\pm5$ frames corresponding to $0.5$ seconds of driving or $\sim5$ meters travel distance).
Our first baseline (``Label Prop.'') is the label transfer approach presented in \cite{Vijayanarasimhan2012ECCV}. To ensure that all baselines have access to the same information, we do not select frames actively but use equidistantly spaced frames for all methods.
We construct a second baseline (``Sparse Track. + GC'') using the feature tracking approach of \cite{Sundaram2010ECCV} to propagate semantic labels from the two closest labeled frames to the target frame. To densify the label map, we apply graph cuts (GC) with contrast sensitive edge potentials \cite{Boykov2004PAMI}.
In order to evaluate the value of 3D information, we implemented a third baseline (``3D Prop. + GC'') which works similar to the previous one, but replaces the sparse tracking part with correspondences obtained by transferring pixels of the two closest labeled frames to the target image via the visible vertices of our 3D mesh followed by graph cuts propagation. 

While all aforementioned baselines require labeled adjacent frames as input at inference time, we consider two more methods that generalize to arbitrary frames.
First, we train the segmentation model of Kr\"ahenb\"uhl \etal \cite{Kraehenbuehl2011NIPS} (``Fully Conn. CRF'') which was also used in \cite{Xie2016CVPR} and which uses a similar inference algorithm as our label transfer method on all annotated adjacent frames of the test sequence.
Finally, we evaluate the deep semantic segmentation network~\cite{Zhao2017CVPR} (``PSPNet'') that also provides dense unary information for our method.  As discussed in \secref{sec:label_transfer}, this model is trained on non-occluded sparse 3D projections combined with the CityScapes training set~\cite{Cordts2016CVPR}. 
Note that neither PSPNet nor our method has access to adjacent annotated frames for training or inference.

We further consider several 3D to 2D label transfer baselines that exploit our 3D annotations without requiring equidistantly labeled 2D annotations. Specifically, we project 3D primitives, meshes or visible 3D points into the 2D image domain, followed by graph cut inference (``3D Primitives + GC''; ``3D Mesh + GC''; ``3D Points + GC'').

\begin{table}[t]
\setlength{\tabcolsep}{2.65pt}
\vspace{0.2cm}
\begin{center}
\begin{tabular}{|p{3.4cm}|>{\centering}p{1.2cm}|>{\centering}p{0.75cm}>{\centering}p{0.75cm}|>{\centering}p{0.75cm}>{\centering}p{0.75cm}|}  \hline
\multirow{2}*{\small Method} & \small Label & \multicolumn{2}{c|}{\small Static} &  \multicolumn{2}{c|}{\small Dynamic}  \tabularnewline
							& source & mIoU &  Acc & mIoU &  Acc  \tabularnewline \hline 
{\small Label Prop.\cite{Vijayanarasimhan2012ECCV} }& 2D & 49.0 & 81.0 & 37.2 & 59.1\tabularnewline
{\small Sparse Track. + GC \cite{Sundaram2010ECCV} }& 2D &51.2 & 79.1 & 8.2 & 12.5\tabularnewline
{\small 3D Prop. + GC }& 2D &72.1 & 87.4 & 14.5 & 21.7\tabularnewline
{\small Fully Conn. CRF \cite{Kraehenbuehl2011NIPS} } & 2D &63.6 & 88.7 & -- & --\tabularnewline
{\small PSPNet \cite{Zhao2017CVPR} }& CS + 3D &67.2 & 90.4 & -- & --\tabularnewline
\hline
{\small 3D Primitives + GC } & 3D &49.4 & 73.4 & -- & --\tabularnewline
{\small 3D Mesh + GC }& 3D& 66.8 & 85.7 & -- & --\tabularnewline
{\small 3D Points + GC }& 3D& 72.6 & 87.8  & -- & --\tabularnewline
{\small Proposed Method }& 3D&\textbf{81.2} &\textbf{93.1} & \textbf{63.5} & \textbf{94.1}\tabularnewline
\hline
\end{tabular}
\end{center}
\vspace{-0.4cm}
\caption{{\bf Comparison to Label Transfer Baselines on Semantic Segmentation Transfer}. We compare our method to 2D label transfer baselines (top) and to 3D to 2D label transfer baselines (bottom). Label source: ``2D'': Labeled neighboring image frames, ``CS'': Cityscapes training images, ``3D'': 3D bounding primitives.}
\label{tab:baseline}
\end{table}

\boldparagraph{Static Objects}
\tabref{tab:ablation_semantic} (left) shows the comparison on $120$ consecutive images of static objects.\footnote{The results differ slightly from those presented in \cite{Xie2016CVPR} as 1) we updated the ground truth labels to be consistent with the extended label definition and 2) we measure mIoU following Cityscapes~\cite{Cordts2016CVPR} while \cite{Xie2016CVPR} reports weighted mIoU.} From the 2D label transfer baselines shown at the top, the mesh transfer method which uses projected 3D information performs best in terms of mIoU. Furthermore, and maybe surprisingly, the sequence-specific fully connected CRF model performs on par or even better than special purpose label transfer methods. \del{According to our experiments,} This is caused by the fact that optical flow (as used in \cite{Vijayanarasimhan2012ECCV,Sundaram2010ECCV}) often fails for street scenes like ours due to large displacements, perspective distortions, textureless regions and challenging lighting conditions. 
Interestingly, PSPNet achieves the best accuracy while performing worse on mIoU. Despite obtaining superior results on large objects (\eg, ``Building''), it struggles with less-occurring classes such as ``Trailer'' and ``Gate''.

The bottom half of \tabref{tab:baseline} (left) compares the proposed method with respect to the 3D to 2D label transfer baselines.
As evidenced by our results, simply projecting 3D primitives or meshes into the image and smoothing via GC does not perform well due to the crude approximation of the geometry. 
Better results are obtained when projecting the visible 3D points followed by spatial propagation.
Finally, we observe that all baselines are outperformed by the proposed method (last row). Note that we also map the $37$ semantic labels of our 3D annotations to the most common $14$ categories considered in the static evaluation images (see \appref{app:label_definition}) for all 3D to 2D label transfer methods.

\begin{table}[t]
\setlength{\tabcolsep}{2.65pt}
\vspace{0.2cm}
\begin{center}
\begin{tabular}{|p{3.5cm}|>{\centering}p{1.0cm}>{\centering}p{1.0cm}|>{\centering}p{1.0cm}>{\centering}p{1.0cm}|}  \hline
\multirow{2}*{\small Method} & \multicolumn{2}{c|}{\small Semantic} &  \multicolumn{2}{c|}{\small Instance}  \tabularnewline
							 &  mIoU &  Acc & mIoU &  Acc\tabularnewline \hline 
{\small LA }&67.6 & 88.7 & 72.3 &86.7 \tabularnewline
{\small LA+3D }&70.4 & 89.7 & 72.9 &88.7 \tabularnewline
{\small LA+PW }&66.6 & 87.9 & 72.9 &85.8\tabularnewline
{\small LA+PW+CO }&76.8 & 91.8 & 81.6 &91.0\tabularnewline
{\small LA+PW+CO+3D }&78.2 & 92.4 & 83.6 &91.7\tabularnewline
{\small Full Model }&\textbf{81.2} & \textbf{93.1} & \textbf{83.7} & \textbf{91.8}\tabularnewline
\hline
{\small Full Model ($90\%$)}&88.3 & 96.0 & 89.0 &94.9\tabularnewline
{\small Full Model ($80\%$)}&92.5 & 97.6 & 91.3 &96.6\tabularnewline
{\small Full Model ($70\%$)}&\textbf{94.3} & \textbf{98.4}& \textbf{92.7} & \textbf{97.4}\tabularnewline
\hline
\end{tabular}

\end{center}
\vspace{-0.4cm}
\caption{{\bf Semantic Instance Segmentation Transfer Ablation} evaluated on static objects. The components are abbreviated as follows:
LA = local appearance ($p^{\cP}$), PW = 2D pairwise constraints ($\psi^{\cP,\cP}$), CO = 3D primitive constraints ($\xi^{\cP}$), 3D = 3D points ($\varphi^{\cL}$,$\psi^{\cP, \cL}$), Full Model = all potentials including 3D pairwise constraints ($\psi^{\cL,\cL}$). Percentages denote fractions of estimated pixels.}
\label{tab:ablation_semantic}
\end{table}

\boldparagraph{Dynamic Objects}
We evaluate our method on dynamic objects against 2D label transfer baselines. Here, we consider all static regions as a single background class during evaluation. Note that we neglect ``Fully Conn. CRF'' and ``PSPNet'' as both methods address semantic segmentation and thus cannot distinguish static and dynamic objects within the same class. 
\tabref{tab:baseline} (right) shows that our method also outperforms all 2D label transfer baselines on dynamic objects.
While the mIoU is calculated over a different set of classes, the average performance of our method on dynamic objects is slightly degraded compared to our result on static objects. Labeling of dynamic objects is more challenging in our annotation pipeline for two reasons:
Since we accumulate 3D points of dynamic points according to the annotated bounding primitives over multiple frames, slight misalignments of the primitives may lead to inaccurate accumulation and thus erroneous 3D cues. Furthermore, the accumulation of deformable objects (``Rider'', ``Person'') leads to noisy 3D point clouds. Despite these challenges, our method achieves satisfying performance on all dynamic objects.

\boldparagraph{Annotation Time Comparison}
While all 2D methods require every 10th frame to be labeled, our method (as well as the other 3D baselines) requires 3D annotations in the form of 3D primitives. Assuming $60$ minutes annotation time per image, this amounts to $20$ hours of annotation time per batch of $200$ frames when labeling one 2D image every $10$th frame, while the respective 3D annotations for this scene can be obtained in about $3$ hours. %
This gain multiplies with the frame rate and the number of cameras (our setup has four).

\begin{figure*}[t]
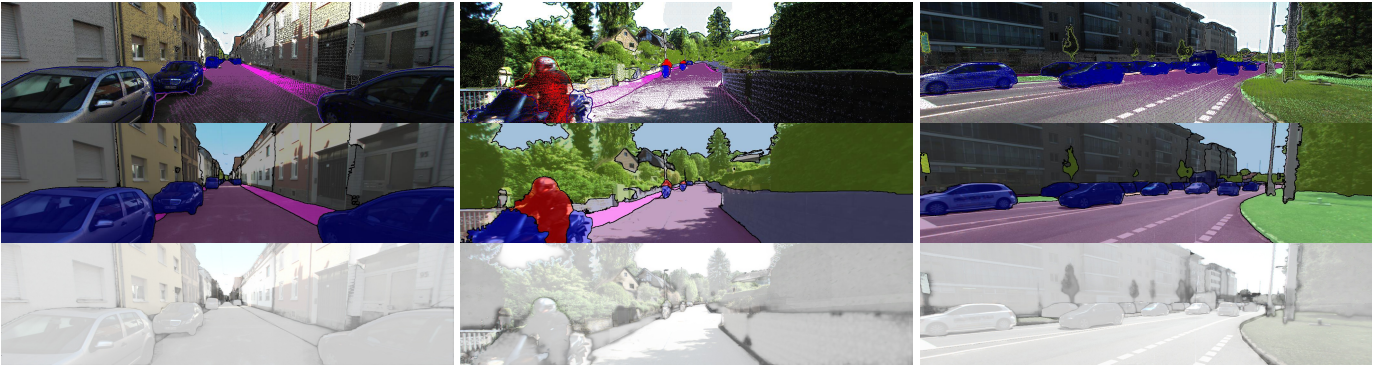

\def\qualwidth{0.33}
\def\qualmargin{0.1}
\newcommand{\qualresult}[4]{%
\begin{minipage}{#1\linewidth}%
\includegraphics[width=\linewidth]{#2/#3_pts.jpg}\\%
\includegraphics[width=\linewidth]{#2/#3_label.jpg}\\%
\includegraphics[width=\linewidth]{#2/#3_conf.jpg}\\%
\vspace{-0.6cm}
\end{minipage}%
\hspace{\qualmargin cm}%
}
\qualresult{\qualwidth}{results/semantics/2013_05_28_drive_0000_sync}{0000006440}{Scene 2}%
\qualresult{\qualwidth}{results/dynamic/2013_05_28_drive_0005_sync}{0000005695}{Scene 4}%
\qualresult{\qualwidth}{results/dynamic/2013_05_28_drive_0010_sync}{0000000680}{Scene 6}\\
\caption{{\bf Qualitative Results on Semantic Instance Segmentation Transfer.} Each subfigure shows from top-to-bottom: the input image with the projected 3D points and inferred semantic segmentation boundaries, the inferred semantic instance segmentation, as well as the confidence map of the inferred label with bright and dark colors indicating high and low confidence, respectively. See supplementary material and text for details. The first scene (1st column) contains only static objects while the others (2nd and 3rd columns) also contain dynamic objects. 
\label{fig:qualitative_results_static}}
\end{figure*}

\boldparagraph{Ablation Study}
We validate the importance of the individual components of our model on semantic segmentation in \tabref{tab:ablation_semantic} (upper left), evaluated on the densely labeled images of static objects. Starting with the appearance classifier $p^{\cP}$ trained on the projected sparse 3D points (``LA''), we incrementally add the terms $\varphi^{\cL}$,$\psi^{\cP, \cL}$ related to the 3D points (``3D''), the semantic pairwise term $\psi^{\cP,\cP}$ between pixels (``PW''), the 3D primitive constraints $\xi^{\cP}$ (``CO'') and finally the 3D pairwise constraints $\psi^{\cL,\cL}$ as specified in \eqref{eq:crf_energy}.
We note that each component is able to increase performance.
We obtain the largest improvement by reasoning about the relationship between points in 3D and pixels in the image.

\boldparagraph{Label Uncertainty}
Here, we leverage our model's awareness of label uncertainty to demonstrate that higher accuracy can be achieved in confident regions.
To quantify uncertainty, we measure the entropy of the label marginal distribution at every pixel, see \figref{fig:qualitative_results_static} (last row). Sorting all pixels according to their entropy allows us to predict the most certain regions in the image. \tabref{tab:ablation_semantic} (bottom) shows our results on static objects when predicting only those parts of the image. Note how this helps to boost our performance to $94.3\%$ mIoU and $98.4\%$ accuracy when predicting at $70\%$ pixel density, demonstrating that our uncertainty estimates are well calibrated. In contrast, uncertainty is not directly accessible in most baseline models as they are deterministic or rely on MAP estimates. In the benchmarks introduced in \secref{sec:benchmark} where our inferred labels are considered as pseudo-ground truth, we adopt confidence weighted evaluation metrics leveraging the uncertainty to take into account the ambiguity in our automatically generated annotations.

\subsubsection{\bf Instance Segmentation Transfer}
As time consistent 2D instance ground truth is hard to obtain, most existing 2D label transfer methods focus on the semantic segmentation problem. Therefore, we chose to evaluate instance segmentation performance in an ablation study. We annotated the classes ``Building'', ``Car'', ``Trailer'', ``Caravan'' and ``Box'' with instances in our 2D ground truth\footnote{While \cite{Xie2016CVPR} uses two sets of parameters for semantic and instance segmentation, we train a single model for instance segmentation and read semantic labels directly from the instance maps. Therefore, our predictions on classes without instance labels are the same in both semantic and instance segmentation maps.}.
For evaluation, we exploit the mIoU metric defined on instances following \cite{Xie2016CVPR}. Specifically, we first match the ground truth instances to the predicted instances. A pixel is then classified as true positive only when its predicted instance index matches the ground truth.
\tabref{tab:ablation_semantic} (right) shows our results. Note how the instance segmentation results are on par with the semantic segmentation, demonstrating our model's intra-class separation ability. Moreover, we also observe higher instance segmentation accuracy when filtering uncertain predictions.

\subsection{Qualitative Evaluation}
\label{sec:results_qualitative}

\figref{fig:qualitative_results_static} illustrates our dense inference results qualitatively for 3 different scenes in terms of semantic instance segmentation on both static and dynamic objects. The first two rows illustrate both semantic and instance labels, where semantic information is color-coded and instances are separated by boundaries. The last row shows the confidence maps. 
While the proposed method is able to delineate most object boundaries satisfyingly, some challenges remain. Errors occur in regions where 3D points are absent due to far distance (1st \& 3rd scene: far building). Another source of errors is inherent label ambiguities that occur for porous objects such as fences or trees (3rd scene: tree boundary) where even 2D ground truth annotation is a hard and ambiguous task. 
Finally,  3D points of dynamic objects are accumulated over multiple frames (2nd \& 3rd scene), providing dense but less accurate 3D cues to the CRF model. 
However, note that our probabilistic inference algorithm is able to successfully identify those uncertain regions as demonstrated in the last row, where far buildings and object boundaries are predicted as less certain compared to other image regions.

\begin{figure*}[t!]
\setlength{\tabcolsep}{1.0pt}
\centering
\begin{tabular}{cccc}
	\rot{ResNet-50} &
	\includegraphics[width=0.32\linewidth]{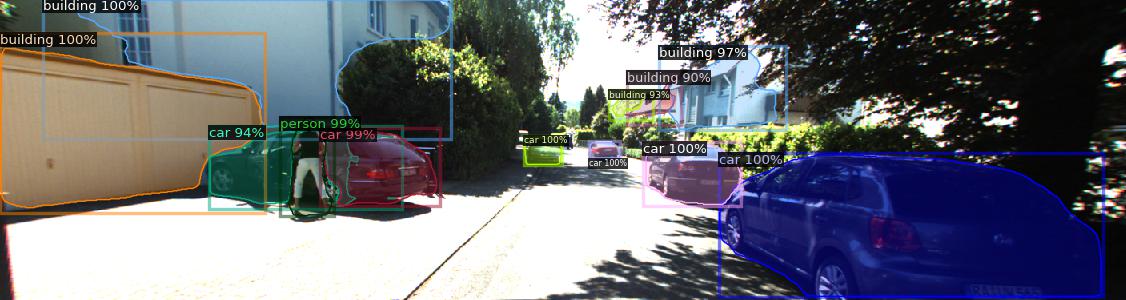} & 
	\includegraphics[width=0.32\linewidth]{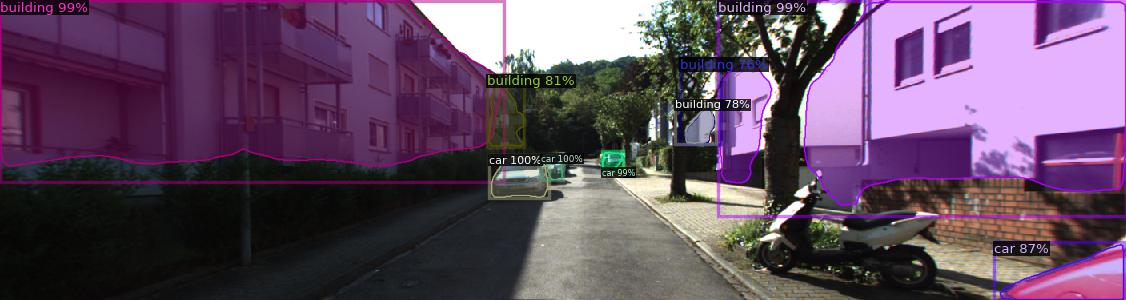} & 
	\includegraphics[width=0.32\linewidth]{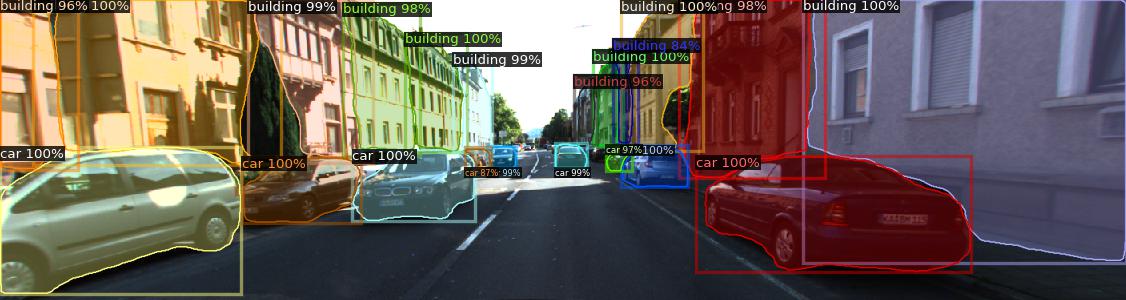} \\
	\rot{ResNet-101} &
	\includegraphics[width=0.32\linewidth]{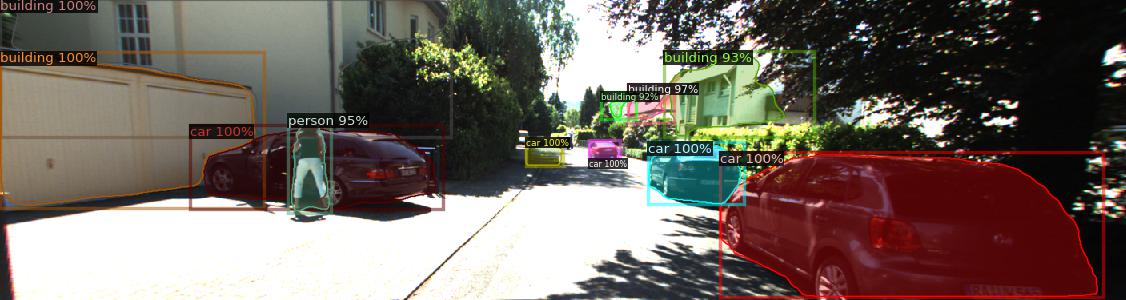} &
	\includegraphics[width=0.32\linewidth]{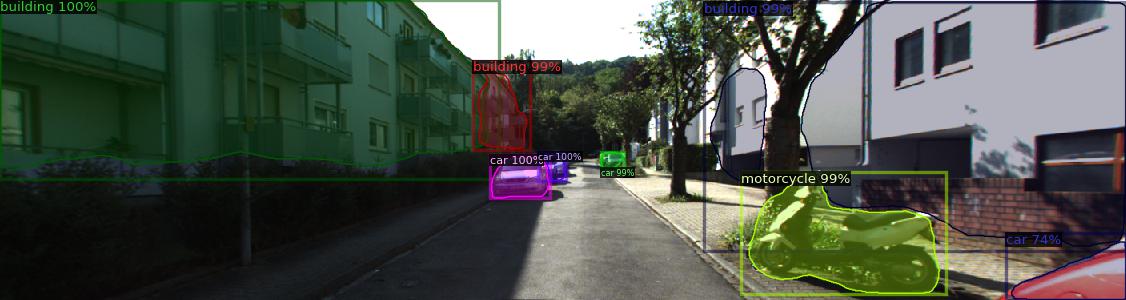} &
	\includegraphics[width=0.32\linewidth]{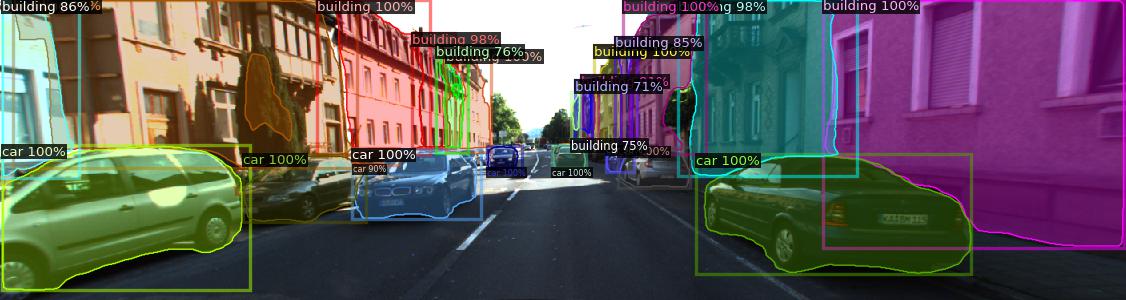} \\
\end{tabular}
\caption{{\bf Qualitative Results for 2D Instance Segmentation.} The first row shows inference results of Mask-RCNN with a ResNet-50 backbone while the second row uses a ResNet-101 backbone. The ResNet-101 backbone leads to better results, \eg, with the ResNet-50 backbone, the car occluded by the person is split into two instances (left) and the motorcycle is not detected (middle). Note that both variants are able to predict “Building” instances after being trained on KITTI-360.
}
\label{fig:benchmark_2d_instance}
\end{figure*}

\begin{table*}[t!]
    \setlength{\tabcolsep}{2pt}
    \begin{subtable}[h]{0.33\textwidth}
		\begin{center}
			
\begin{tabular}{|p{2.0cm}|>{\centering}p{1.5cm}>{\centering}p{1.5cm}|}  \hline
Method & mIoU$_\text{class}$ & mIoU$_\text{category}$  \tabularnewline 
\hline
FCN~\cite{Long2015CVPR} &54.0 & 77.6 \tabularnewline 
PSPNet~\cite{Zhao2017CVPR} & \textbf{64.9} & \textbf{82.2} \tabularnewline 
\hline 
\end{tabular}
		\end{center}
		\vspace{-0.3cm}
		\caption{2D Semantic Segmentation}
		\label{tab:benchmark_2d_semantic}
		\vspace{0.1cm}
	\end{subtable}
    \begin{subtable}[h]{0.33\textwidth}
		\begin{center}

\begin{tabular}{|p{1.3cm}|>{\centering}p{1.2cm}|>{\centering}p{1.15cm}>{\centering}p{1.15cm}|} \hline
Method & Backbone & AP & AP$_{50}$ \tabularnewline 
\hline
Mask R- & Res. 50  & 19.5 & 36.3 \tabularnewline 
CNN~\cite{He2020PAMI}& Res. 101 & \textbf{20.9} & \textbf{40.1} \tabularnewline 
\hline 
\end{tabular}

		\end{center}
		\vspace{-0.3cm}
		\caption{2D Instance Segmentation}
		\vspace{0.1cm}
		\label{tab:benchmark_2d_instance}
	\end{subtable}
    \begin{subtable}[h]{0.33\textwidth}
		\begin{center}
			\begin{tabular}{|p{2.0cm}|>{\centering}p{1.5cm}>{\centering}p{1.5cm}|} \hline
Method & AP$_{50}$ & AP$_{25}$ \tabularnewline 
\hline
BoxNet~\cite{Qi2019ICCV}  & \textbf{4.1} & 23.6  \tabularnewline 
VoteNet~\cite{Qi2019ICCV} & 3.4 & \textbf{30.6} \tabularnewline 
\hline 
\end{tabular}
		\end{center}
		\vspace{-0.3cm}
		\caption{3D Bounding Box Detection}
		\vspace{0.1cm}
		\label{tab:benchmark_3d_bbox}
	\end{subtable}\\
    \begin{subtable}[h]{0.33\textwidth}
		\begin{center}
			\begin{tabular}{|p{2.0cm}|>{\centering}p{1.5cm}>{\centering}p{1.5cm}|}  \hline
Method & mIoU$_\text{class}$ & mIoU$_\text{category}$  \tabularnewline 
\hline
PointNet~\cite{Qi2017CVPR} &13.1 & 30.4 \tabularnewline 
PointNet++~\cite{Qi2017NIPS} & \textbf{35.7} & \textbf{58.3}\tabularnewline 
\hline 
\end{tabular}

		\end{center}
		\vspace{-0.2cm}
		\caption{3D Semantic Segmentation}
		\vspace{0.1cm}
		\label{tab:benchmark_3d_semantic}
	\end{subtable}
    \begin{subtable}[h]{0.33\textwidth}
		\begin{center}

\begin{tabular}{|p{2.7cm}|>{\centering}p{1.15cm}>{\centering}p{1.15cm}|}  \hline
Method &AP & AP$_{50}$  \tabularnewline
\hline
PointNet++~\cite{Qi2017NIPS}+~\cite{Ester1996}&23.7 & 40.1  \tabularnewline
PointGroup~\cite{JiangCVPR2020}  & \textbf{34.8} & \textbf{53.6} \tabularnewline
\hline
\end{tabular}
		\end{center}
		\vspace{-0.2cm}
		\caption{3D Instance Segmentation}
		\vspace{0.1cm}
		\label{tab:benchmark_3d_instance}
	\end{subtable}
    \begin{subtable}[h]{0.33\textwidth}
		\begin{center}
			\begin{tabular}{|p{1.5cm}|>{\centering}p{2.3cm}>{\centering}p{1.2cm}|}
\hline
Method & Acc / Cmp / F$_1$ & mIoU$_\text{class}$ \tabularnewline
\hline
Raw Input & \textbf{98.2} / 19.1 / 32.4 & --  \tabularnewline
Enc-Dec & 41.4 / \textbf{41.2} / \textbf{41.3} & 9.1 \tabularnewline
\hline
\end{tabular}

		\end{center}
		\vspace{-0.2cm}
		\caption{Semantic Scene Completion}
		\vspace{0.1cm}
		\label{tab:benchmark_3d_holistic}
	\end{subtable}
	\vspace{-0.3cm}
	\caption{{\bf Quantitative Results for 2D \& 3D Scene Understanding on Various Different Tasks.}}
\end{table*}

\section{Dataset \& Benchmarks}
\label{sec:benchmark}

We apply the proposed label transfer method to all frames captured by perspective cameras, resulting in 2$\times$78k 2D semantic/instance segmentation maps, 1.0B 3D semantic points and 172.4M 3D instance points. We provide a statistical analysis of the 2D \& 3D labels in \appref{app:statistics}. 
We further deploy an online evaluation server and establish benchmarks on a set of challenging tasks relevant to autonomous driving.
For all tasks, we split the data at the batch-level into disjoint training, validation and held-out test sets as specified in \appref{app:split}. 
Specifically, we leverage KITTI-360 to address tasks at the intersection of vision, graphics and robotics which are commonly viewed as relevant towards achieving full autonomy, including tasks within the scope of semantic scene understanding, novel view synthesis and semantic SLAM. We now describe each task and the corresponding evaluation protocol in detail. Furthermore, we introduce initial baselines for each task.

\subsection{Semantic Scene Understanding}
In this section, we establish scene perception benchmarks in both 2D image space and 3D domain. 
We first implement benchmarks for the traditional tasks of 2D semantic segmentation and 2D instance segmentation on perspective images, using the inferred semantic/instance segmentation maps as pseudo ground-truth. While not the main focus of this work, we establish these standard 2D benchmarks to investigate whether there is a performance gap between methods operating in 2D and 3D.
Furthermore, as our label definition is compatible with Cityscapes, this benchmark opens up the possibility for studying domain adaption across datasets in future work.
Next, we establish benchmarks in the 3D domain, including bounding box detection and semantic/instance segmentation.
Moreover, we consider a semantic scene completion task where the goal is to simultaneously complete the scene and infer corresponding semantic labels given limited observations. This task allows autonomous vehicles to hallucinate future possibilities and thus can benefit downstream tasks, \eg, predictive control.

\boldparagraph{2D Semantic Segmentation} 
We train and evaluate 2D segmentation baselines on the densely labeled images in KITTI-360. We consider two well-known methods, Fully Convolutional Neural Network (FCN)~\cite{Long2015CVPR} and Pyramid Scene Parsing Network (PSPNet)~\cite{Zhao2017CVPR}, as a reference. 
Following Cityscapes~\cite{Cordts2016CVPR}, we adopt mean intersection over union (mIoU) at two semantic granularities, \ie, classes and categories, 
where 19 classes are grouped into 7 coarse-grained categories. 
To account for label uncertainty, the mIoU is weighted by the confidence of our pseudo-ground truth labels. A formal definition of our metrics and a detailed definition of the classes and categories can be found in \appref{app:benchmark_2d_semantic}.
\tabref{tab:benchmark_2d_semantic} shows that, unsurprisingly, PSPNet outperforms the na\"ive FCN on the test set. 

\boldparagraph{2D Instance Segmentation} 
We use the established Mask R-CNN framework~\cite{He2020PAMI} with different backbones as our baselines, see \tabref{tab:benchmark_2d_instance}. We measure the Average Precision (AP) weighted by the label confidence over 10 thresholds, ranging from 0.5 to 0.95 with a step size of 0.05. The mean AP is then calculated over 7 classes that contain instance labels. We also compare mean AP$_{50}$ given a threshold of 0.5. Both \tabref{tab:benchmark_2d_instance} and \figref{fig:benchmark_2d_instance} suggest that Mask R-CNN with a deeper backbone leads to better performance. Note that we provide instance segmentation labels of ``Buildings'' which are not available for other outdoor datasets~\cite{Cordts2016CVPR,Huang2020PAMI,Caesar2020CVPR,Geyer2020ARXIV}. This information allows future works to explore scene compositionality~\cite{Liao2020CVPR,Ost2021CVPR,Niemeyer2021CVPR} in real-world street scenes.

\begin{figure*}[t!]
\centering
\begin{subfigure}{.33\linewidth}
	\includegraphics[width=\linewidth]{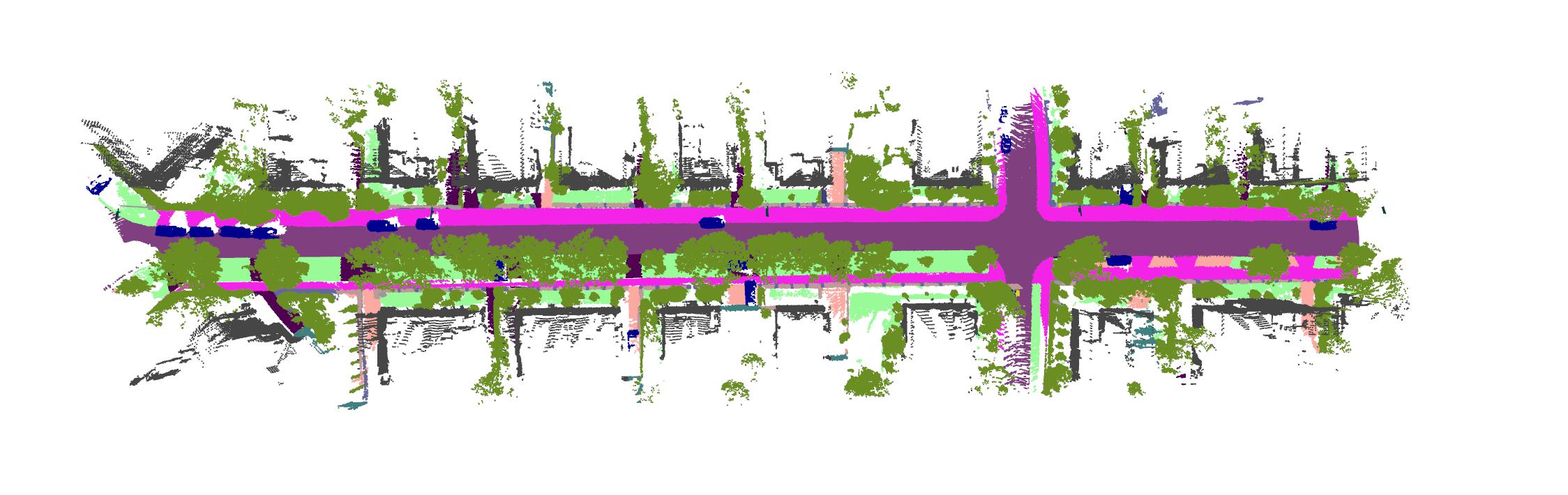} 
	\caption{Semantic GT}
\end{subfigure}
\begin{subfigure}{.33\linewidth}
	\includegraphics[width=\linewidth]{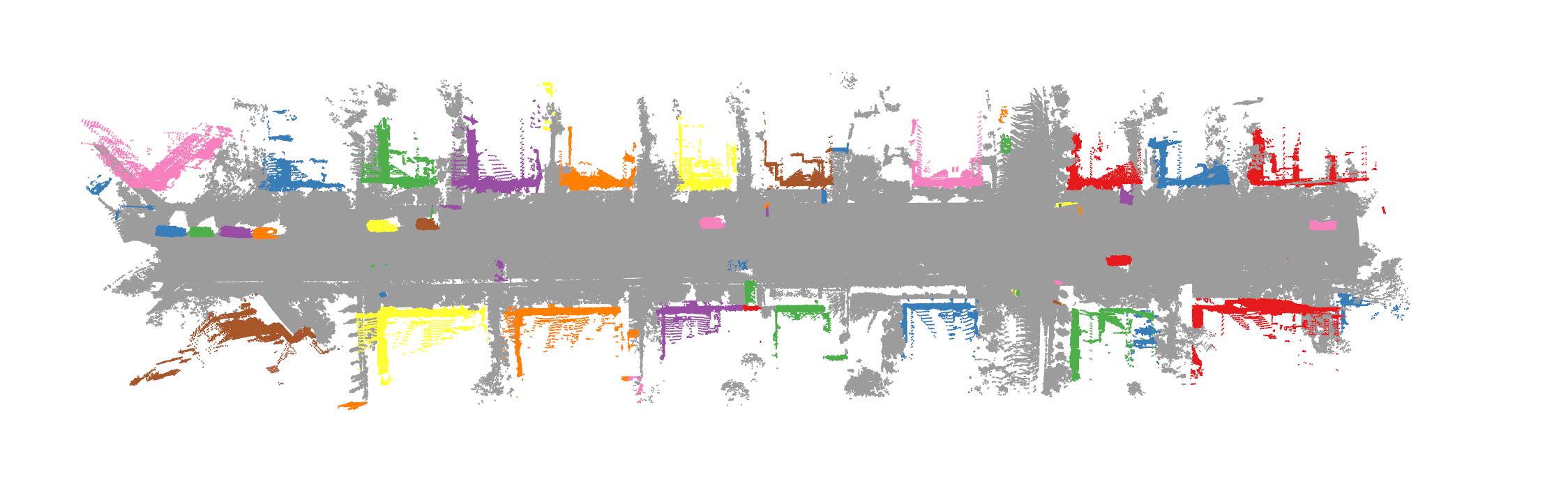} 
	\caption{Instance GT}
\end{subfigure}
\begin{subfigure}{.33\linewidth}
	\includegraphics[width=\linewidth]{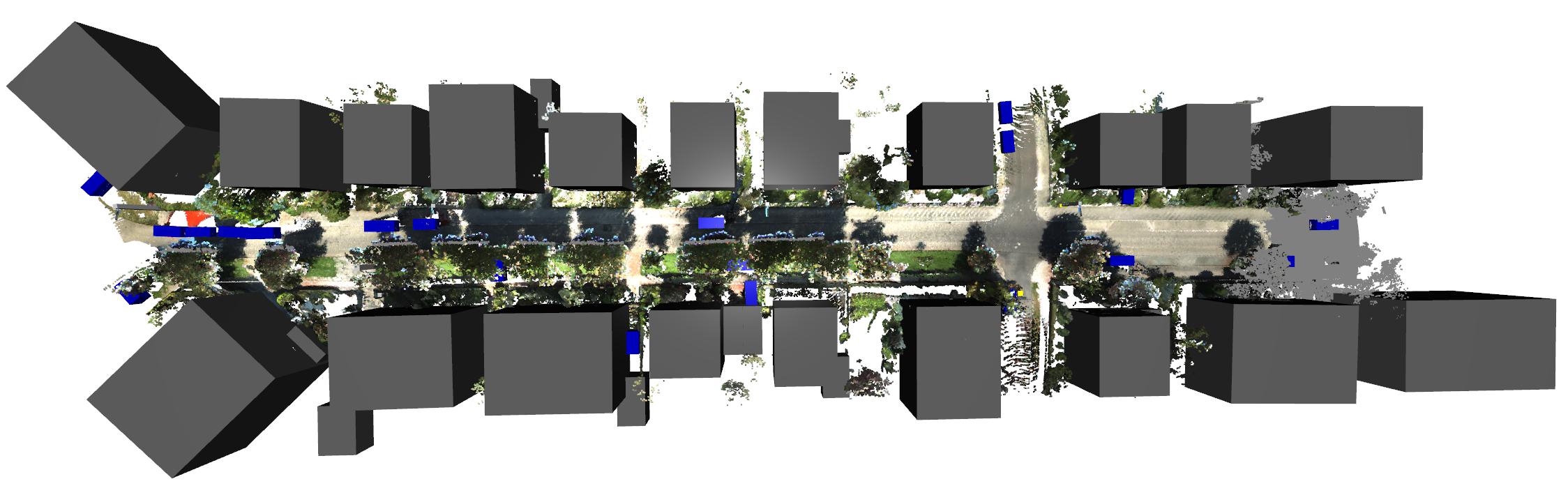}
	\caption{Bbox GT} 
\end{subfigure}\\
\begin{subfigure}{.33\linewidth}
	\includegraphics[width=\linewidth]{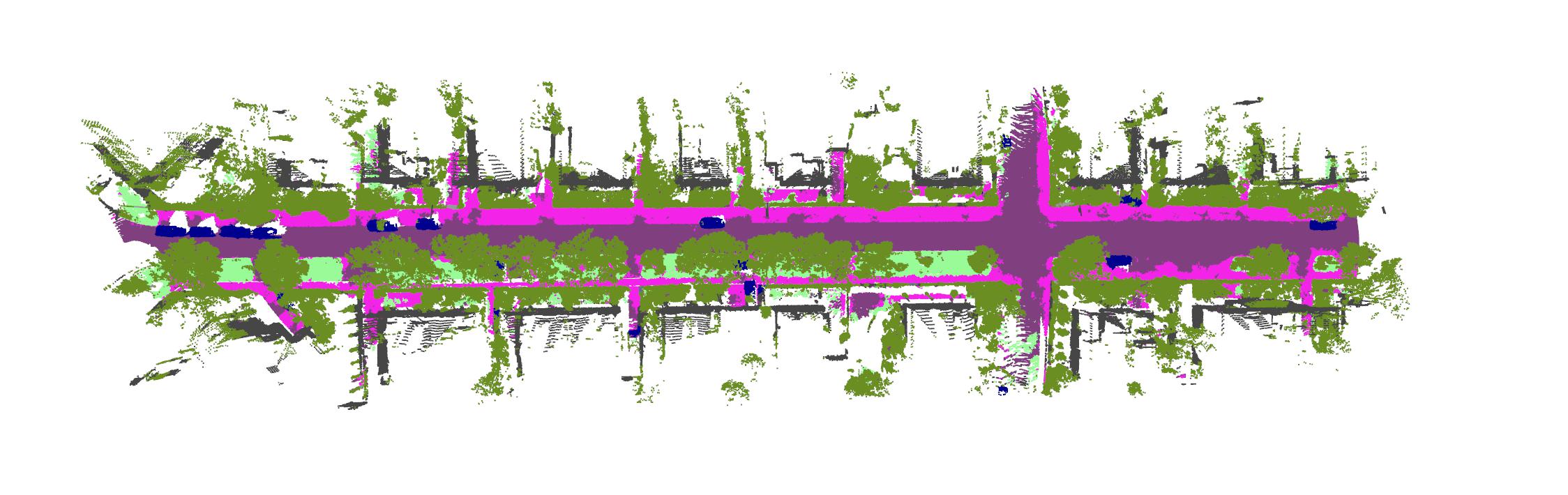} 
	\caption{PointNet++~\cite{Qi2017NIPS}}
	\label{fig:benchmark_3d_semantic}
\end{subfigure}
\begin{subfigure}{.33\linewidth}
	\includegraphics[width=\linewidth]{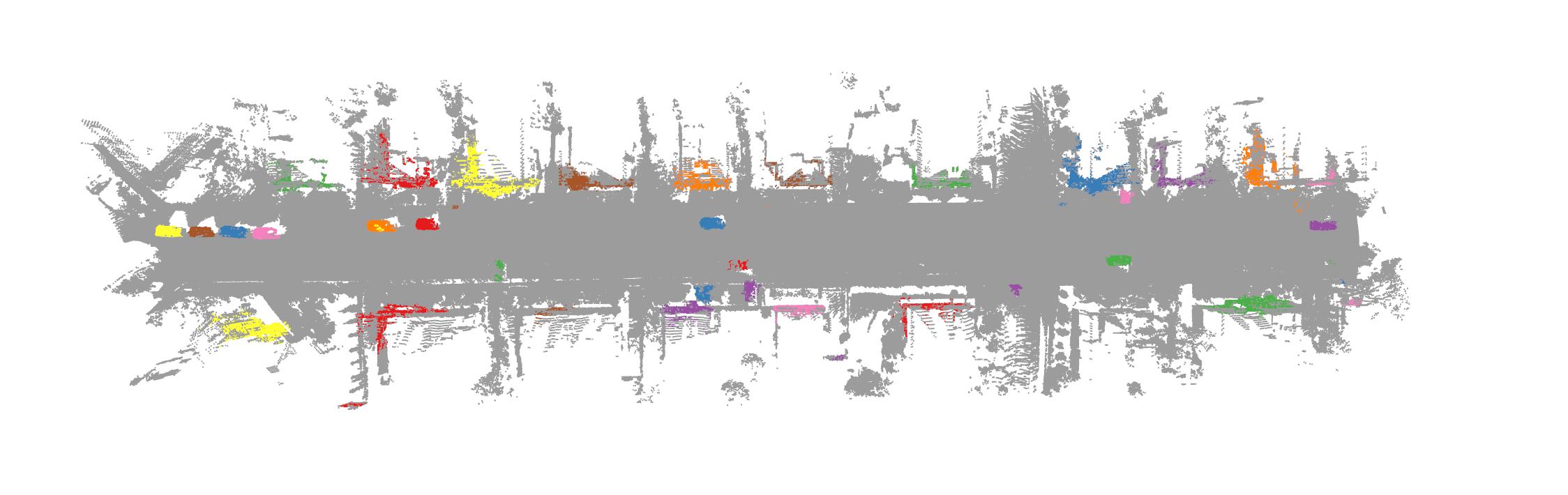} 
	\caption{PointGroup~\cite{JiangCVPR2020}}
	\label{fig:benchmark_3d_instance}
\end{subfigure}
\begin{subfigure}{.33\linewidth}
	\includegraphics[width=\linewidth]{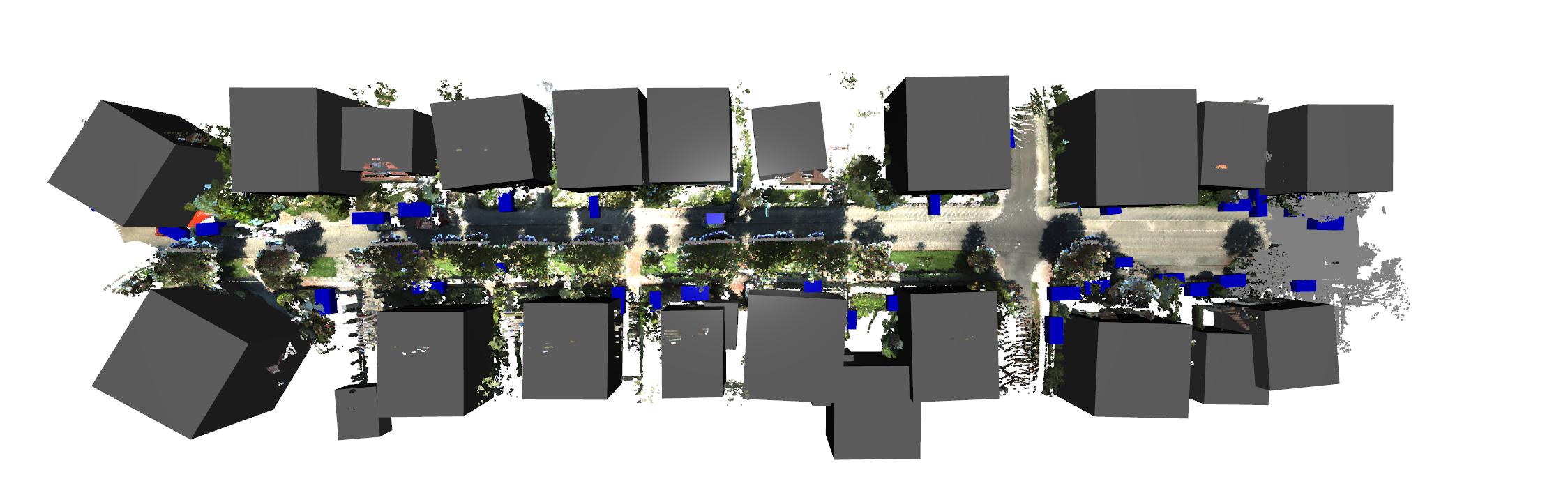}
	\caption{VoteNet~\cite{Qi2019ICCV}}
	\label{fig:benchmark_3d_bbox}
\end{subfigure}
\caption{{\bf Qualitative Results for 3D Scene Perception.} We establish benchmarks for 3D semantic/instance segmentation and 3D bounding box detection. This figure shows the ground truth and the prediction of a baseline method for each sub-task.}
\label{fig:benchmark_3d}
\end{figure*}

\boldparagraph{3D Bounding Box Detection} 
In this benchmark we measure the mean AP over two classes, ``Building'' and ``Car'', since it is particularly challenging for learning-based algorithms to generalize well to other classes with fewer training samples. Following~\cite{Qi2019ICCV}, the mean AP is calculated at a threshold of 0.25 and 0.5, respectively.
We consider VoteNet~\cite{Qi2019ICCV} and its simplified version BoxNet~\cite{Qi2019ICCV} as baseline methods. Both methods require 3D point locations as input and output 3D bounding boxes and their semantic labels. 
\tabref{tab:benchmark_3d_bbox} suggests that VoteNet can make reasonable predictions for both building and cars while it fails to predict 3D bounding boxes with high IoU values, see \figref{fig:benchmark_3d_bbox}. 

\boldparagraph{3D Semantic Segmentation} 
We establish a 3D semantic segmentation benchmark on the accumulated point clouds, where PointNet~\cite{Qi2017CVPR} and PointNet++~\cite{Qi2017NIPS} are trained and evaluated as baselines. Both methods take as input point locations and colors to predict a semantic label for each 3D point. Following the 2D semantic segmentation task, we measure mIoU weighted by label confidence over classes and categories, respectively. \tabref{tab:benchmark_3d_semantic} shows the quantitative comparison and \figref{fig:benchmark_3d_semantic} illustrates the performance of PointNet++. Interestingly, comparing \tabref{tab:benchmark_2d_semantic} and  \tabref{tab:benchmark_3d_semantic} shows that the 3D semantic segmentation baselines’ overall performances are inferior compared to the 2D semantic segmentation methods, suggesting that parsing the semantic meaning of  irregularly structured 3D point clouds remains more challenging and requires further work.

\begin{figure*}[t!]
\centering
\setlength{\tabcolsep}{0.0pt}
\centering
\begin{tabular}{cc|cc|cc}
	\includegraphics[width=0.16\linewidth]{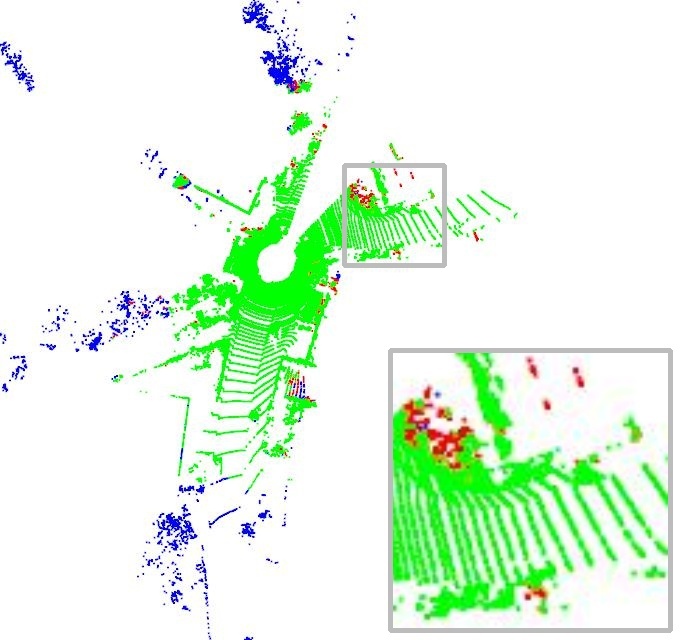} & 
	\includegraphics[width=0.16\linewidth]{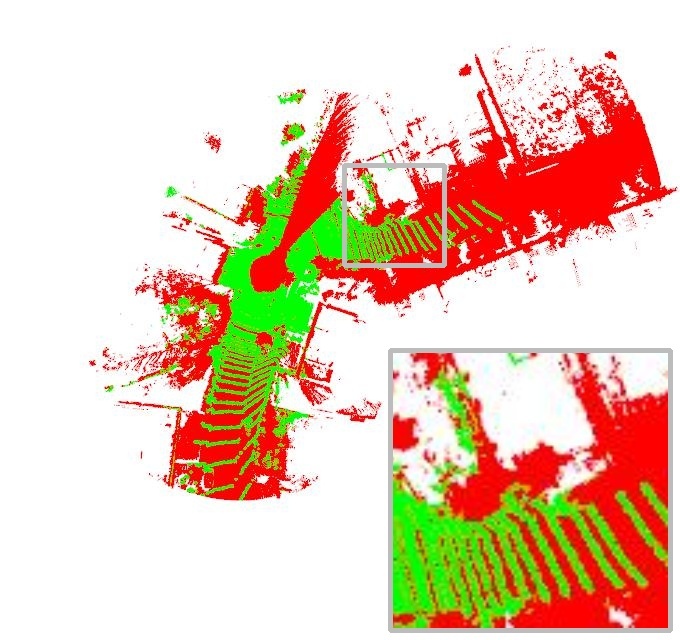} &
	\includegraphics[width=0.16\linewidth]{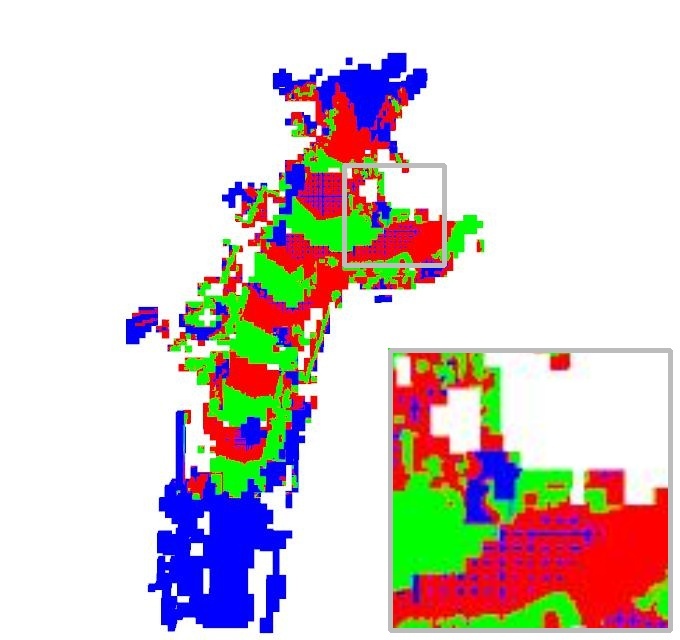} & 
	\includegraphics[width=0.16\linewidth]{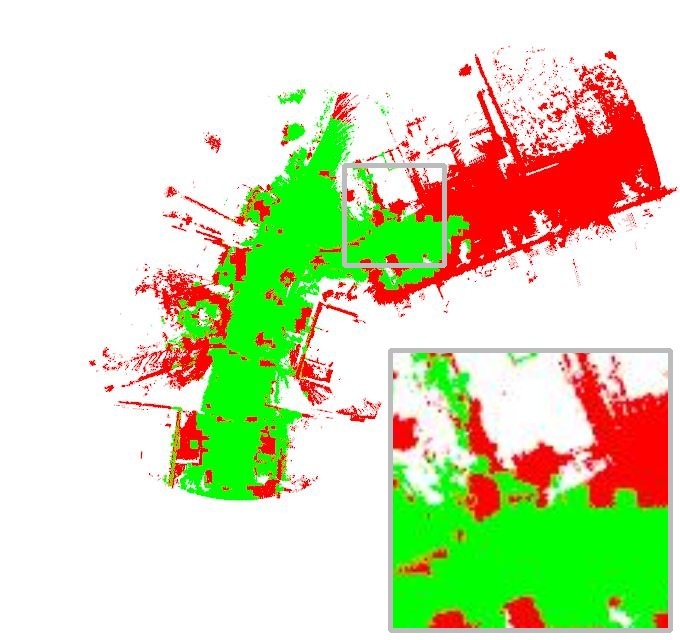} &
	\includegraphics[width=0.16\linewidth]{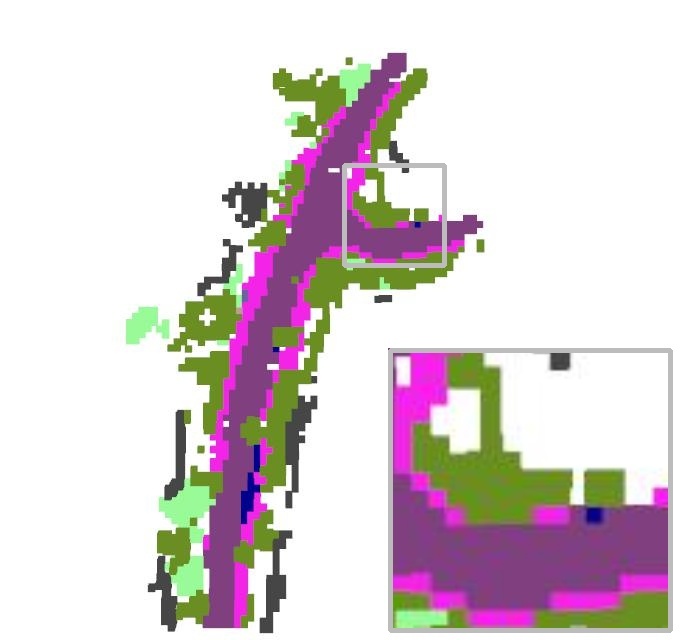} &
	\includegraphics[width=0.16\linewidth]{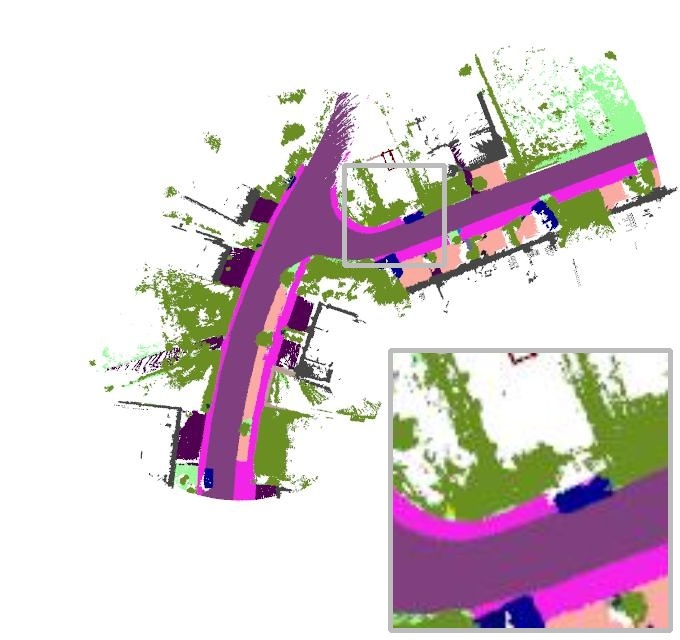} \\
	Raw Input Acc. & Raw Input Comp. & Enc-Dec Acc. & Enc-Dec Comp. & Enc-Dec Semantic & GT Semantic \\
\end{tabular}
\caption{{\bf Qualitative Results for Semantic Scene Completion} evaluated at a distance threshold of 20cm. Green denotes complete/accurate, red denotes incomplete/inaccurate and blue denotes points in unobserved region.}
\label{fig:benchmark_3d_holistic}
\end{figure*}

\boldparagraph{3D Instance Segmentation} 
We evaluate 3D instance segmentation results for ``Building'' and ``Car''.  
Specifically, we measure the mean AP over a set of thresholds ranging from 0.5 to 0.95 with a step size of 0.05 and AP at a threshold of 0.25 and 0.5.
As a first simple baseline, we na\"ively cluster semantically labeled points into instances. We use PointNet++~\cite{Qi2017NIPS} for semantic segmentation and DBSCAN~\cite{Ester1996} for clustering.
We further evaluate PointGroup~\cite{JiangCVPR2020} as a state-of-the-art method for 3D instance segmentation which takes as input point locations and colors. \tabref{tab:benchmark_3d_instance} demonstrates that PointGroup outperforms the na\"ive clustering-based method. The qualitative result of PointGroup is shown in \figref{fig:benchmark_3d_instance}. 
While the 2D and 3D results in \tabref{tab:benchmark_2d_instance} and \tabref{tab:benchmark_3d_instance} are defined over different sets of classes, we provide detailed results on each class in \appref{app:benchmark_3d_instance}.
 Interestingly, 3D instance segmentation methods achieve better performance on ``Car'' than 2D methods while performing worse on ``Building''.
We hypothesize that unlike in 2D where occlusions strongly impact the results (\eg \figref{fig:benchmark_2d_instance} left, pedestrian standing in front of a car), cars can be more easily separated in 3D.
As for buildings, many instances are spatially connected (\eg, \figref{fig:benchmark_2d_instance} right), making the instance segmentation task harder in 3D where boundaries are harder to detect on the sparse point cloud.

\boldparagraph{Semantic Scene Completion}
While standard scene perception tasks aim to predict a semantic label for each observed scene point, the semantic scene completion task additionally requires predicting geometry and semantics in unobserved regions. Given a single LiDAR scan as input, this task requires semantic scene completion within a corridor of 30m around the vehicle poses of a 100m trajectory.
For evaluation, we measure reconstruction quality and semantic prediction accuracy. The former measures geometric accuracy independent of semantics, using \textit{completeness} and \textit{accuracy} over a range of distance thresholds following common practice~\cite{Schops2017CVPR}. We consider a threshold of 20cm as the main metric.
As our ground truth reconstruction may not be complete, we evaluate accuracy only in observed regions. We further measure the F$_1$ score as the harmonic mean of the completeness and the accuracy. The semantic prediction quality is conditioned on the geometric reconstruction. Specifically, we measure the confidence weighted \textit{mIoU} over the same set of thresholds where a true positive prediction is made when 1) a ground truth point is classified as complete at the given threshold and 2) its closest reconstructed point has the correct label. See \appref{app:benchmark_scene_complt} for more details of the ground truth construction and evaluation metrics.

We consider two baselines for this task, both taking a single raw LiDAR frame as input. For calibration, we implement a na\"ive baseline which returns the input as output. The second baseline is a learning-based approach where we use an encoder-decoder architecture to predict the complete scene structure from the raw LiDAR scan. More details about this baseline can be found in \appref{app:benchmark_scene_complt}. \tabref{tab:benchmark_3d_holistic} and \figref{fig:benchmark_3d_holistic} illustrate the results. As expected, the raw LiDAR scans are accurate but incomplete. The learning-based approach instead achieves higher completeness but the predictions are less accurate. For the learning-based approach we also predict a semantic label at each 3D point. \figref{fig:benchmark_3d_holistic} shows that the model is able to correctly predict semantic labels at a coarse level but struggles to predict smaller objects like cars.

\boldparagraph{Discussion}
Our results show that 3D semantic segmentation is harder than 2D semantic segmentation.
In contrast, our conclusions for instance segmentation vary for different classes. Some classes, \eg, cars, are easier to segment in 3D, suggesting further works can explore 3D information to enhance 2D instance segmentation. 3D bounding box detection remains challenging, especially when a high IoU is desired. Lastly, while inferring dense geometry and semantics from raw sparse observations can benefit autonomous driving, completing the scene and predicting semantics jointly is a difficult task that requires further research.

\begin{figure*}[t!]
\centering
\begin{subfigure}{.33\linewidth}
	\includegraphics[width=\linewidth]{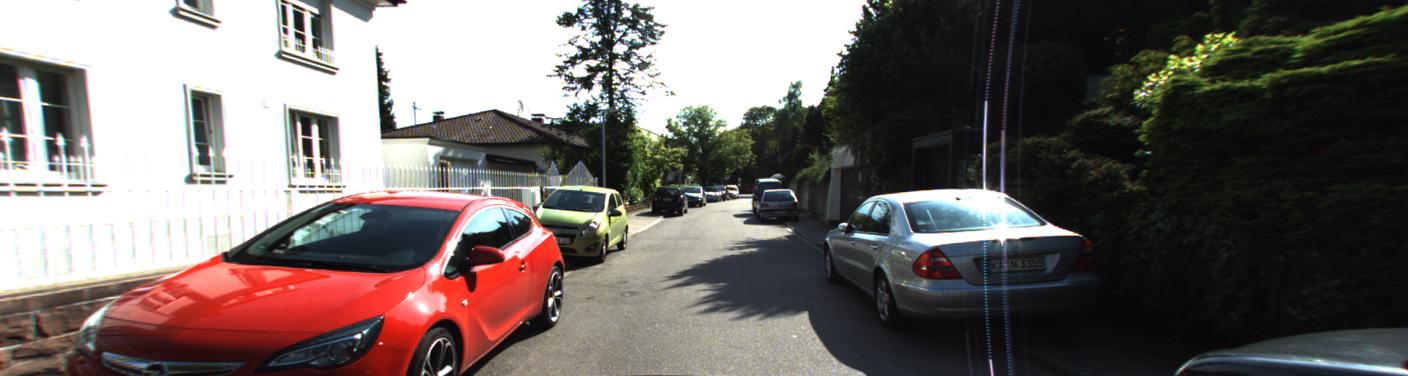} 
	\caption{GT Image}
\end{subfigure}%
\begin{subfigure}{.33\linewidth}
	\includegraphics[width=\linewidth]{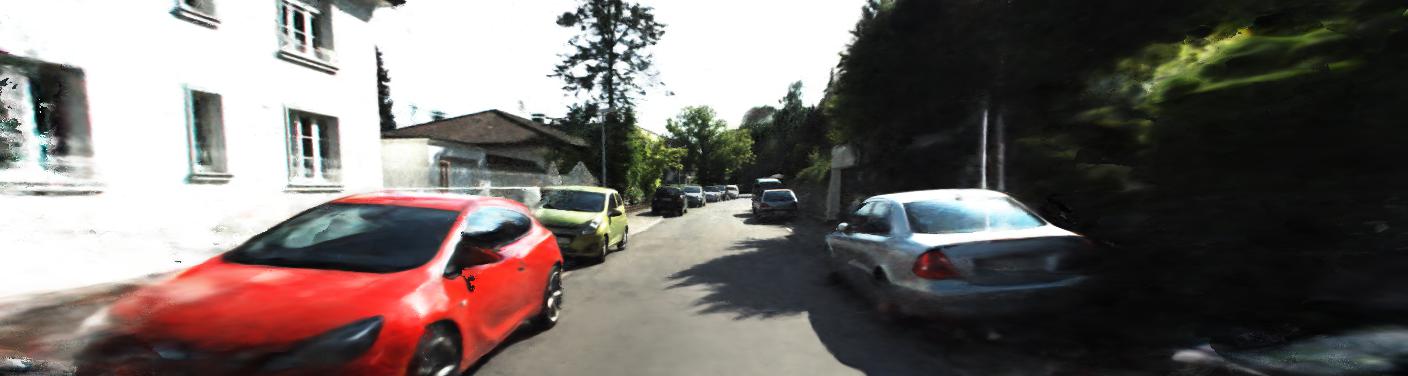} 
	\caption{NeRF~\cite{Mildenhall2020ECCV}}
\end{subfigure}%
\begin{subfigure}{.33\linewidth}
	\includegraphics[width=\linewidth]{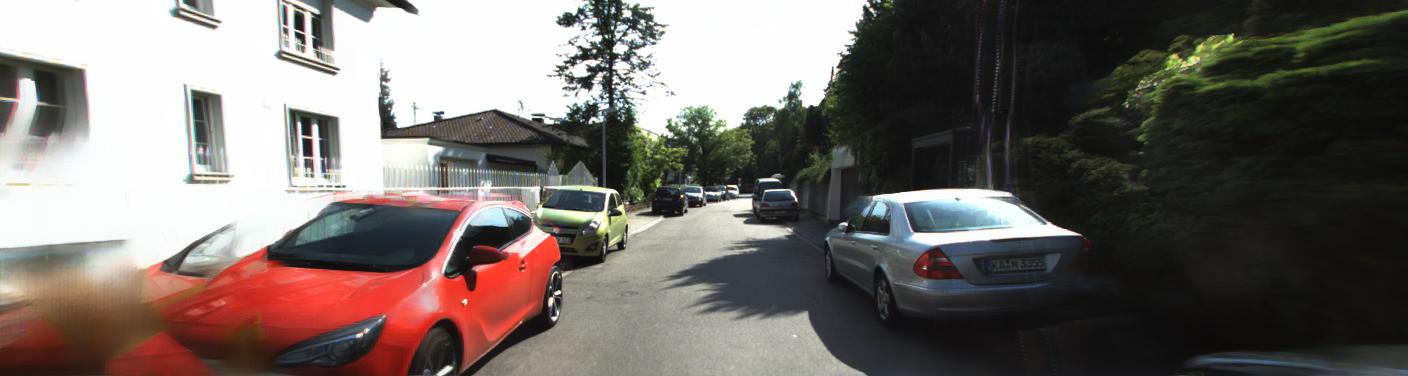} 
	\caption{FVS~\cite{Riegler2020ECCV}}
\end{subfigure}\vspace{0.1cm}\\%
\begin{subfigure}{.33\linewidth}
	\includegraphics[width=\linewidth]{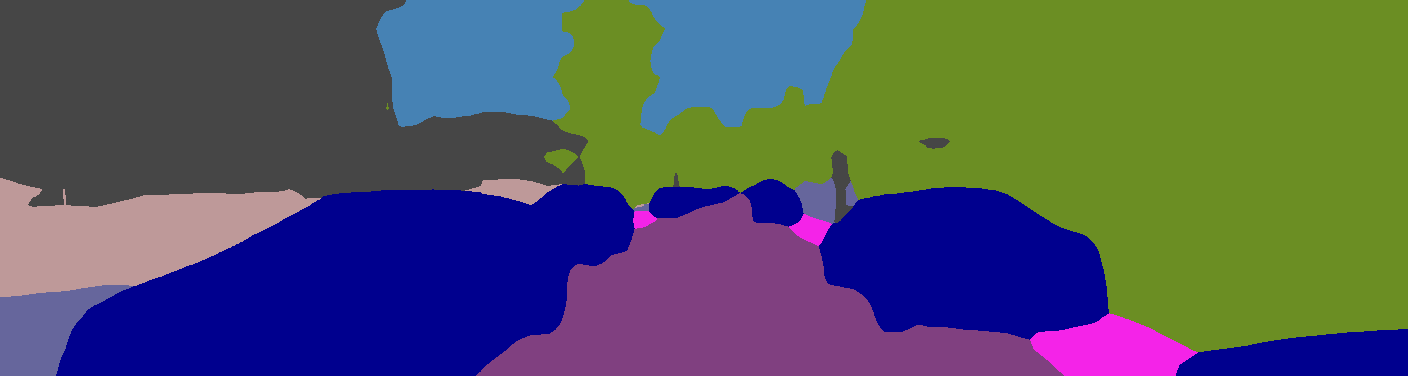}
	\caption{GT Image + PSPNet~\cite{Zhao2017CVPR}}
\end{subfigure}%
\begin{subfigure}{.33\linewidth}
	\includegraphics[width=\linewidth]{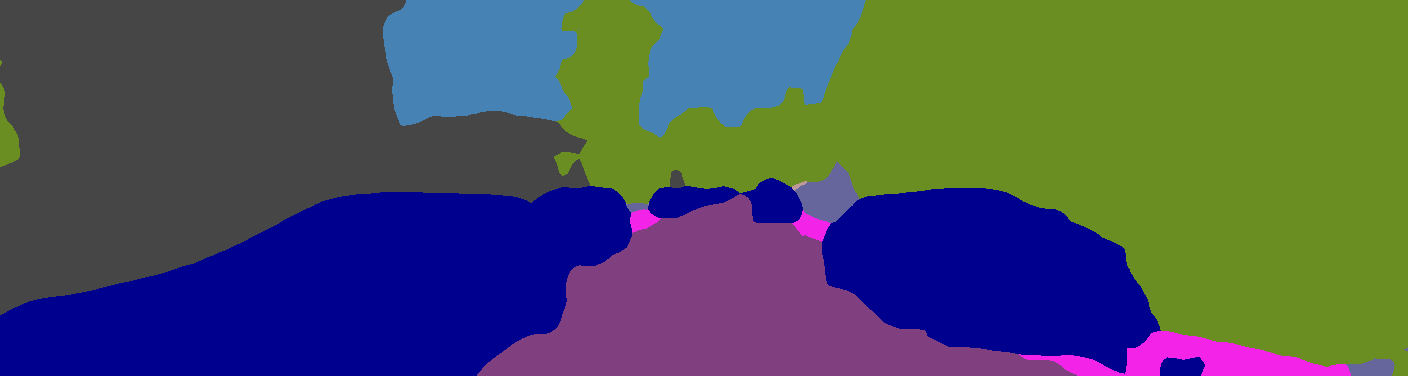}
	\caption{NeRF~\cite{Mildenhall2020ECCV} + PSPNet~\cite{Zhao2017CVPR}}
\end{subfigure}%
\begin{subfigure}{.33\linewidth}
	\includegraphics[width=\linewidth]{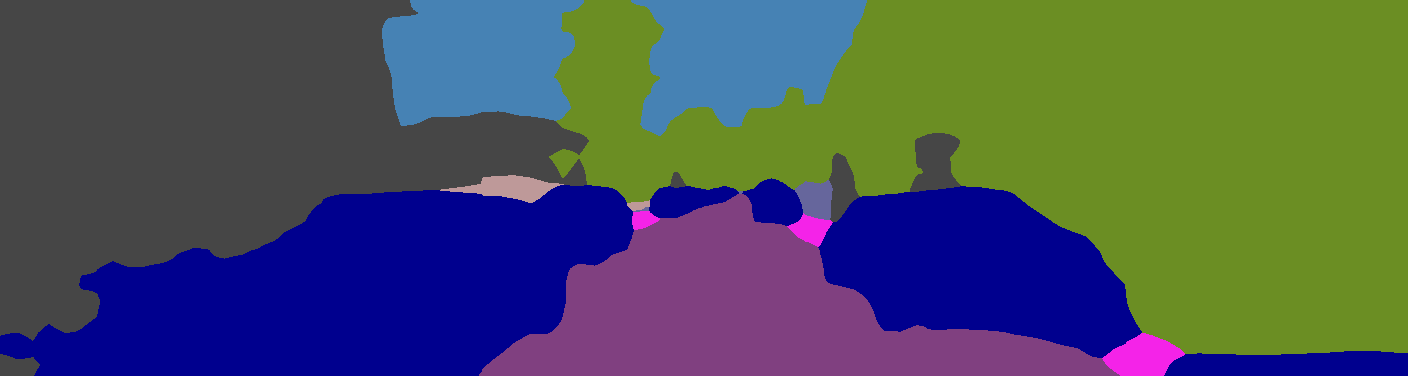} 
	\caption{FVS~\cite{Riegler2020ECCV} + PSPNet~\cite{Zhao2017CVPR}}
\end{subfigure}%
\caption{{\bf Qualitative Results for Novel View Appearance \& Semantic Synthesis.} The first row shows the GT image and novel view appearance synthesis results. The second row shows the corresponding semantic segmentation using PSPNet~\cite{Zhao2017CVPR}.}
\label{fig:benchmark_nvs_label}
\end{figure*}

\subsection{Novel View Synthesis}
Simulation is an essential tool for training and evaluating autonomous vehicles. While existing methods trained in simulated scenes struggle to generalize to real scenes, creating a simulation environment based on real-world images is a promising direction to close the gap between real-world scenarios and synthetic environments~\cite{Yang2020CVPR,Chen2021CVPR}. We thus establish challenging benchmarks towards this goal, including novel view appearance synthesis and novel view semantic synthesis.

\boldparagraph{Novel View Appearance Synthesis} 
In this benchmark, we are interested in novel view RGB image synthesis for driving scenarios. While we evaluate on a set of held-out perspective images, the benchmark participant can choose from a set of input modalities\footnote{The used input modalities will be indicated on the leaderboard.}, including posed perspective/fisheye images or accumulated point clouds. 
For perspective and fisheye images, we release approximately $50\%$ of the frames for training and use the remaining $50\%$ for testing. In addition, the evaluation server also provides a harder setting with a $90\%$ drop rate. See appendix for details.
The point cloud is accumulated over all frames where each point fuses colors from different viewpoints.
We adopt three standard evaluation metrics for this benchmark: peak signal-to-noise ratio (PSNR), structural similarity index (SSIM), and perceptual metric (LPIPS)~\cite{Zhang2018CVPR2}.

We evaluate two sets of baselines on two different input modalities.
We first consider a na\"ive baseline using the accumulated point cloud (PCL) as input. Specifically, 
we project non-occluded colored points to test viewpoints, followed by nearest neighbor interpolation to fill in the missing values. As there are no 3D points in the sky, we heuristically assign a mean blue color to the sky region. 
We further consider several state-of-the-art baselines for image-based novel view synthesis, including methods based on Neural Radiance Fields~\cite{Mildenhall2020ECCV,Barron2021ICCV,Deng2021ARXIV} or per-view depth maps~\cite{Riegler2020ECCV,Kopanas2021CGF}.
Results in \tabref{tab:benchmark_nvs} (left) and \figref{fig:benchmark_nvs_label} (1st row) reveal the challenges of this benchmark. 
We observe that NeRF~\cite{Mildenhall2020ECCV} shows promising results but struggles to synthesize fine structure. FVS performs better in rendering fine details (\eg, license plate) but exhibits noticeable artifacts due to the inaccurate underlying geometry (\eg, left car). Interestingly, FVS/PBNR performs better in LPIPS but has a lower PSNR compared to NeRF-based methods, suggesting that LPIPS is more sensitive to fine detail than larger regional errors.

\begin{table}[t!]
\setlength{\tabcolsep}{2.65pt}
\begin{center}
\begin{tabular}{|p{1.80cm}|>{\centering}p{0.60cm}|>{\centering}p{0.85cm}>{\centering}p{0.85cm}>{\centering}p{0.85cm}|>{\centering}p{1.00cm}>{\centering}p{1.40cm}|}  \hline
Method & Input & PSNR  & SSIM & LPIPS & mIoU$_\text{class}$ & mIoU$_\text{category}$  \tabularnewline 
\hline
GT Image & -- & -- & -- & -- & \textbf{72.0} & \textbf{83.6}  \tabularnewline
PCL & PC  & 12.81 & 0.576 & 0.549 & 39.4 & 46.8 \tabularnewline
NeRF~\cite{Mildenhall2020ECCV} & PI & 21.18 & 0.779 & 0.343 & 53.0 & 73.9 \tabularnewline
mip-NeRF~\cite{Barron2021ICCV} & PI  & \textbf{21.54} & 0.778 & 0.365 & 51.2 & 72.0 \tabularnewline
DS-NeRF~\cite{Deng2021ARXIV} & PI  & 21.28	& 0.777	& 0.347 & 54.8 & 75.5 \tabularnewline
FVS~\cite{Riegler2020ECCV} & PI  & 20.00 & 0.790 & \textbf{0.193} & 67.1 & 78.5 \tabularnewline
PBNR~\cite{Kopanas2021CGF} & PI  & 19.91 & \textbf{0.811} & \textbf{0.191}  & 65.1 & 77.8 \tabularnewline
\hline
\end{tabular}
\end{center}
\vspace{-0.3cm}
\caption{{\bf Quantitative Results for Novel View Appearance \& Novel View Semantic Synthesis using a $50\%$ drop rate.} Input: ``PC'' denotes the accumulated point cloud and ``PI'' means perspective images. }
\label{tab:benchmark_nvs}
\end{table}

\begin{figure*}[t!]
\centering
\begin{subfigure}{.33\linewidth}
	\includegraphics[width=\linewidth]{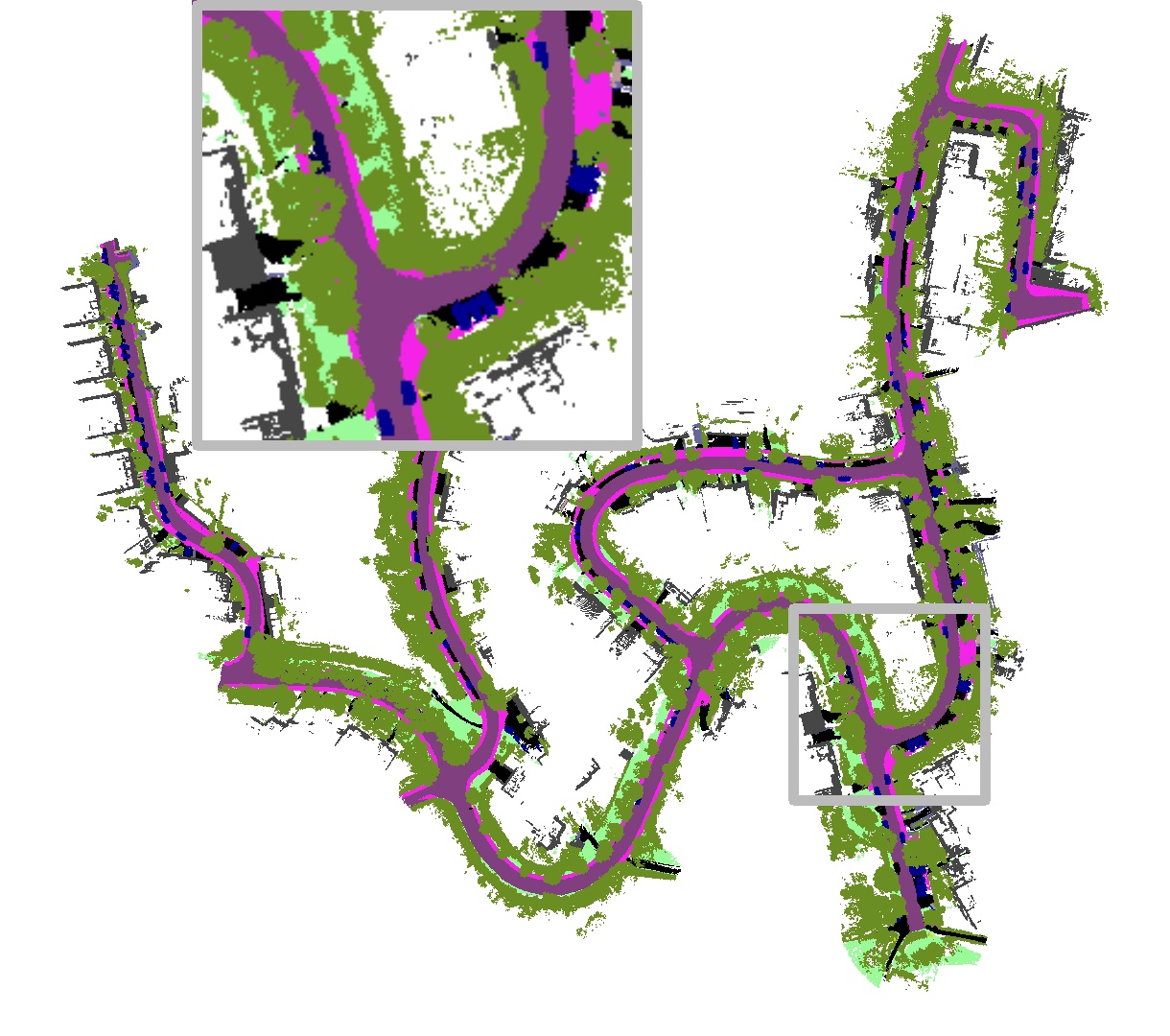} 
	\caption{GT Semantic Reconstruction}
\end{subfigure}
\begin{subfigure}{.33\linewidth}
	\includegraphics[width=\linewidth]{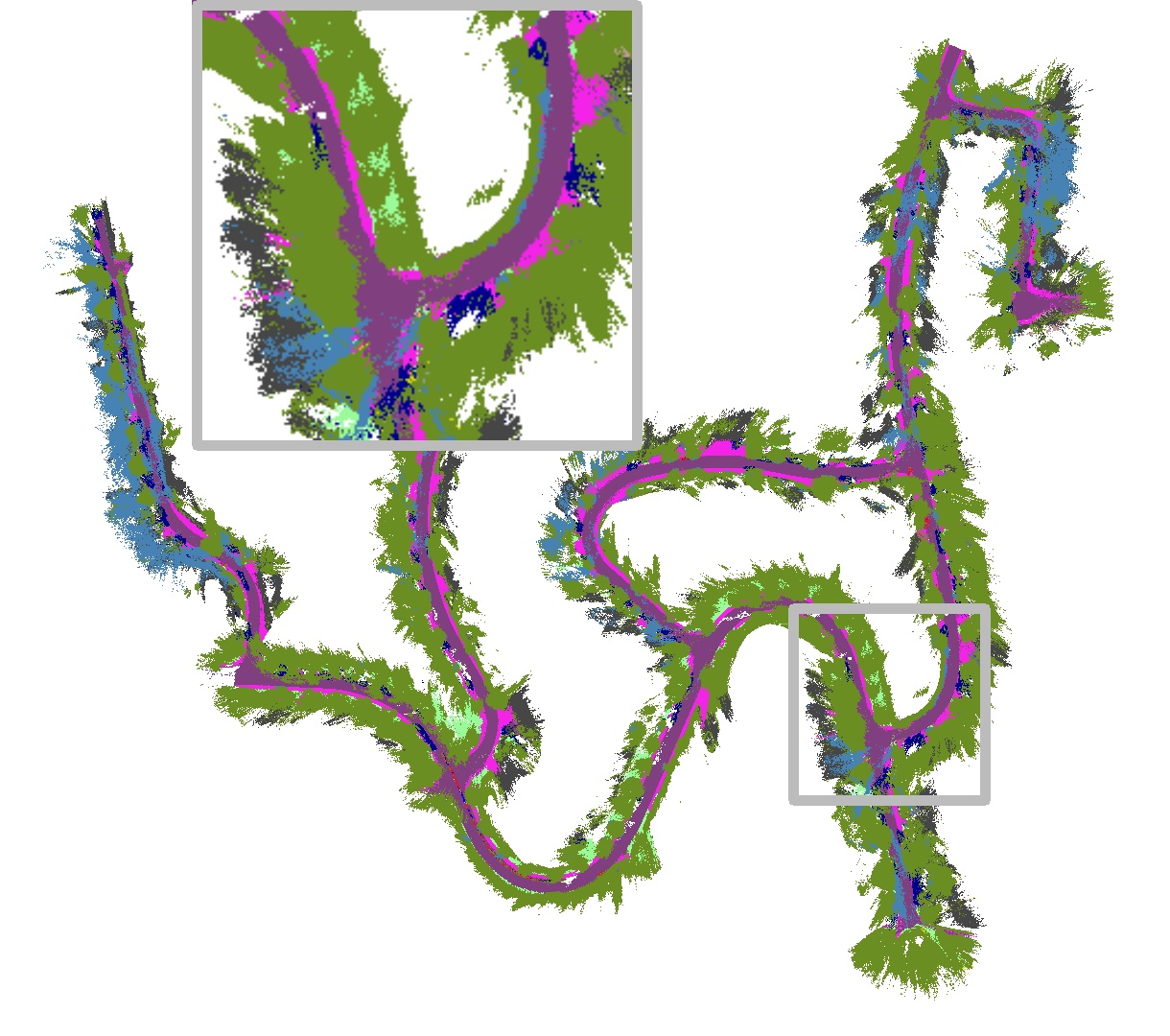} 
	\caption{ORB-SLAM2 + SGM}
\end{subfigure}
\begin{subfigure}{.33\linewidth}
	\includegraphics[width=\linewidth]{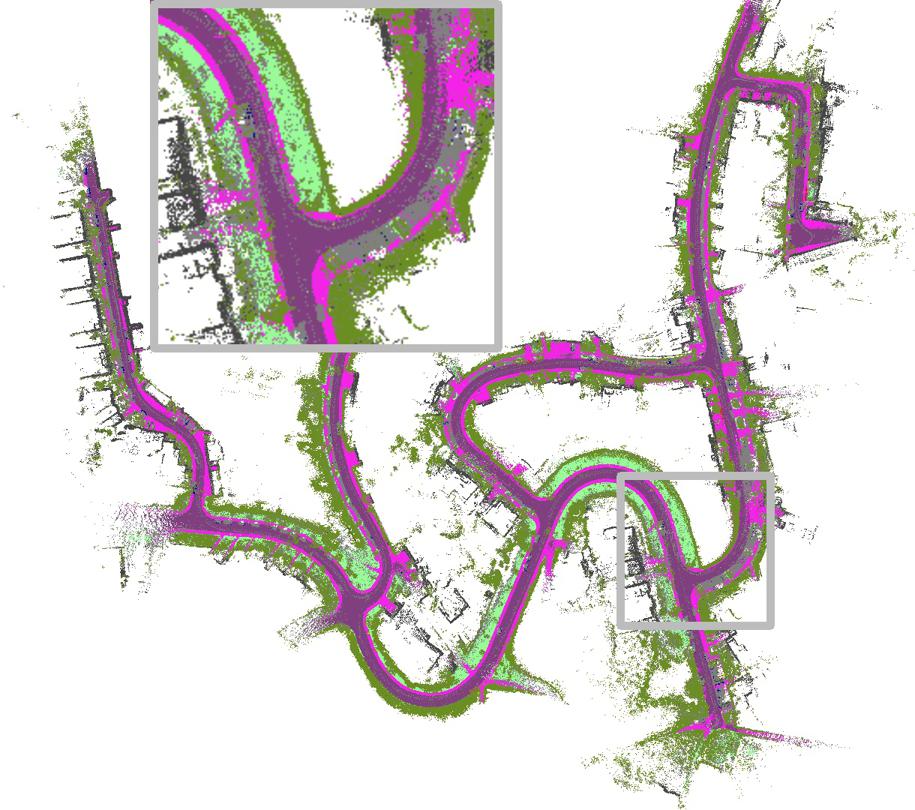} 
	\caption{SUMA++}
\end{subfigure}
\caption{{\bf Qualitative Results for Semantic Mapping} on test sequence 3 colored based on semantic class labels. }
\label{fig:benchmark_slam_semantic}
\end{figure*}

\boldparagraph{Novel View Semantic Synthesis}
An important property of simulation environments like CARLA \cite{Dosovitskiy2017CORL} is that they provide not only RGB images but also auxiliary information like semantic label maps. Towards a real-world simulator with the same capability, we therefore 
consider a novel benchmark that requires joint novel view and semantic synthesis. The input data for this task is the same as for the novel view synthesis task, while the methods are tasked to predict both an RGB image and a semantic segmentation map at a given target camera pose. Therefore, the evaluation metric of this task additionally comprises mIoU for semantic segmentation as shown in \tabref{tab:benchmark_nvs} (right). As no prior work has addressed this problem yet, we consider a na\"ive two-stage solution as baseline to bootstrap this benchmark, \ie, we apply an existing semantic segmentation model (PSPNet~\cite{Zhao2017CVPR}) on the synthesized images. For comparison, we also evaluate the semantic segmentation performance on the original ground truth images (GT Image). Note that the artifacts in the synthesized images lead to a significant performance drop for semantic segmentation. As illustrated in \figref{fig:benchmark_nvs_label}, the fence is misclassified as building when the synthesized images are taken as input to PSPNet, despite that the fence is still visible in these images. It is also interesting to note that semantic segmentation performance is aligned with the LPIPS metric, as both apply pre-trained networks on synthesized images.

\boldparagraph{Discussion}
Our baselines reveal the different challenges in novel view appearance synthesis with different input modalities.  While point clouds provide a good representation of 3D geometry, it is not easy to model view dependency. When instead taking a sparse set of multi-view images as input, the task is similarly difficult despite little variation in the camera orientation. We believe that future works should explore the combination of different input modalities to improve image fidelity further. Moreover, 
given the low performance of our simple baselines on the novel view semantic synthesis task, there is a large potential for future improvements, \ie, by learning view and semantic synthesis jointly.

\subsection{Semantic SLAM}
We further establish a semantic SLAM benchmark at the intersection of robotics and computer vision.
Here, the goal is to simultaneously estimate poses and reconstruct a \textit{semantic} map from monocular/stereo images and/or LiDAR scans. 
While there is a growing interest in evaluating indoor semantic reconstructions of SLAM algorithms at room-level~\cite{Wald2018RAL,Rosinol2020ICRA}, existing works on outdoor semantic SLAM typically evaluate only pose estimation while ignoring the quality of the semantic reconstruction~\cite{Brasch2018ICRA,Chen2019IROS}. Considering that the semantic reconstruction is valuable on its own for down-stream tasks, \eg planning ~\cite{Ding2019RAL}, we thus additionally evaluate geometric and semantic reconstructions where the latter is enabled by the dense semantic annotations of KITTI-360. 
For this benchmark, the test sequences are separated from those used for 3D scene perception such that the accumulated point cloud is held out from the public.

\boldparagraph{Localization}
Given an estimated trajectory, we adopt the standard Absolute Pose Error (APE) and Relative Pose Error (RPE)~\cite{Grupp2017EVO} as metrics for evaluating pose estimation. We consider four test sequences for this task and report the evaluation results on each test sequence without averaging.

We evaluate two baseline methods, ORB-SLAM2~\cite{Artal2017TR} and SUMA++~\cite{Chen2019IROS}, where the former takes stereo images as input and the latter is applied on LiDAR scans. \tabref{tab:benchmark_slam_pose} compares the localization results of ORB-SLAM2 and SUMA++. 
For both methods, the APE exceeds 2 meters and the RPE is around 2\% in general.
ORB-SLAM2 achieves better overall performance compared to SUMA++, suggesting that the stereo images of our dataset contain rich features for the purpose of localization. One possibility for improving localization accuracy is to exploit the 3D bounding boxes and instance labels available in our dataset \cite{Yang2019TR,Nicholson2019RAL}.

\begin{table}[t!]
	\begin{subtable}{\linewidth}
		\setlength{\tabcolsep}{2.65pt}
		\begin{center}
		\begin{tabular}{|>{\centering}p{0.5cm}|>{\centering}p{1.6cm}>{\centering}p{1.6cm}|>{\centering}p{1.6cm}>{\centering}p{1.6cm}|}
\hline
Test & \multicolumn{2}{c|}{ORB-SLAM2} & \multicolumn{2}{c|}{SUMA++}\tabularnewline
Seq. & APE (m) & RPE (\%) & APE (m) & RPE (\%) \tabularnewline
\hline
0 & \textbf{1.53} $\pm$ 0.74 & \textbf{2.42} $\pm$ 1.34 & 2.27 $\pm$ 1.38 & 2.66 $\pm$ 2.12 \tabularnewline
1 & \textbf{2.22} $\pm$ 0.78 & 2.46 $\pm$ 1.37 & 2.87 $\pm$ 1.50 & \textbf{2.43} $\pm$ 1.80 \tabularnewline
2 & \textbf{2.12} $\pm$ 0.94 & \textbf{1.50} $\pm$ 1.01 & 4.62 $\pm$ 4.24 & 2.90 $\pm$ 2.90 \tabularnewline
3 & \textbf{1.79} $\pm$ 0.96 & \textbf{1.72} $\pm$ 1.22 & 2.77 $\pm$ 1.44 & 2.88 $\pm$ 2.42 \tabularnewline
\hline
\end{tabular}
		\end{center}
		\vspace{-0.3cm}
		\caption{Localization. RPE evaluated with a delta unit of 1 meter.}
		\vspace{0.1cm}
		\label{tab:benchmark_slam_pose}
	\end{subtable}
	\begin{subtable}{\linewidth}
		\setlength{\tabcolsep}{2.65pt}
		\begin{center}
		\begin{tabular}{|>{\centering}p{0.5cm}|>{\centering}p{2.4cm}>{\centering}p{0.8cm}|>{\centering}p{2.4cm}>{\centering}p{0.8cm}|}
\hline
Test & \multicolumn{2}{c|}{ORB-SLAM2 + PSPNet} & \multicolumn{2}{c|}{SUMA++}\tabularnewline
Seq. & Acc./ Comp. / F$_{1}$ & mIoU & Acc./ Comp. / F$_{1}$ & mIoU \tabularnewline
\hline
0 & 78.5 /  \textbf{72.8} / \textbf{75.5} & \textbf{35.3} & \textbf{90.8} / 63.1 / 74.5 & 19.9 \tabularnewline
1 & 81.8 /  \textbf{76.8} / \textbf{79.2} & \textbf{31.6} & \textbf{89.3} / 62.9 / 73.8 & 17.3 \tabularnewline
2 & 82.5 /  \textbf{70.8} / \textbf{76.2} & \textbf{30.4} & \textbf{89.6} / 64.5 / 75.0 & 17.6 \tabularnewline
3 & 84.3 /  \textbf{79.1} / \textbf{81.6} & \textbf{32.7} & \textbf{94.2} / 66.3 / 77.8 & 22.8 \tabularnewline
\hline
\end{tabular}

		\end{center}
		\vspace{-0.3cm}
		\caption{Semantic mapping evaluated at a threshold of 10cm.}
		\label{tab:benchmark_slam_recon}
	\end{subtable}
	\caption{{\bf Quantitative Results for Semantic SLAM.}  Evaluated on $4$ test sequences.}
\end{table}

\boldparagraph{Geometric and Semantic Mapping}
We measure the quality of the geometric and semantic mapping using the same metrics considered in the semantic scene completion benchmark. As richer input observations are available in this task, we adopt a smaller distance threshold of 10cm as the main metric. Specifically, we first measure geometry accuracy using completeness and accuracy, and then evaluate semantics on the completed ground truth points via the confidence weighted mIoU. 
As the mapping accuracy is highly correlated with the APE, we compare ground truth and estimated reconstruction in local windows to minimize the impact of pose drifts. Each local window consists of $50$ consecutive frames and is aligned to the ground truth based on the trajectory, see \appref{app:benchmark_slam_mapping} for more details.

We use the same baselines considered in the localization benchmark.
As ORB-SLAM2 does not provide dense reconstruction nor semantic information, we obtain dense semantic reconstruction by unprojecting 2D semantic segmentations (PSPNet~\cite{Zhao2017CVPR}) using depth maps from semi-global matching (SGM)~\cite{Hirschmueller2008PAMI}. SUMA++ aims for semantic SLAM and estimates poses and a semantic surfel map from LiDAR scans. We experimentally observe that it is sufficient to take the center of the surfels as the reconstructed points.

\tabref{tab:benchmark_slam_recon} and \figref{fig:benchmark_slam_semantic} show the reconstruction and semantic prediction results. We observed that both baselines produce good reconstructions on the ground region. For regions above the ground, SUMA++ is less complete as it only uses LiDAR scans and thus the maximum height is limited. ORB-SLAM2 + SGM results in higher completeness but worse accuracy. In terms of semantic predictions, SUMA++ produces reasonable results on the LiDAR scans but struggles to achieve good overall performance due to its low completeness. In contrast, ORB-SLAM2 + PSPNet contains more flying points due to the outliers of stereo matching (\eg, sky points colored blue). Exploring semantic information to remove sky points may further improve the performance of this baseline.

\boldparagraph{Discussion}
We evaluate localization accuracy of existing SLAM methods and suggest exploring 3D instance-level information in further works.
Further, reconstructing accurate geometry and semantics remains a challenging task. Our benchmark allows to investigate important questions towards solving this challenging task, \eg, which input modality is better suited for this task, whether semantic prediction and geometric reconstruction can benefit each other and if joint optimization is desirable.

\section{Conclusion}

We present KITTI-360, a large scale 3D video dataset comprising 300k images and laser point clouds with consistent semantics in both 2D and 3D. We create a WebGL-based annotation tool and annotate both static and dynamic objects in 3D. We propose a method to obtain dense semantic instance labels from annotated 3D primitives. In the presence of 3D data, our method yields better results compared to several 2D label transfer baselines while lowering the annotation time. 

Furthermore, we establish novel online benchmarks for several challenging tasks at the intersection of computer vision, graphics and robotics. We evaluate several baselines for each benchmark. Our results show that existing methods achieve satisfactory results on well-established benchmarks, \eg, 2D/3D segmentation, where inference is directly performed on given observations. However, it is much harder to solve tasks that require jointly recovering the geometry, appearance and estimating the semantics as in the newly introduced tasks for semantic scene completion, novel view appearance/semantic synthesis and semantic SLAM.
We hope that our dataset, online benchmarks and annotation tools will fertilize new research across communities, fostering progress towards the grand goal of full autonomy.

\ifCLASSOPTIONcompsoc
  \section*{Acknowledgments}
\else
  \section*{Acknowledgment}
\fi

The authors thank Siyuan Peng, Bernhard Jaeger, Shrisha Bharadwaj, Apratim Bhattacharyya, Paul Henderson, and Zehao Yu for their help in implementing the baselines, Kashyap Chitta, Katja Schwarz, and Yue Wang for proofreading, and SurfingTech for annotating parts of the dataset. Andreas Geiger was supported by the ERC Starting Grant LEGO-3D (850533) and the DFG EXC number 2064/1 - project number 390727645. Yiyi Liao was supported by the German Federal Ministry of Education and Research (BMBF): T{\"u}bingen AI Center, FKZ: 01IS18039A.

{
\bibliographystyle{IEEEtranS}
\bibliography{bibliography_short,bibliography,bibliography_custom}
}

\vspace{-1 cm}%
\begin{IEEEbiography}[{\includegraphics[width=1in,clip,keepaspectratio]{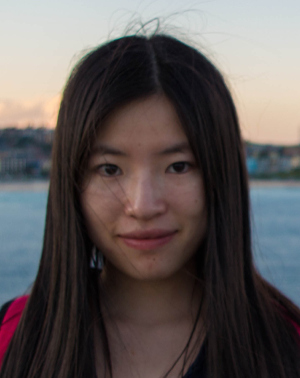}}]{Yiyi Liao}
received her Ph.D. degree from the Department of Control Science and Engineering, Zhejiang University, China in 2018. She is currently a PostDoctoral researcher at the Autonomous Vision Group, University of T{\"u}bingen and Max Planck Institute for Intelligent Systems, Germany. Her research interests include 3D vision and scene understanding.
\end{IEEEbiography}
\vspace{-1 cm}%

\begin{IEEEbiography}[{\includegraphics[width=1in,clip,keepaspectratio]{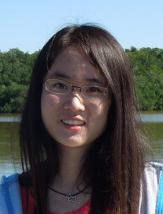}}]{Jun Xie} 
received her Ph.D. degree from the Electrical Engineering Department of University of Washington in 2016. She is currently a research engineer at Google. Her research interests include image segmentation and 3D vision.
\end{IEEEbiography}
\vspace{-1 cm}%

\begin{IEEEbiography}[{\includegraphics[width=1in,clip,keepaspectratio]{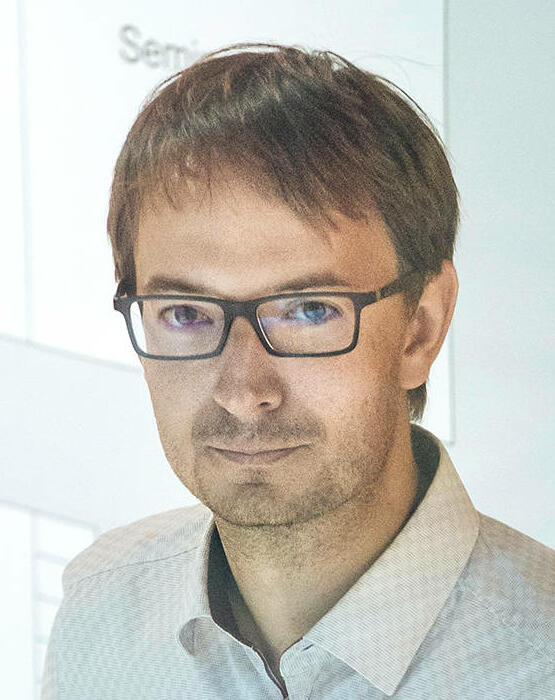}}]{Andreas Geiger}
received his Diploma in computer science and his Ph.D. degree from Karlsruhe Institute of Technology in 2008 and 2013. Currently, he is leading the Autonomous Vision Group at the University of T{\"u}bingen and the Max Planck institute for Intelligent Systems in T{\"u}bingen. His research interests include computer vision, machine learning and scene understanding with a focus on self-driving vehicles.
\end{IEEEbiography}

\labelLastPage
\newpage
\pagenumbering{gobble}

\clearpage
\begin{appendices}

\section{Annotation Data Preparation}
\label{app:pcd_accumulation_anno}
\subsection{Point Cloud Accumulation}
To facilitate annotation, we accumulate all laser measurements in a common world coordinate system and augment them with 3D points from stereo matching \cite{Hirschmueller2008PAMI}. To reduce outliers of stereo matching, we consider only points up to $15$m distance, and apply left-right as well as forward-backward consistency checks over 5 frames. We fuse all 3D points sequentially and ignore a point closer than $5$ cm to its nearest neighbor in the fused point cloud. This downsampling operation allows for reducing the data loading traffic and memory of the web based annotation tool.

\subsection{Dense Point Cloud on Dynamic Objects}
\label{app:dynamic_detection}
Our simple dynamic object detection consists of two steps. In the first step, we apply volumetric fusion over a sequential of laser scans and search for 3D points located in (mostly) free regions. As a dynamic object always moves in the free space, the voxels along its moving trajectory are occupied occasionally and thus the expected status over time should be free. Hence the dynamic points can be detected by finding points inside of all ``free'' voxels. Specifically, we build a 3D occupancy volumetric grid $\cV$ for each batch by fusing Velodyne observations using Octomap \cite{Hornung2013AR}. Given a set of measurements $z_{1:t}$ from frame $1$ to $t$, the occupancy probability of each voxel $v\in \cV$ is updated as follow:
\begin{align}
p(v|z_{1:t}) & = \begin{cases}
 \max(\min(p(v|z_{1:t-1}) + p(v|z_t), p_{max}), p_{min})  \\ 
 \hspace{3.5cm} \text{ if } p(v|z_{1:t-1})>p_{min} \\ 
 p_{min} \\
 \hspace{3.5cm} \text{ if }  p(v|z_{1:t-1})=p_{min}
\end{cases}
\end{align}
where $p(v|z_{1:t})$ is the log-odds of the probability, $p_{max}$ and $p_{min}$ are the upper bound and lower bound of the log-odds respectively. Note that we clamp a voxel to be free if there is sufficient evidence from previous frames to support its free status. We denote the set of free voxels as $\widetilde{\cV}$.

\begin{figure}[t]
\begin{subfigure}{.5\linewidth}
\includegraphics[width=\linewidth]{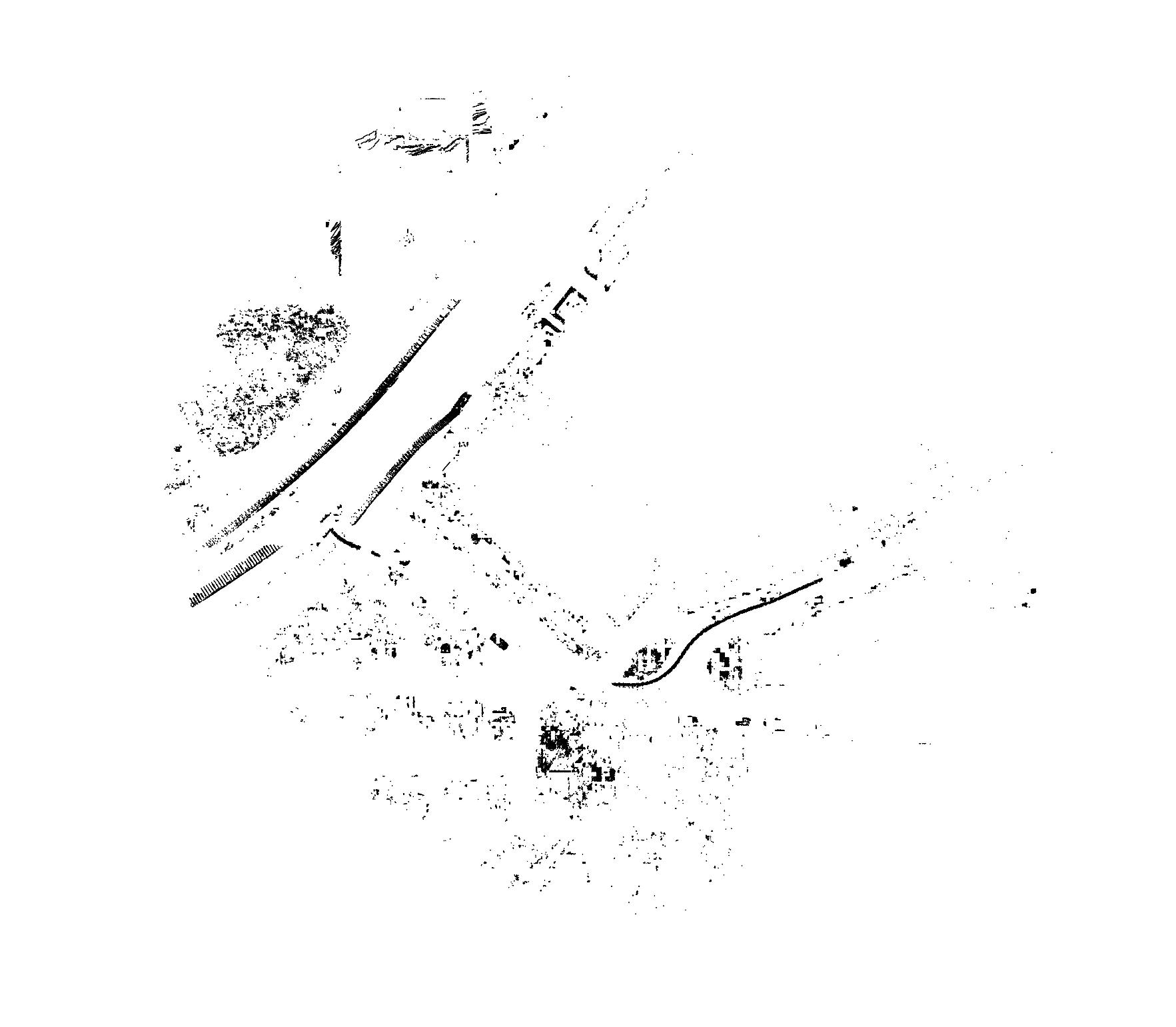}
\caption{Fine-grained detection}
\label{fig:low_level}
\end{subfigure}%
\hfill
\begin{subfigure}{.5\linewidth}
\includegraphics[width=\linewidth]{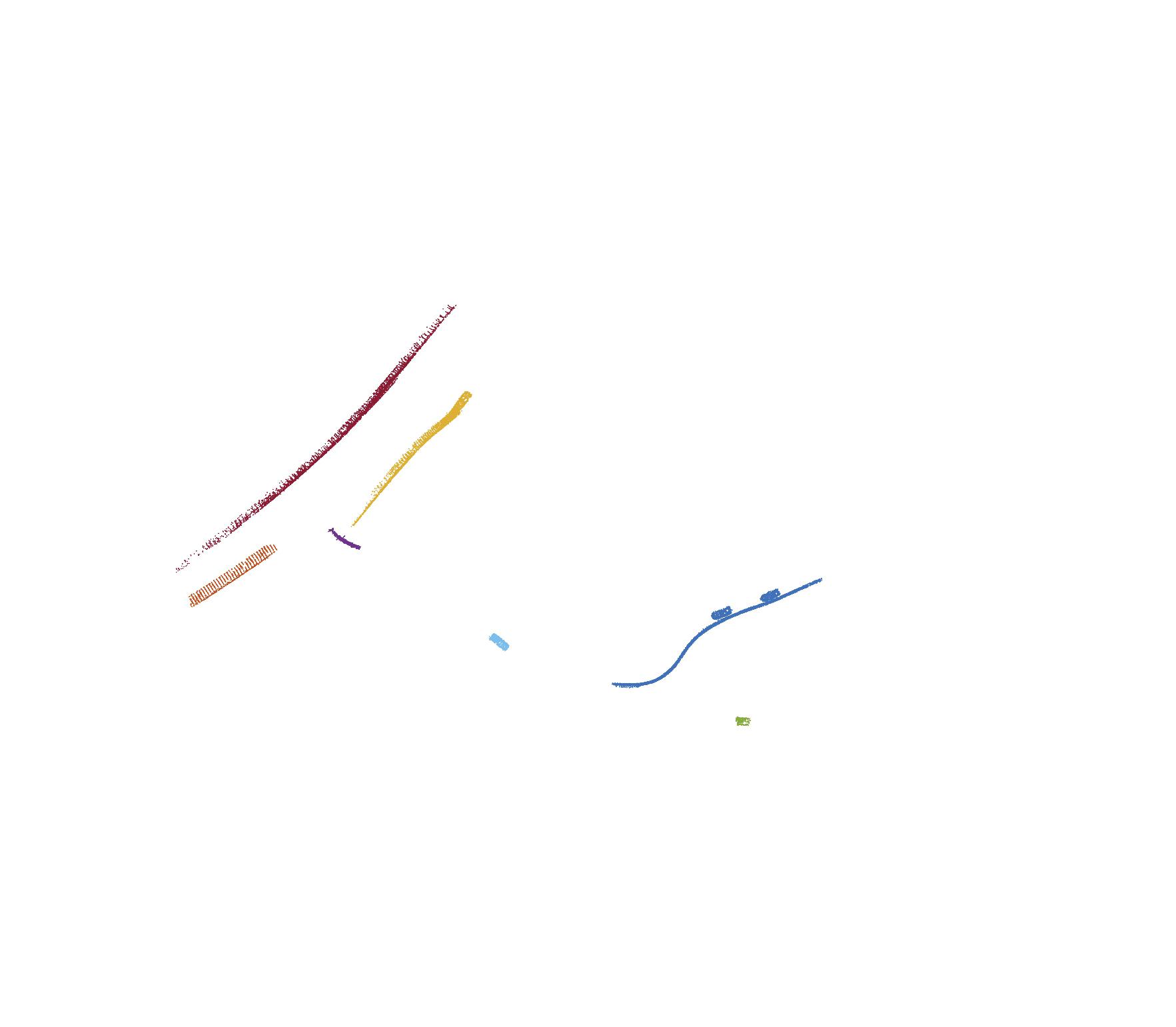}
\caption{Coarse-grained detection}
\label{fig:high_level}
\end{subfigure}%
\caption{{\bf Dynamic Object Detection.} (\subref{fig:low_level}) Detected dynamic points with decision on 3D grids.
(\subref{fig:high_level}) Detected dynamic points with decision on point cloud clusters.
}
\label{fig:dynamic_detection}
\end{figure}

Due to the noise in measurements and poses, the free voxels may also contain many 3D points of static objects as shown in \figref{fig:low_level}. It is hard to avoid these false positive detections as each voxel is classified as free or occupied independently. Therefore, we consider a second step to filter out clusterings of noisy detections. Specifically, we segment the original accumulated point cloud into a branch of clusters using the Region Growing algorithm\footnote{\url{https://pcl.readthedocs.io/projects/tutorials/en/latest/region_growing_segmentation.html}}, and we calculate the occupancy probability of each cluster $c$ based on the detection in the first step:
\begin{align}
p(c) & =  \frac{1}{N} \sum_i^N \left [ c_i \in  \widetilde{\cV} \right ]
\end{align}
where $c_i$ denotes a single point in the cluster, $N$ denotes number of points, and $c_i \in  \widetilde{\cV}$ means that $c_i$ is spatially located within ${\cV}$. A cluster $c$ is considered as dynamic if $p(c)$ is larger than a given threshold. With the second step, we are able to filter out false positive detections as shown in \figref{fig:high_level}. It is acceptable if a few false positive detections remains since the dense point cloud will be further labeled by our annotators.

\begin{figure}[t]
\begin{subfigure}{.5\linewidth}
\centering
\includegraphics[width=0.95\linewidth]{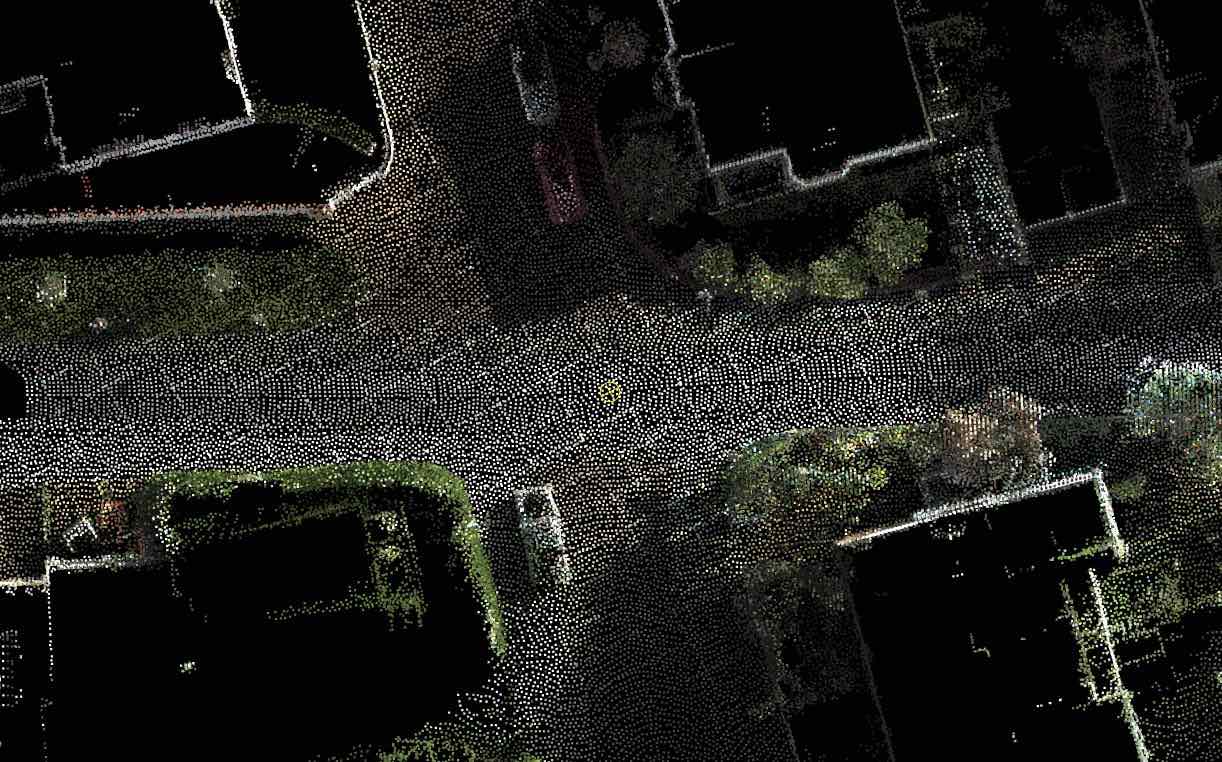}
\caption{RGB}
\label{fig:pcl_color1}
\end{subfigure}%
\hfill
\begin{subfigure}{.5\linewidth}
\centering
\includegraphics[width=0.95\linewidth]{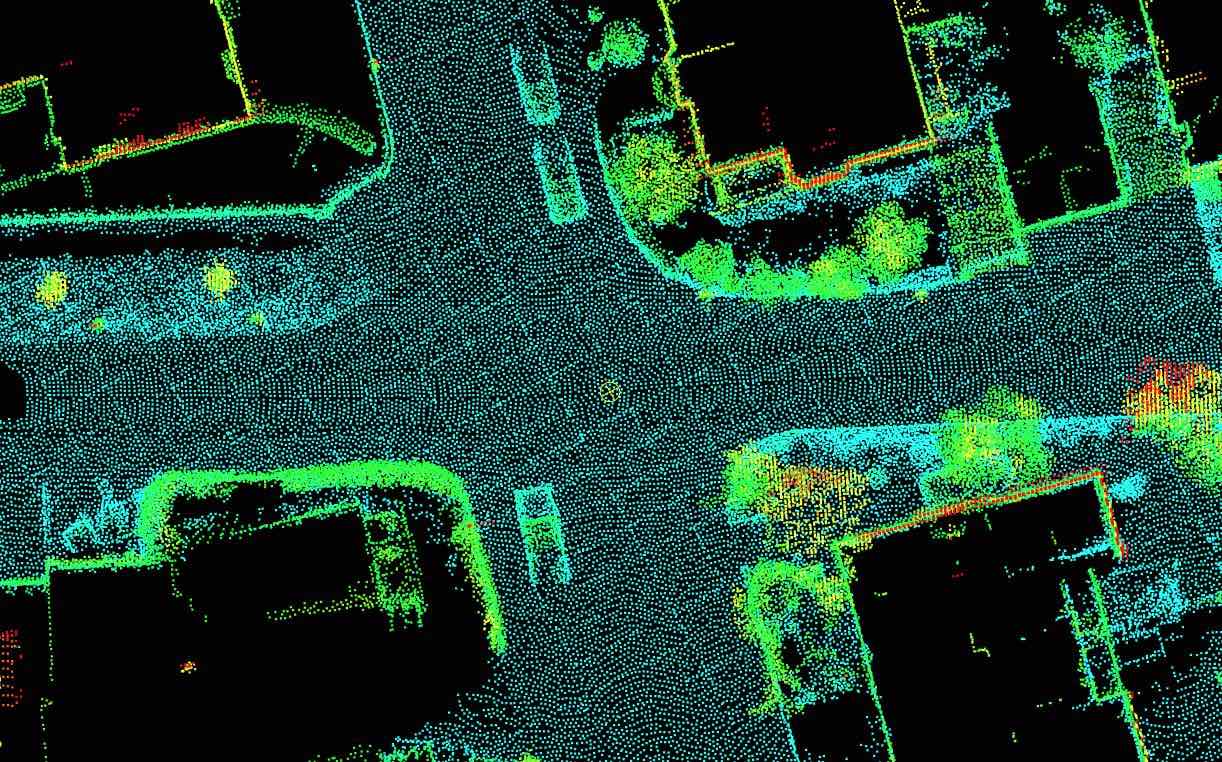}
\caption{Color \wrt height}
\label{fig:pcl_color2}
\end{subfigure}%
\\
\begin{subfigure}{.5\linewidth}
\centering
\includegraphics[width=0.95\linewidth]{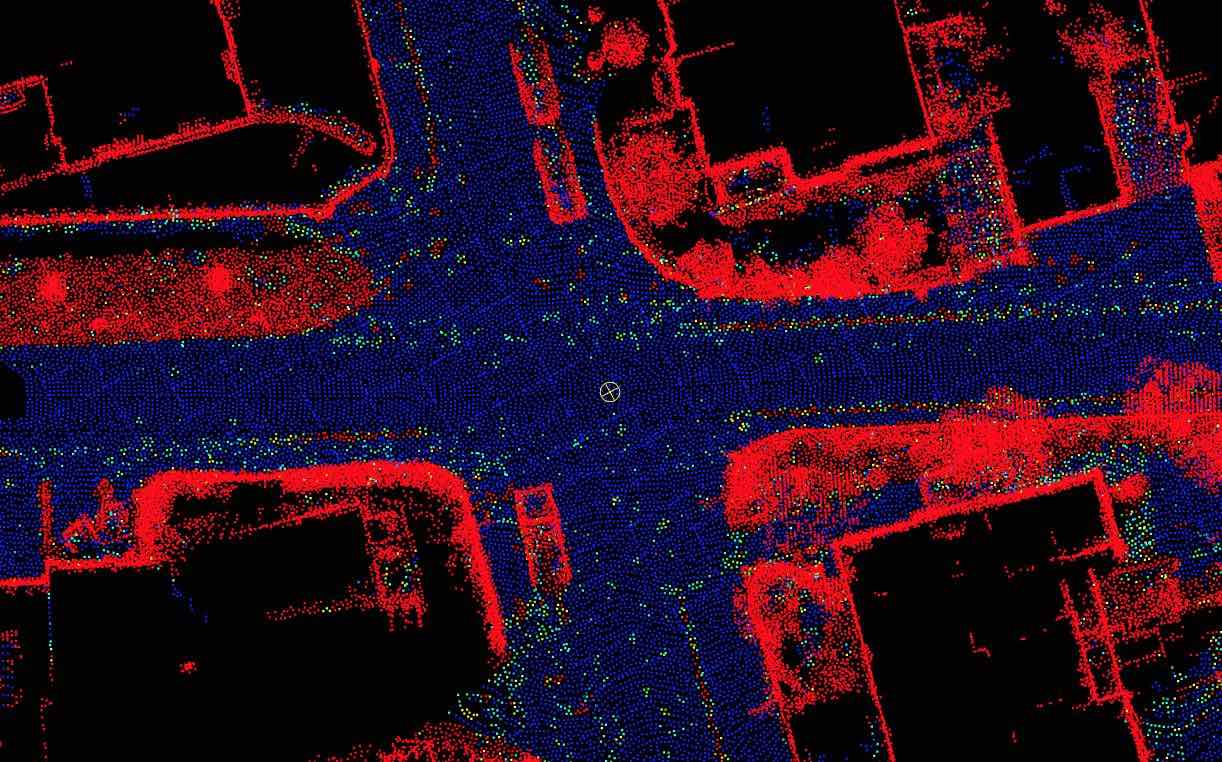}
\caption{Color \wrt height variance}
\label{fig:pcl_color3}
\end{subfigure}%
\hfill
\begin{subfigure}{.5\linewidth}
\centering
\includegraphics[width=0.95\linewidth]{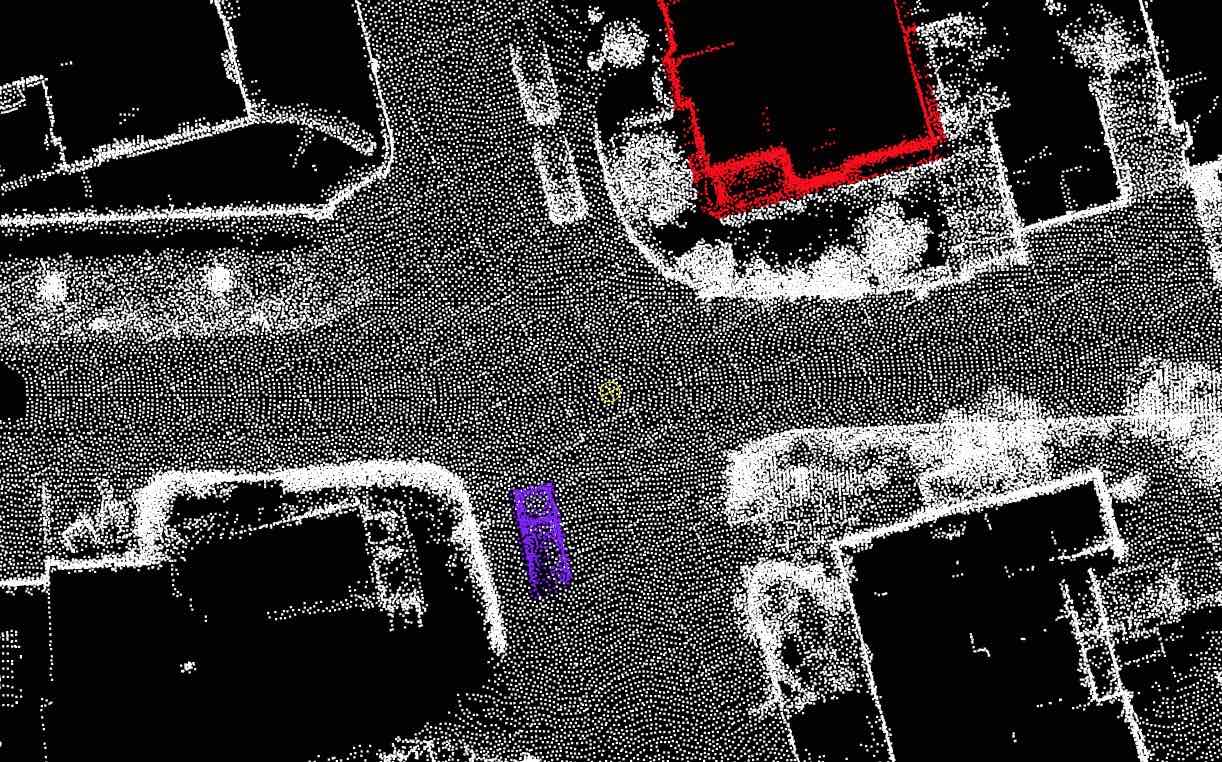}
\caption{Color \wrt annotation}
\label{fig:pcl_color4}
\end{subfigure}%
\caption{{\bf Color Codings} of 3D points supported by our annotation interface.
}
\label{fig:pcl_color_coding}
\end{figure}

\begin{figure}[t]
\begin{subfigure}{.49\linewidth}
\includegraphics[width=\linewidth]{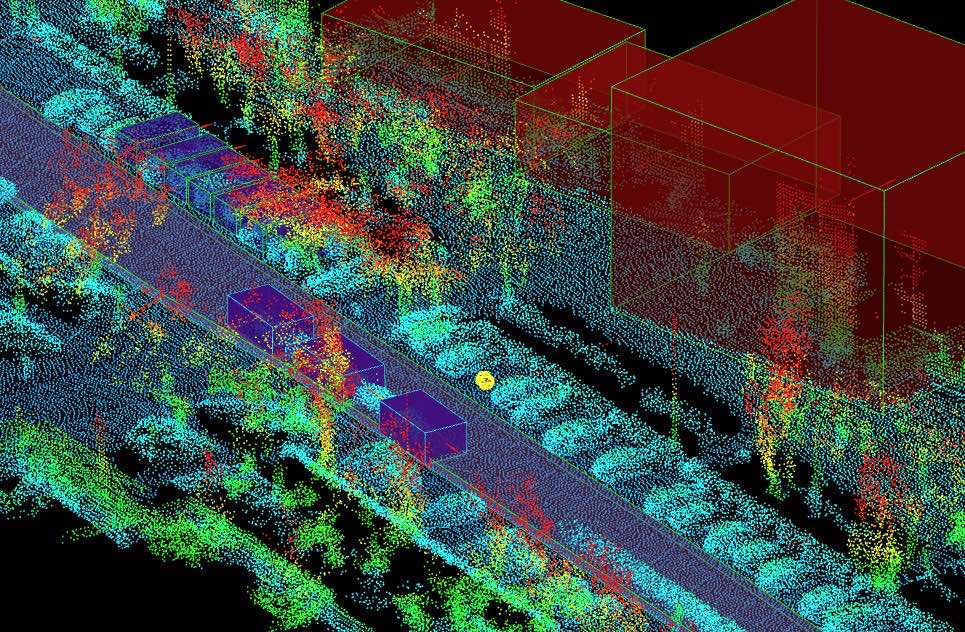}
\caption{Normal viewport}
\end{subfigure}%
\hfill
\begin{subfigure}{.49\linewidth}
\includegraphics[width=\linewidth]{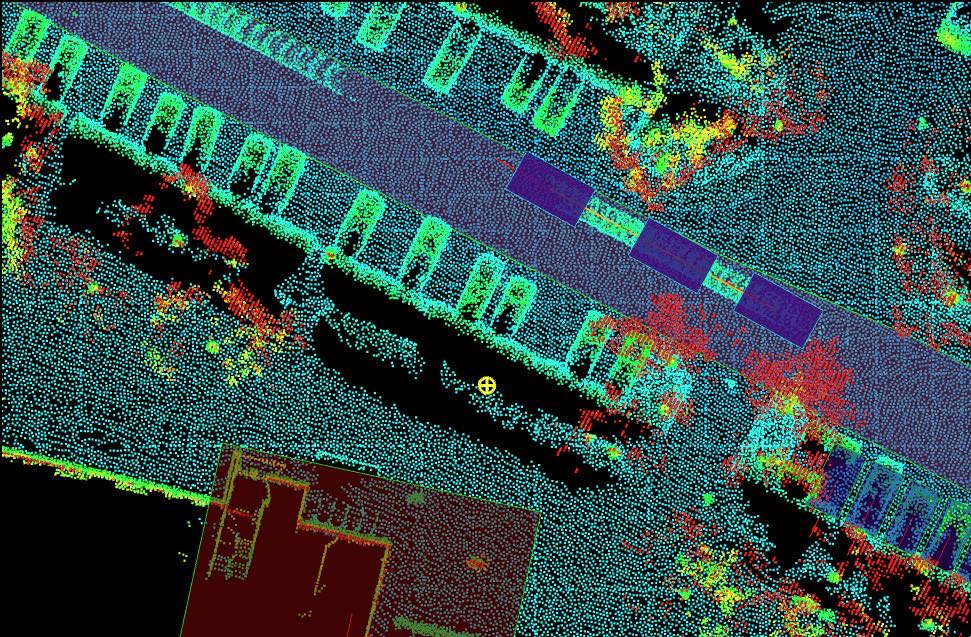}
\caption{Orthographic viewport}
\end{subfigure}%
\caption{\bf{3D Viewport.}}
\label{fig:3d_views}
\end{figure}

\begin{figure}[t]
\center
\includegraphics[width=0.85\linewidth]{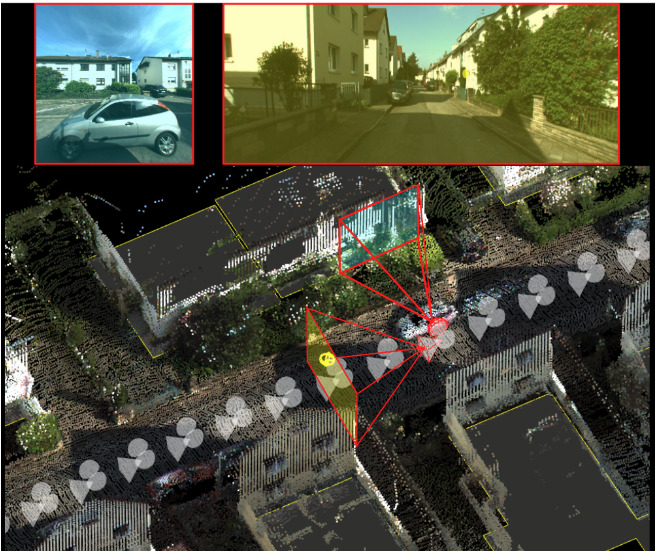}
\caption{{\bf 2D Camera View.}  We illustrate fisheye and perspective camera views and virtual camera poses.}
\label{fig:camera_view}
\end{figure}

\begin{figure}[t]
\center
\includegraphics[width=0.9\linewidth]{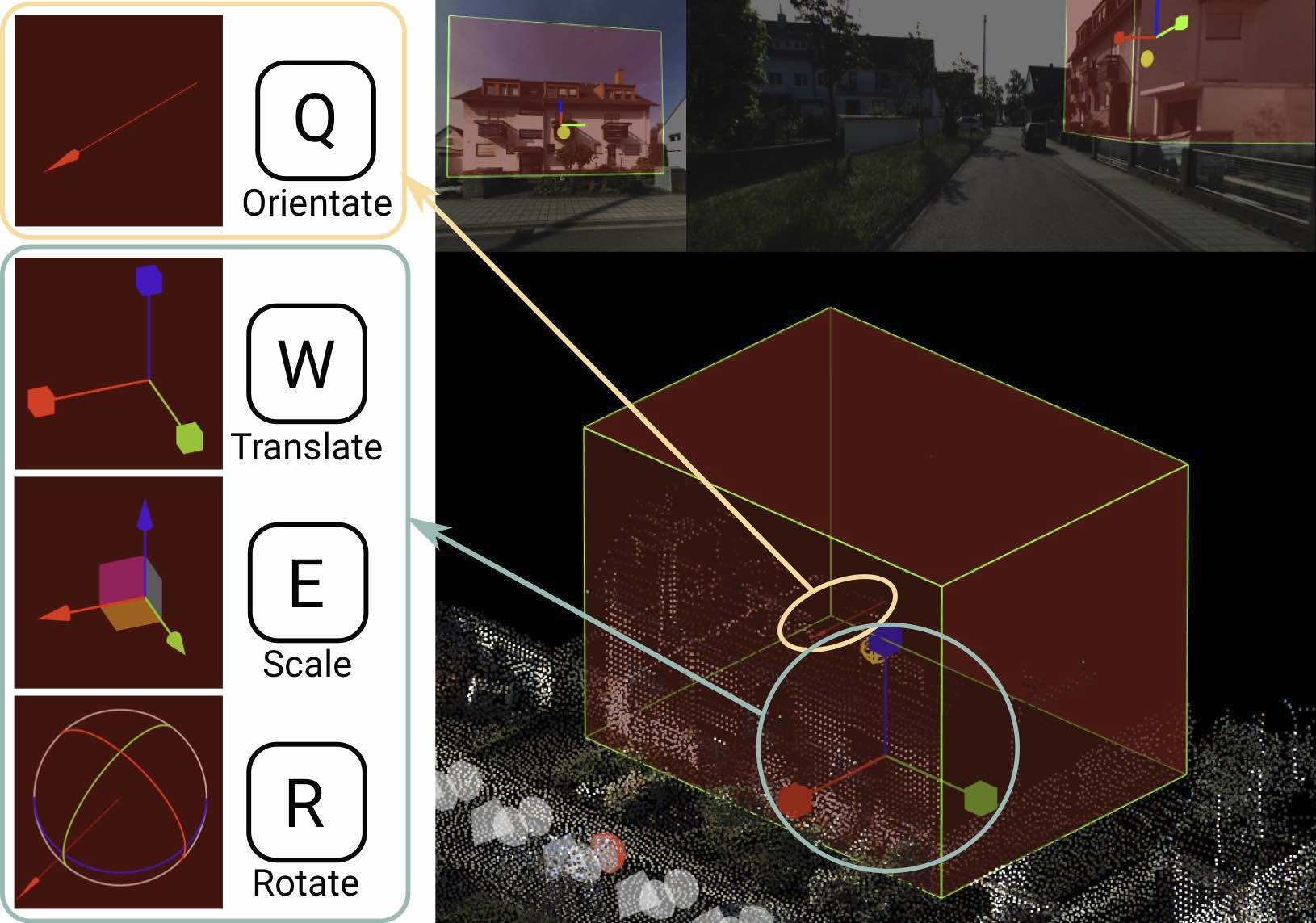}
\caption{{\bf Bounding Primitive Manipulation.}
} 
\label{fig:anno_shortcut1}
\end{figure}

\begin{figure}[t]
\center
\includegraphics[width=0.9\linewidth]{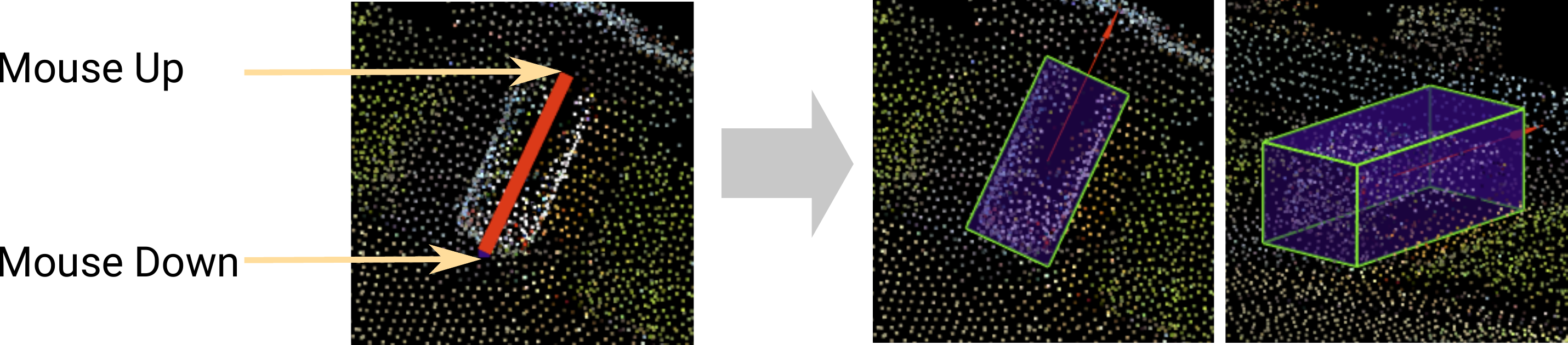}
\caption{{\bf Fast Object Annotation.}
} 
\label{fig:anno_shortcut2}
\end{figure}

\begin{figure}[t]
\begin{subfigure}{.54\linewidth}
\includegraphics[width=\linewidth]{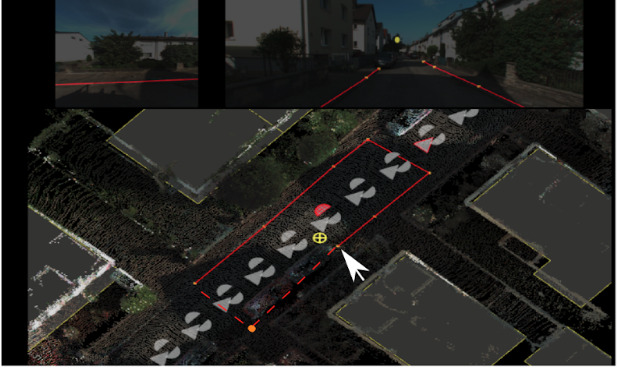}
\caption{Ground annotation process}
\end{subfigure}%
\hfill
\begin{subfigure}{.45\linewidth}
\includegraphics[width=\linewidth]{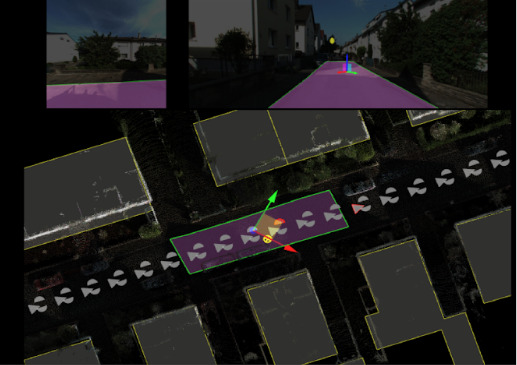}
\caption{Finished annotation}
\end{subfigure}%
\caption{\bf Ground Annotation.}
\label{fig:ground_annotation}
\end{figure}

\section{Annotation Interface}
\label{app:ann_interface}
In this section, we demonstrate the annotation tool and process in detail.

\subsection{Annotation Scene}

\boldparagraph{Color Coding} \figref{fig:pcl_color_coding} shows different color codings of 3D points that we provide in the annotation tool. Annotators can choose different color codings accordingly. 
For example, \figref{fig:pcl_color3} helps annotators to identify the boundary between ``Road'' and ``Sidewalk'' and \figref{fig:pcl_color4} allows annotators to check unannotated region (shown as white).

\boldparagraph{3D Viewpoint}
To assist annotators to better visualize the scene, we also provide different viewports, namely, normal and orthographic viewports as shown in \figref{fig:3d_views}. The orthographic viewport helps annotators accurately identify stuff classes' boundaries and annotate individual objects efficiently (see ``fast object annotation mode'' in \secref{app:ann_functions}). Besides, annotators can also adjust point size and brightness to work with different levels of detail.

\boldparagraph{2D Camera View}
For better perceiving the scene, we also show fisheye and perspective images as well as the pose of each camera, enabling annotators to select informative viewpoints efficiently, as shown in \figref{fig:camera_view}.

\subsection{Annotation Functions}
\label{app:ann_functions}
We provide a few shortcuts and functions to facilitate the annotation process.

\boldparagraph{Bounding Primitive Manipulation} 
To best enclose the 3D object, each bounding primitive can be manipulated with translation, scaling, and rotation, resulting in $9$ degrees of freedom. In addition, annotators are also asked to assign an orientation to each bounding primitive to understand objects' (especially instances) orientation. See \figref{fig:anno_shortcut1} for each operation's shortcuts. 

\boldparagraph{Bounding Primitive Copy} 
We also provide a ``copy'' shortcut to allow annotators to quickly insert new annotations with the same label and similar pose to previously annotated objects. This is especially useful for objects appearing frequently with similar sizes such as building and car.

\boldparagraph{Fast Object Annotation} 
We support quickly annotating a subset of object classes by simply drawing a line along the object. This fast annotation mode is enabled under the orthographic view for a few pre-selected classes, where the annotator draws a line along the longest side in the middle. While this line only specifies the length of the object in one dimension and its orientation, we heuristically place a bigger bounding primitive centered at this line and iteratively shrink the bounding primitive until it touches any non-ground points. As shown in \figref{fig:anno_shortcut2}, this simple technique allows for efficient and accurate annotation.

\boldparagraph{Object-Centric Mode}
To enable a clear observation of a single object from the accumulated point cloud, we also help annotators with the ``object-centric'' mode. With a single 3D bounding box selected, triggering the ``object-centric'' mode hides all the other bounding primitives as well as points far from the selected primitive. In addition, both front view and side views images are automatically switched to the ones in which the selected primitive is most visible. 

\boldparagraph{Completeness Check}
As illustrated in \figref{fig:pcl_color4}, the annotator can check the completeness level of the annotation by visualizing the point cloud based on existing bounding primitives. Specifically, we color each 3D point if it is enclosed by a bounding primitive and leave the unlabeled region as white. This helps the annotator easily identify any unlabeled 3D points.

\subsection{Ground Annotation}
\label{app:ann_ground}
Ground bounding primitives are simply annotated as 2D polygons. The extruded height of the ground polygon is automatically determined as follows: for each vertex $v=\{x,y\}$ on the polygon, we first search the nearest camera of this given vertex, and assign the height of the camera to this vertex as its initial height $\hat{z}$. Then we search nearest neighbors of point $\{x,y,\hat{z}\}$ in the 3D point cloud, and update height $z$ as the median height of these nearest neighbors. See \figref{fig:ground_annotation} as an example for ``Road'' annotation. Annotators can also modify the 2D polygon anytime by dragging its control points.

\subsection{Dynamic Object Annotation}
\label{app:dynamic_annotation}

We implement a semi-automatic annotation scheme to label dynamic objects efficiently. 
Our semi-automatic annotation relies on two assumptions: the size of the dynamic object is fixed over time, and its trajectory is smooth. Therefore, the required annotation is reduced to adding posed 3D primitives at several keyframes. Our annotation tool then automatically places the remaining primitives along the trajectory. The smooth trajectory is obtained via spline interpolation based on the primitives at the keyframes. We annotate articulated dynamic objects, e.g., pedestrians, using the maximum extent bounding box.

\begin{figure}[t]
\begin{subfigure}{.9\linewidth}
\includegraphics[width=\linewidth]{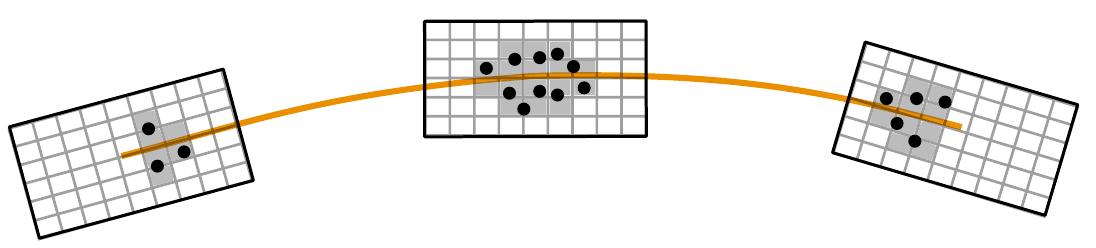}
\caption{Annotated bounding primitives}
\label{fig:dynamic_illustration_01}
\end{subfigure}%
\hfill
\begin{subfigure}{.9\linewidth}
\includegraphics[width=\linewidth]{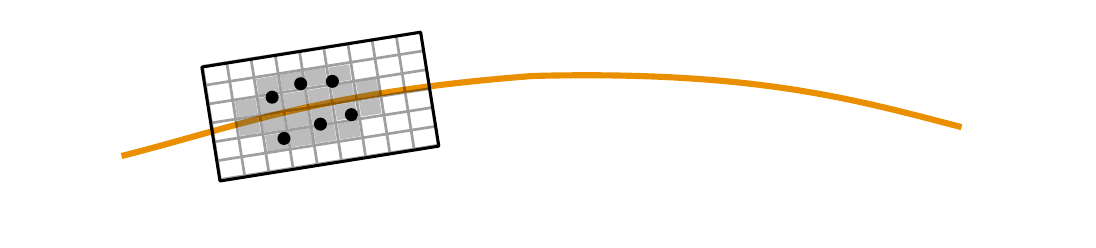}
\caption{Automatically generated bounding primitive}
\label{fig:dynamic_illustration_02}
\end{subfigure}%
\caption{{\bf Semi-Automatic Dynamic Object Annotation.} (\subref{fig:dynamic_illustration_01}) We discretize the annotated 3D primitives and fuse the occupancy status.
(\subref{fig:dynamic_illustration_02}) We search along the spline and find the matching bounding primitive by maximizing the overlap.
}
\label{fig:dynamic_illustration}
\end{figure}

As the speed of the dynamic object may not be constant, we place the remaining primitives based on the observed 3D points at each timestamp. Specifically, we first discretize the annotated 3D primitives into voxels and fuse the occupancy status of each voxel over all annotated primitives as shown in \figref{fig:dynamic_illustration_01}. A voxel is considered as \textit{occupied} if it is occupied in any of the annotated 3D primitives, otherwise it is \textit{free}. This fused occupancy status is considered an ``occupancy template'' for searching matching 3D primitive along the trajectory. Given a timestamp between the first and the last annotated timestamp, we slide the 3D primitive on the interpolated spline and calculate the occupancy status based on 3D points collected at the given timestamp. We choose the pose that provides the maximum overlap with the occupancy template, see \figref{fig:dynamic_illustration_02}. 

\begin{figure}[t]
\center
\includegraphics[width=\linewidth]{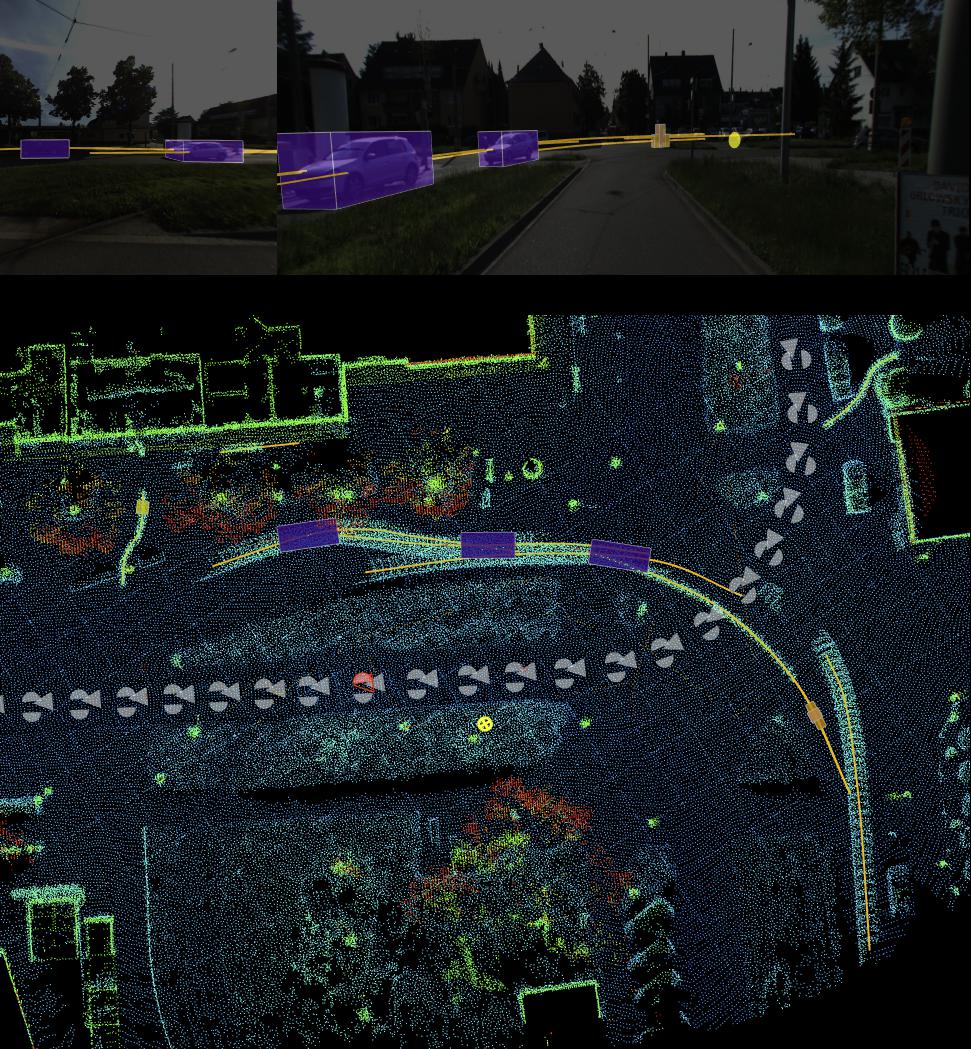}
\caption{{\bf Dynamic Annotations.} The above 3D primitives are automatically generated along the interpolated spline visualized in orange.
}
\label{fig:dynamic_anno}
\end{figure}

To facilitate users' interaction, we plot the interpolated spline in the annotation tool and allow the annotator to refine the spline by simply adjusting the 3D bounding primitives inserted at the keyframes. We also display the automatically generated 3D primitives to help the user check if they are accurate. The poses of each automatically generated bounding primitive can also be adjusted if necessary. Typically, it requires $2\sim5$ annotated keyframes to produce accurate annotations for the full moving trajectory. \figref{fig:dynamic_anno} illustrates the dynamic annotation at a given timestamp. 

\section{Label Definition}
\label{app:label_definition}
\tabref{tab:labels} shows the definition of the $37$ classes that we use for annotating 3D scenes. We adhere to the definition of Cityscapes as close as possible while a few inconsistent label definitions are inevitable as our annotations are performed in 3D. For example, the ``Traffic Sign'' in Cityscapes only includes the front side while we consider both the front and the back. We do not distinguish the back as each traffic sign is labeled by a single 3D bounding primitive and it might be observed in 2D from both sides. Thus, each traffic sign has a consistent label regardless of which side it is observed.

\begin{table*}
\scriptsize
\setlength{\tabcolsep}{2.27pt}
\renewcommand{\arraystretch}{1.2}
\begin{center}
\begin{tabular}
{p{1.7cm}p{2.9cm}p{1.0cm}p{12.0cm}} \hline
Category  & Class & Instance & Definition  \tabularnewline
\hline
\multirow{1}{*}{flat}    	& road   & \xmark               & 
Horizontal surfaces on which cars usually drive,
including road markings. Typically delimited by
curbs, rail tracks, or parking areas. However, road
is not delimited by road markings and thus may
include bicycle lanes or roundabouts. \tabularnewline     
	& sidewalk     & \xmark           & 
Horizontal surfaces designated for pedestrians or
cyclists. Delimited from the road by some obstacle, e.g. curbs or poles (might be small), but not
only by markings. Often elevated compared to the
road and often located at the side of a road. The
curbs are included in the sidewalk label. Also includes the walkable part of traffic islands, as well
as pedestrian-only zones, where cars are not allowed to drive during regular business hours. If
it’s an all-day mixed pedestrian/car area, the correct label is ground. \tabularnewline      
	& parking      & \xmark           & 
Horizontal surfaces that are intended for parking
and separated from the road, either via elevation
or via a different texture/material, but not separated merely by markings.\tabularnewline      
	& rail track   & \xmark           & 
Horizontal surfaces on which only rail cars can
normally drive. If rail tracks for trams are embedded in a standard road, they are included in the
road label. \tabularnewline 
\hline     
\multirow{1}{*}{construction}	& building $^\dagger$    & \cmark           &
Includes structures that house/shelter humans,
e.g. low-rises, skyscrapers.
Translucent buildings made of glass still receive
the label building. Also includes scaffolding attached to buildings. \tabularnewline      
	& garage$^*$     & \cmark     &
Structures for parking. \tabularnewline 
	& wall           & \xmark    & 
Individually standing walls that separate two (or
more) outdoor areas, and do not provide support
for a building. \tabularnewline      
	& fence          & \xmark        &
Structures with holes that separate two (or more)
outdoor areas, sometimes temporary. \tabularnewline      
	& guard rail     & \xmark        &
Metal structure located on the side of the road to
prevent serious accidents. Rare in inner cities,
but occur sometimes in curves. Includes the bars
holding the rails  \tabularnewline      
	& bridge         & \xmark       &
Bridges (on which the ego-vehicle is not driving)
including everything (fences, guard rails) permanently attached to them. \tabularnewline      
	& tunnel         & \xmark        &
Tunnel walls and the (typically dark) space encased by the tunnel, but excluding vehicles. \tabularnewline     
	& gate$^*$       & \xmark        & 
A passageway or opening in a wall of fence for entrance or exit. \tabularnewline
	& stop$^*$       & \cmark        & 
A place where a bus or train stops for passengers to get on or off. It intends to protect waiting pedestrians from rain, wind and snow. \tabularnewline
\hline 
\multirow{1}{*}{object}	& pole$^\dagger$                   & \cmark  &
Long, vertically oriented poles, e.g. sign
poles or traffic light poles. This does not include objects mounted on the pole, which have a
larger diameter than the pole itself (e.g. most street
lights). \tabularnewline
	& smallpole$^*$      & \cmark        &
Small, vertically oriented poles, e.g. sign
poles or traffic light poles. This does not include objects mounted on the pole, which have a
larger diameter than the pole itself (e.g. most street
lights).  \tabularnewline      
	& traffic light      & \cmark      &
The traffic light box without its poles in all orientations and for all types of traffic participants, e.g.
regular traffic light, bus traffic light, train traffic
light \tabularnewline      
	& traffic sign $^\dagger$   & \cmark  &
Signs installed by the state/city authority with the purpose of conveying information
to drivers/cyclists/pedestrians, e.g. traffic signs,
parking signs, direction signs, or warning reflector posts. \textit{Both frontal and back side are included.} \tabularnewline    
	& lamp$^*$               & \cmark     & 
A lamp usually mounted on a pole and constituting one of a series spaced at intervals along a public road or highway. \tabularnewline
	& trash bin$^*$          & \cmark     & 
A container that holds materials that have been thrown away. \tabularnewline
	& vending machine$^*$    & \cmark     & 
An automated machine for selling merchandise. \tabularnewline
	& box$^*$                & \cmark     & 
Any rigid typically rectangular container excluding trash bin and vending machine. Some examples are electric box, honey bucket, package, etc. \tabularnewline 
\hline
\multirow{1}{*}{nature}	& vegetation      & \xmark       &
Trees, hedges, and all kinds of vertically growing vegetation. Plants attached to buildings/walls/fences are not annotated separately,
and receive the same label as the surface they are
supported by. \tabularnewline      
	& terrain          & \xmark             &
Grass, all kinds of horizontally spreading vegetation, soil, or sand. These are areas that are not
meant to be driven on. This label may also include a possibly adjacent curb. Single grass stalks
or very small patches of grass are not annotated
separately and thus are assigned to the label of the
region they are growing on.  \tabularnewline    
\hline
\multirow{1}{*}{sky}	& sky            & \xmark               & 
Open sky (without tree branches/leaves) \tabularnewline      
\hline
\multirow{1}{*}{human}	& person         & \cmark                &
All humans that would primarily rely on their legs
to move if necessary.  \tabularnewline      
	& rider         & \cmark                 &
Humans relying on some device for movement. This includes drivers, passengers, or riders
of bicycles, motorcycles. \tabularnewline   
\hline   
\multirow{1}{*}{vehicle}	       & car          & \cmark                 &
This includes cars, jeeps, SUVs, vans with a continuous body shape (i.e. the driver’s cabin and
cargo compartment are one). Does not include
trailers, which have their own separate class. \tabularnewline      
	& truck       & \cmark                   &
This includes trucks, vans with a body that is separate from the driver’s cabin, pickup trucks, as well
as their trailers. \tabularnewline      
	& bus         & \cmark                   &
This includes buses that are intended for 9+ persons for public or long-distance transport. \tabularnewline      
	& caravan     & \cmark                   &
Vehicles that (appear to) contain living quarters.
This also includes trailers that are used for living
and has priority over the trailer class. \tabularnewline      
	& trailer     & \cmark                   &
Includes trailers that can be attached to any vehicle, but excludes trailers attached to trucks. The
latter are included in the truck label. \tabularnewline      
	& train       & \cmark                   &
All vehicles that move on rails, e.g. trams, trains. \tabularnewline      
	& motorcycle     & \cmark                &
This includes motorcycles, mopeds, and scooters
without the driver or other passengers. The latter
receive the label rider \tabularnewline      
	& bicycle        & \cmark                &
This includes bicycles without the cyclist or other
passengers. The latter receive the label rider. \tabularnewline      
\hline
\multirow{1}{*}{void}	& unknown construction & \cmark   & 
All remaining construction regions which are not mentioned in the ``construction'' session\tabularnewline
	& unknown vehicle   & \cmark     & 
All remaining vehicle regions which are not mentioned in the ``vehicle'' session\tabularnewline
	& unknown object    & \cmark     & 
All remaining object regions which are not mentioned in the ``object'' session\tabularnewline
\hline
\end{tabular}

\end{center}
$^*$ classes do not exist in Cityscapes \\
$^\dagger$ classes with definitions slightly different from Cityscapes\\
\caption{{\bf Label Definition.} We adhere to the label definition of Cityscapes as close as possible. Inconsistent classes are marked.}
\label{tab:labels}
\end{table*}

\section{More Details of Label Transfer Inference}

\subsection{Accumulation of Input Point Cloud}
\label{app:pcd_accumulation}
In contrast to the point cloud accumulation for \textit{annotation} as introduced in \appref{app:pcd_accumulation_anno}, here we need to distinguish static and dynamic points for label transfer \textit{inference} and thus determine visible points on every frame. Specifically, we consider a point static if it is not enclosed by any dynamic bounding primitive and accumulate all static points first. For each dynamic object, we retrieve all points inside the corresponding bounding primitives $\{b^m_t\}$ for every timestamp $t$ and accumulate them in the canonical object-centered coordinate system by taking the inverse transform of the object pose defined by $\{b^m_t\}$ (world-to-object transformation). Next, we insert the accumulated dynamic point clouds back to the world coordinate following the object pose (object-to-world transformation). This allows us to obtain dense 3D points during inference for both static and dynamic regions.

\subsection{Accumulation of Inferred 3D Label}
\label{app:inference_accumulation}
Our inference is performed individually on each frame defined over the corresponding 2D pixels and visible 3D points. To obtain 3D labels on the accumulated point clouds, we thus fuse 3D labels obtained from each frame. 
Specifically, if a 3D point is visible in multiple frames, we take the majority of its inferred classes as its final label. The confidence of this 3D point is also averaged over confidence values of these points of the majority label. If a 3D point is not visible in any of the frames but is uniquely labeled by a single class, we assign this unique label to the 3D point and a confidence value of $1.0$. In the remaining cases, we treat the 3D point's label as unknown.

\subsection{Pixel Unary Potentials of Ground Objects}
The first term of the pixel unary potential is a binary feature $\xi^{\cP}_i(s_i)\in\{0,1\}$ which indicates admissible labels. For non-planar object classes, $\xi^{\cP}_i(s_i)$ is obtained based on the projection of \textit{3D} primitives, whereas planar object classes directly use projections of \textit{2D} polygons to obtain more accurate boundaries. As introduced in \appref{app:ann_ground}, the ground bounding primitives are extruded to 3D to enclose the 3D points. This leads to oversized 2D projections, making it hard to determine the boundary between two adjacent ground object annotations (\eg, ``Road'' and ``Sidewalk''). As opposed to our conference version~\cite{Xie2016CVPR}, which exploits a geometric unary potential to address this problem, we instead directly project the 2D polygons of the ground objects before extruding them to 3D. This allows us to obtain accurate ground object boundaries from $\xi^{\cP}_i(s_i)$ and avoid the complexity introduced by the additional geometric unary term.

\subsection{Instance Augmentation of Pixel Unary Potentials}
\label{app:ablation_object_boudary}
\begin{figure*}[t]
\def\qualwidth{0.5}
\def\qualmargin{0.1}
\begin{minipage}{0.5\linewidth}%
\includegraphics[width=\linewidth]{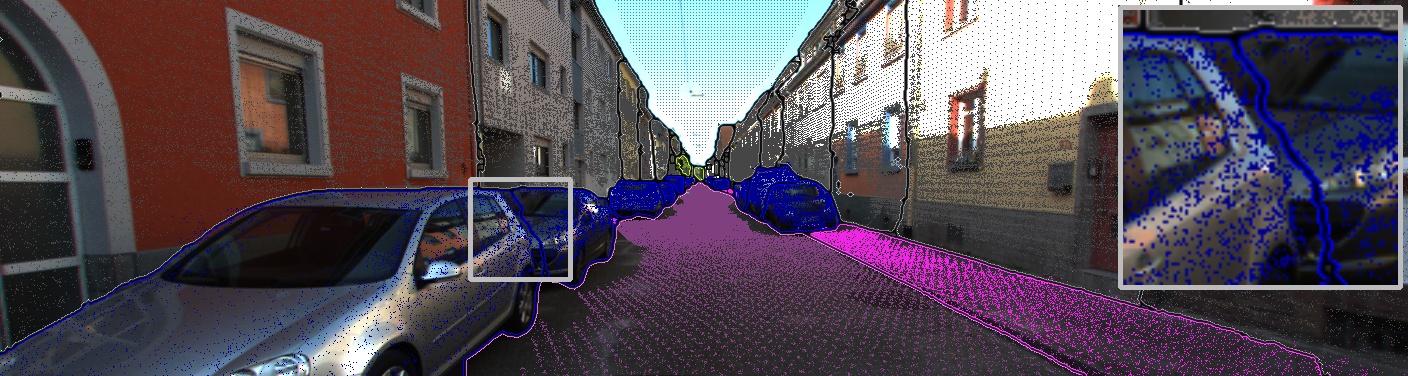}\\%
\includegraphics[width=\linewidth]{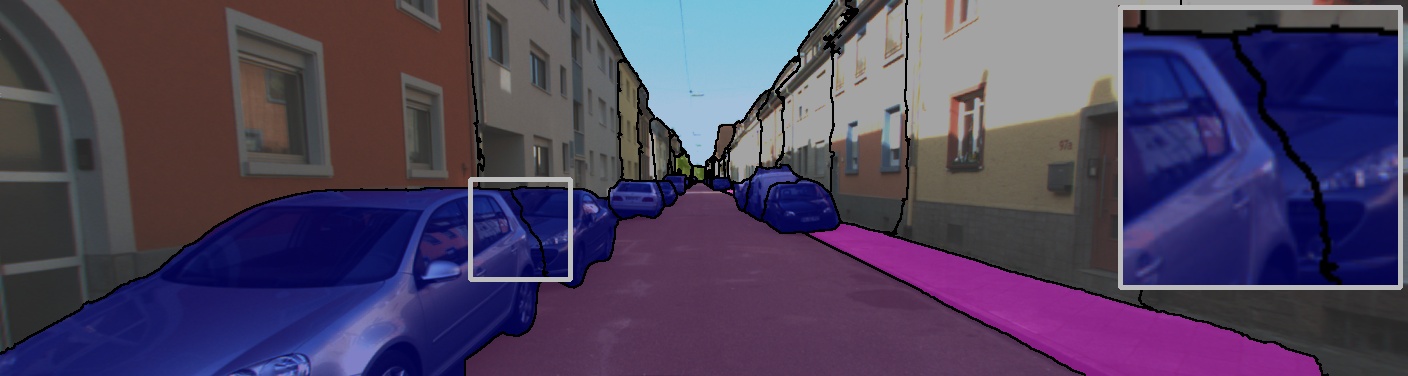}\\%
\vspace{-0.6cm}
\subcaption{Without Instance Unary}
\label{fig:unary_instance_wo}
\end{minipage}%
\hspace{\qualmargin cm}%
\begin{minipage}{0.5\linewidth}%
\includegraphics[width=\linewidth]{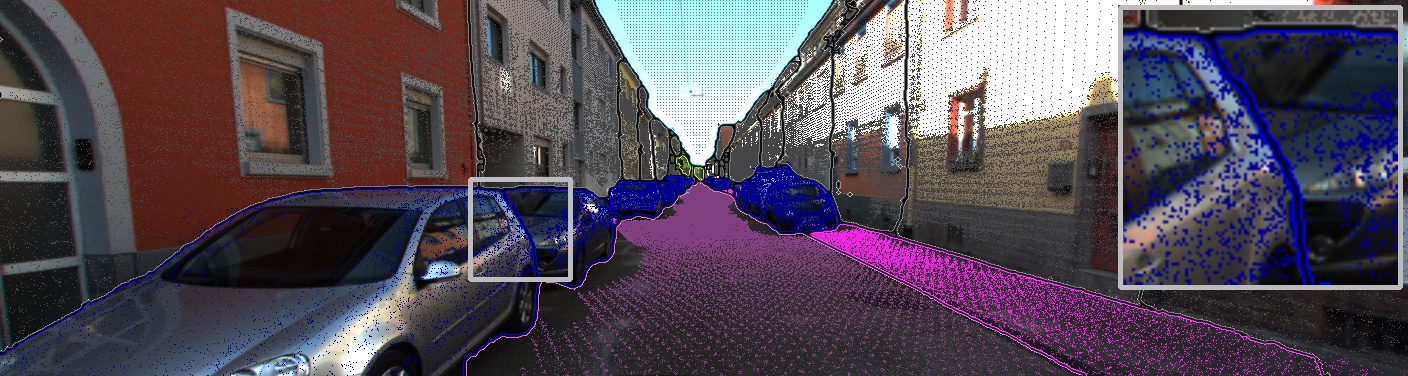}\\%
\includegraphics[width=\linewidth]{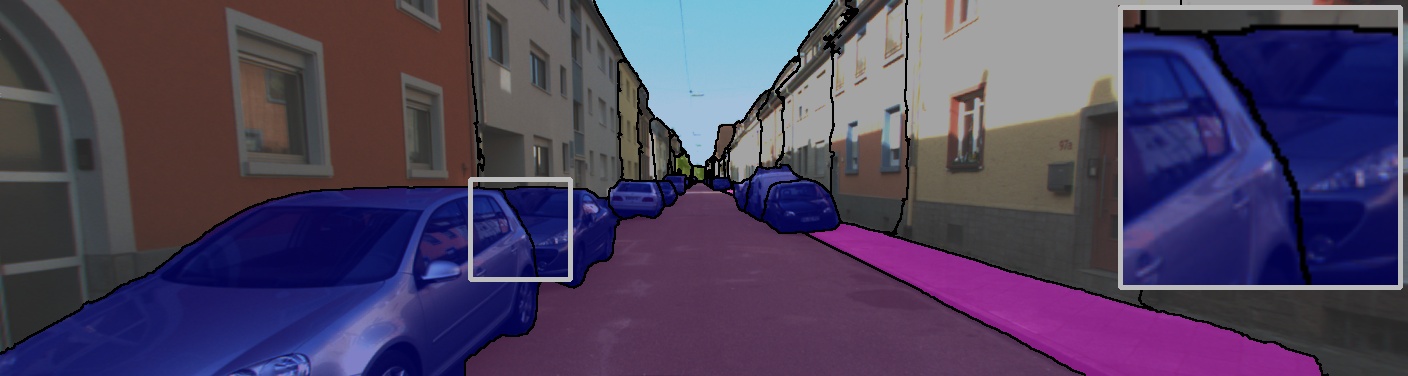}\\%
\vspace{-0.6cm}
\subcaption{With Instance Unary}
\label{fig:unary_instance_with}
\end{minipage}%
\hspace{\qualmargin cm}%
\caption{{\bf Ablation Study of Instance Unary.} The top row shows the input image with the projected 3D points and inferred semantic segmentation boundaries. The second row shows the inferred semantic instance segmentation. As can be seen from (\subref{fig:unary_instance_wo}), it is challenging to distinguish instance boundaries given sparse projections of point clouds. This can be effectively improved by incorporating instance unary as shown in (\subref{fig:unary_instance_with}). }
\vspace{0.3cm}
\label{fig:ablation_object_boundary}
\end{figure*}

We adopt a state-of-the-art panoptic segmentation method UPSNet~\cite{Xiong2019CVPR} pre-trained on Cityscapes~\cite{Cordts2016CVPR} to obtain instance hypotheses for ``Car'', ``Truck'', and ``Pedestrian''. We first run UPSNet on our images to get probability maps of all instances. For each annotated instance within the aforementioned classes, we retrieve matched instances from the predictions of UPSNet. More specifically, given a set of 3D points annotated with one instance (\eg, one car) and a probability map of one instance predicted by UPSNet, we consider they match if more than $50\%$ of the 3D points fall into the high-probability region of the predicted instance. 
This allows for improving instance boundaries as shown in \figref{fig:ablation_object_boundary}.

\section{More Results of Label Transfer Inference}
\label{app:label_transfer_result}

\subsection{Detailed Quantitative Comparisons}
\label{app:label_transfer_result_quantitative}
Here, we show detailed quantitative comparisons for individual classes. \tabref{tab:baseline_static_detailed} and \tabref{tab:baseline_dynamic_detailed} show quantitative comparison to label transfer baselines on static and dynamic objects, respectively. We evaluate the intersection over union (IoU) of each class where the mIoU is the average over all classes. We further show detailed ablation study in \tabref{tab:ablation_semantic_detailed} and \tabref{tab:ablation_instance_detailed} for semantic and instance label transfer.

\subsection{Qualitative Comparison to Baselines}

We compare our method qualitatively to several 2D-to-2D and 3D-to-2D label transfer baselines in \figref{fig:qualitative_results_baselines_app}. Note how the 2D-to-2D label transfer baselines fail in the presence of strong occlusions and large displacements.
\vspace{-0.2cm}

\def\qualwidth{0.3}
\def\qualmargin{0.1}
\newcommand{\qualresultnew}[3]{%
\begin{figure*}[h]
\centering
\begin{minipage}{#1\linewidth}%
\includegraphics[width=\linewidth]{results_sup/#2/active_label_propagation/output/#3_label.jpg}\\%
\includegraphics[width=\linewidth]{results_sup/#2/active_label_propagation/output/#3_err.jpg}\\%
\vspace{-0.5cm}%
\subcaption{Label Prop. \cite{Vijayanarasimhan2012ECCV}}
\vspace{0.1cm}
\end{minipage}%
\hspace{\qualmargin cm}%
\begin{minipage}{#1\linewidth}%
\includegraphics[width=\linewidth]{results_sup/#2/sparseTrack/output/#3_label.jpg}\\%
\includegraphics[width=\linewidth]{results_sup/#2/sparseTrack/output/#3_err.jpg}\\%
\vspace{-0.5cm}%
\subcaption{Sparse Track. + GC \cite{Sundaram2010ECCV}}
\vspace{0.1cm}
\end{minipage}%
\hspace{\qualmargin cm}%
\begin{minipage}{#1\linewidth}%
\includegraphics[width=\linewidth]{results_sup/#2/3dTrack/output/#3_label.jpg}\\%
\includegraphics[width=\linewidth]{results_sup/#2/3dTrack/output/#3_err.jpg}\\%
\vspace{-0.5cm}%
\subcaption{3D Prop. + GC}
\vspace{0.1cm}
\end{minipage}%
\hspace{\qualmargin cm}%
\begin{minipage}{#1\linewidth}%
\includegraphics[width=\linewidth]{results_sup/#2/fullyConnect/output/#3_label.jpg}\\%
\includegraphics[width=\linewidth]{results_sup/#2/fullyConnect/output/#3_err.jpg}\\%
\vspace{-0.5cm}%
\subcaption{Fully Conn. CRF \cite{Kraehenbuehl2011NIPS}}
\vspace{0.1cm}
\end{minipage}%
\hspace{\qualmargin cm}%
\begin{minipage}{#1\linewidth}%
\includegraphics[width=\linewidth]{results_sup/#2/pspnet/output/#3_label.jpg}\\%
\includegraphics[width=\linewidth]{results_sup/#2/pspnet/output/#3_err.jpg}\\%
\vspace{-0.5cm}%
\subcaption{PSPNet \cite{Zhao2017CVPR}}
\vspace{0.1cm}
\end{minipage}%
\hspace{\qualmargin cm}%
\begin{minipage}{#1\linewidth}%
\includegraphics[width=\linewidth]{results_sup/#2/result_Box/#3_label.jpg}\\%
\includegraphics[width=\linewidth]{results_sup/#2/result_Box/#3_err.jpg}\\%
\vspace{-0.5cm}%
\subcaption{3D Primitives + GC}
\vspace{0.1cm}
\end{minipage}%
\hspace{\qualmargin cm}%
\begin{minipage}{#1\linewidth}%
\includegraphics[width=\linewidth]{results_sup/#2/result_Mesh/#3_label.jpg}\\%
\includegraphics[width=\linewidth]{results_sup/#2/result_Mesh/#3_err.jpg}\\%
\vspace{-0.5cm}%
\subcaption{3D Mesh + GC}
\vspace{0.1cm}
\end{minipage}%
\hspace{\qualmargin cm}%
\begin{minipage}{#1\linewidth}%
\includegraphics[width=\linewidth]{results_sup/#2/result_Pts/#3_label.jpg}\\%
\includegraphics[width=\linewidth]{results_sup/#2/result_Pts/#3_err.jpg}\\%
\vspace{-0.5cm}%
\subcaption{3D Points + GC}
\vspace{0.1cm}
\end{minipage}%
\hspace{\qualmargin cm}%
\begin{minipage}{#1\linewidth}%
\includegraphics[width=\linewidth]{results_sup/#2/ours/output/#3_label.jpg}\\%
\includegraphics[width=\linewidth]{results_sup/#2/ours/output/#3_err.jpg}\\%
\vspace{-0.5cm}%
\subcaption{Proposed Method}
\vspace{0.1cm}
\end{minipage}%
\hspace{\qualmargin cm}%
\vspace{-0.3cm}
\caption{{\bf Qualitative Comparison to Baselines.} Each subfigure shows from top-to-bottom: the input image with inferred semantic segmentation and the errors with respect to 2D ground truth annotation where colors indicate ground truth labels.}
\label{fig:qualitative_results_baselines_app}
\end{figure*}
}
\qualresultnew{\qualwidth}{baseline_new}{result/result_0000_003425}%

\subsection{Qualitative Comparison of Ablation Study}
\figref{fig:qualitative_results_ablation_app} compares different variants of our label transfer model. Consistent with the quantitative analysis, our full model achieves the best performance.

\def\qualwidth{0.3}
\def\qualmargin{0.1}
\newcommand{\qualresultabla}[3]{%
\begin{figure*}[h]
\centering
\begin{minipage}{#1\linewidth}%
\includegraphics[width=\linewidth]{results_sup/#2/dynamic_method1/#3_label.jpg}\\%
\includegraphics[width=\linewidth]{results_sup/#2/dynamic_method1/#3_err.jpg}\\%
\vspace{-0.5cm}%
\subcaption{LA}
\vspace{0.1cm}
\end{minipage}%
\hspace{\qualmargin cm}%
\begin{minipage}{#1\linewidth}%
\includegraphics[width=\linewidth]{results_sup/#2/dynamic_method2/#3_label.jpg}\\%
\includegraphics[width=\linewidth]{results_sup/#2/dynamic_method2/#3_err.jpg}\\%
\vspace{-0.5cm}%
\subcaption{LA + 3D}
\vspace{0.1cm}
\end{minipage}%
\hspace{\qualmargin cm}%
\begin{minipage}{#1\linewidth}%
\includegraphics[width=\linewidth]{results_sup/#2/dynamic_method3/#3_label.jpg}\\%
\includegraphics[width=\linewidth]{results_sup/#2/dynamic_method3/#3_err.jpg}\\%
\vspace{-0.5cm}%
\subcaption{LA + PW}
\vspace{0.1cm} 
\end{minipage}%
\hspace{\qualmargin cm}%
\begin{minipage}{#1\linewidth}%
\includegraphics[width=\linewidth]{results_sup/#2/dynamic_method4/#3_label.jpg}\\%
\includegraphics[width=\linewidth]{results_sup/#2/dynamic_method4/#3_err.jpg}\\%
\vspace{-0.5cm}%
\subcaption{LA + PW + CO }
\vspace{0.1cm}
\end{minipage}%
\hspace{\qualmargin cm}%
\begin{minipage}{#1\linewidth}%
\includegraphics[width=\linewidth]{results_sup/#2/dynamic_method5/#3_label.jpg}\\%
\includegraphics[width=\linewidth]{results_sup/#2/dynamic_method5/#3_err.jpg}\\%
\vspace{-0.5cm}%
\subcaption{LA + PW + CO + 3D}
\vspace{0.1cm}
\end{minipage}%
\hspace{\qualmargin cm}%
\begin{minipage}{#1\linewidth}%
\includegraphics[width=\linewidth]{results_sup/#2/dynamic/#3_label.jpg}\\%
\includegraphics[width=\linewidth]{results_sup/#2/dynamic/#3_err.jpg}\\%
\vspace{-0.5cm}%
\subcaption{Full Model}
\vspace{0.1cm}
\end{minipage}%
\hspace{\qualmargin cm}%
\vspace{-0.3cm}
\caption{{\bf Qualitative Results for Ablation Study.} Each subfigure shows from top-to-bottom: the input image with inferred semantic segmentation, and the errors with respect to 2D ground truth annotation where colors indicate ground truth labels.}
\label{fig:qualitative_results_ablation_app}
\end{figure*}
}
\qualresultabla{\qualwidth}{ablation_new}{instance_vis_key2/2013_05_28_drive_0000_sync/0000001672}%

\begin{table*}[t]
\setlength{\tabcolsep}{2.65pt}
\begin{center}
\begin{tabular}{|p{3.5cm}|>{\centering}p{0.7cm}>{\centering}p{0.7cm}>{\centering}p{0.7cm}>{\centering}p{0.7cm}>{\centering}p{0.7cm}>{\centering}p{0.7cm}>{\centering}p{0.7cm}>{\centering}p{0.7cm}>{\centering}p{0.7cm}>{\centering}p{0.7cm}>{\centering}p{0.7cm}>{\centering}p{0.7cm}>{\centering}p{0.7cm}>{\centering}p{0.7cm}|>{\centering}p{0.7cm}|>{\centering}p{0.7cm}|}  \hline
\small Method & \small Road & \small Park & \small Sdwlk & \small Terr & \small Bldg & \small Vegt & \small Car & \small Trler & \small Crvn & \small Gate & \small Wall & \small Fence & \small Box & \small Sky & \small mIoU & \small Acc \tabularnewline \hline
{\small Label Prop.\cite{Vijayanarasimhan2012ECCV} }&85.9 & 27.8 & 59.4 & 48.7 & 78.8 & 67.5 & 51.0 & 12.6 & 48.1 & 25.8 & 47.1 & 44.3 & 0.0 & 89.6 & 49.0 & 81.0 \tabularnewline
{\small Sparse Track. + GC \cite{Sundaram2010ECCV} }&83.4 & 33.0 & 38.7 & 49.8 & 76.2 & 61.0 & 75.1 & 73.5 & 78.8 & 24.6 & 6.7 & 25.0 & 11.6 & 79.0 & 51.2 & 79.1 \tabularnewline
{\small 3D Prop. + GC }&77.1 & 50.5 & 75.4 & 64.5 & 83.1 & 76.8 & 82.1 & 91.2 & \textbf{91.7} & 62.6 & 59.9 & 58.1 & 55.5 & 81.6 & 72.1 & 87.4 \tabularnewline
{\small Fully Conn. CRF \cite{Kraehenbuehl2011NIPS} }&90.1 & 46.4 & 67.4 & 61.3 & 88.3 & 78.4 & 85.6 & 48.9 & 78.1 & 30.5 & 33.7 & 45.6 & 43.1 & 92.7 & 63.6 & 88.7 \tabularnewline
{\small PSPNet \cite{Zhao2017CVPR} }&\textbf{95.6} & 46.2 & 77.1 & 64.8 & 88.9 & 81.7 & 91.5 & 46.5 & 84.0 & 30.6 & 41.7 & 50.2 & 52.3 & 89.2 & 67.2 & 90.4 \tabularnewline
\hline
{\small 3D Primitives + GC }&81.7 & 31.4 & 45.9 & 22.5 & 59.6 & 56.7 & 63.0 & 67.1 & 61.7 & 42.3 & 25.5 & 52.3 & 31.3 & 50.3 & 49.4 & 73.4 \tabularnewline
{\small 3D Mesh + GC }&91.7 & 54.6 & 67.6 & 31.4 & 81.3 & 72.1 & 85.2 & 93.5 & 86.0 & 59.4 & 35.9 & 61.2 & 50.1 & 65.6 & 66.8 & 85.7 \tabularnewline
{\small 3D Points + GC }&93.5 & 62.2 & 76.5 & 37.2 & 82.0 & 74.1 & 87.5 & \textbf{94.7} & 85.7 & 73.2 & 52.2 & 69.0 & \textbf{61.1} & 68.0 & 72.6 & 87.8 \tabularnewline
{\small Proposed Method }&95.2 & \textbf{72.9} & \textbf{84.5} & \textbf{67.9} & \textbf{90.3} & \textbf{84.2} & \textbf{92.2} & 93.4 & 90.8 & \textbf{78.8} & \textbf{64.3} & \textbf{73.1} & 56.8 & \textbf{92.8} & \textbf{81.2} & \textbf{93.1} \tabularnewline
\hline
\end{tabular}
\end{center}
\caption{{\bf Comparison to Label Transfer Baselines on Semantic Segmentation Transfer of Static Objects}. We compare our method to 2D label transfer baselines (top) and to 3D to 2D label transfer baselines (bottom) on $120$ consecutive images of static objects.}
\label{tab:baseline_static_detailed}
\end{table*}

\begin{table*}[t]
\setlength{\tabcolsep}{2.65pt}
\begin{center}
\begin{tabular}{|p{3.5cm}|>{\centering}p{1.0cm}>{\centering}p{1.0cm}>{\centering}p{1.0cm}>{\centering}p{1.0cm}>{\centering}p{1.0cm}>{\centering}p{1.0cm}>{\centering}p{1.0cm}|>{\centering}p{1.0cm}|>{\centering}p{1.0cm}|>{\centering}p{1.0cm}|}  \hline
\small Method & \small Car & \small Truck & \small Trailer & \small Motor & \small Bicycle & \small Rider & \small Person &\small mIoU &\small Acc \tabularnewline \hline 
{\small Label Prop.\cite{Vijayanarasimhan2012ECCV} }&28.5 & 53.2 & \textbf{71.7} & 11.1 & 42.0 & 32.2 & 21.8 & 37.2 & 59.1 \tabularnewline
{\small Sparse Track. + GC \cite{Sundaram2010ECCV} }&10.1 & 22.0 & 13.2 & 5.2 & 1.4 & 5.6 & 0.1 & 8.2 & 12.5 \tabularnewline
{\small 3D Prop. + GC }&15.7 & 40.3 & 24.1 & 2.2 & 0.2 & 2.0 & 17.1 & 14.5 & 21.7 \tabularnewline
{\small Proposed Method }&\textbf{77.7} & \textbf{85.1} & 69.8 & \textbf{60.5} & \textbf{42.9} & \textbf{49.5} & \textbf{59.0} & \textbf{63.5} & \textbf{94.1} \tabularnewline
\hline
\end{tabular}
\end{center}
\caption{{\bf Comparison to Label Transfer Baselines on Semantic Segmentation Transfer of Dynamic Objects}. We compare our method to 2D label transfer baselines (top) and to 3D to 2D label transfer baselines (bottom) on $120$ consecutive images that contain dynamic objects.}
\label{tab:baseline_dynamic_detailed}
\end{table*}

\begin{table*}[t]
\setlength{\tabcolsep}{2.65pt}
\begin{center}
\begin{tabular}{|p{3.5cm}|>{\centering}p{0.7cm}>{\centering}p{0.7cm}>{\centering}p{0.7cm}>{\centering}p{0.7cm}>{\centering}p{0.7cm}>{\centering}p{0.7cm}>{\centering}p{0.7cm}>{\centering}p{0.7cm}>{\centering}p{0.7cm}>{\centering}p{0.7cm}>{\centering}p{0.7cm}>{\centering}p{0.7cm}>{\centering}p{0.7cm}>{\centering}p{0.7cm}|>{\centering}p{0.7cm}|>{\centering}p{0.7cm}|}  \hline
\small Method & \small Road & \small Park & \small Sdwlk & \small Terr & \small Bldg & \small Vegt & \small Car & \small Trler & \small Crvn & \small Gate & \small Wall & \small Fence & \small Box & \small Sky & \small mIoU & \small Acc \tabularnewline \hline{\small LA }&95.4 & 34.0 & 48.5 & 66.6 & 88.4 & 82.8 & 91.8 & 54.2 & 88.9 & 61.7 & 53.8 & 52.0 & 53.6 & 89.6 & 68.7 & 89.1 \tabularnewline
{\small LA+3D }&95.3 & 48.9 & 75.7 & 66.6 & 85.8 & 81.5 & 92.1 & 56.1 & \textbf{91.2} & 68.3 & 58.8 & 45.7 & 48.4 & 88.7 & 71.7 & 90.1 \tabularnewline
{\small LA+PW }&95.4 & 26.5 & 38.3 & 66.2 & 88.5 & 83.5 & 92.0 & 56.9 & 89.4 & 66.2 & 51.6 & 50.0 & 51.2 & 89.8 & 67.5 & 88.2 \tabularnewline
{\small LA+PW+CO }&\textbf{95.5} & 70.0 & 82.5 & 67.4 & 89.8 & 83.7 & \textbf{92.4} & 87.2 & 89.0 & 75.4 & 57.9 & 68.3 & \textbf{60.0} & 89.9 & 79.2 & 92.5 \tabularnewline
{\small LA+PW+CO+3D }&95.1 & 72.7 & 84.0 & 67.3 & 90.3 & 84.1 & 92.2 & 93.1 & 90.8 & 77.3 & 63.0 & 72.1 & 56.7 & \textbf{92.8} & 80.8 & 93.0 \tabularnewline
{\small Full Model }&95.2 & \textbf{72.9} & \textbf{84.5} & \textbf{67.9} & \textbf{90.3} & \textbf{84.2} & 92.2 & \textbf{93.4} & 90.8 & \textbf{78.8} & \textbf{64.3} & \textbf{73.1} & 56.8 & 92.8 & \textbf{81.2} & \textbf{93.1} \tabularnewline
\hline
{\small Full Model (90\%) }&97.5 & 83.1 & 92.0 & 80.3 & 93.9 & 90.1 & 95.1 & 95.0 & 93.2 & 86.3 & 74.6 & 81.4 & 79.9 & 93.7 & 88.3 & 96.0 \tabularnewline
{\small Full Model (80\%) }&98.4 & \textbf{89.2} & 94.2 & 89.5 & 96.4 & 94.2 & 96.6 & 96.3 & 95.5 & 93.5 & 80.2 & 86.4 & 90.0 & 95.3 & 92.5 & 97.6 \tabularnewline
{\small Full Model (70\%) }&\textbf{98.8} & 88.8 & \textbf{95.0} & \textbf{92.2} & \textbf{97.7} & \textbf{95.8} & \textbf{97.3} & \textbf{97.1} & \textbf{96.7} & \textbf{96.1} & \textbf{85.2} & \textbf{88.2} & \textbf{95.0} & \textbf{96.3} & \textbf{94.3} & \textbf{98.4} \tabularnewline
\hline
\end{tabular}

\end{center}
\caption{{\bf Ablation Study on Semantic Segmentation of Static Objects.} This table shows the importance of the different components in our model on all $120$ images. The components are abbreviated as follows:
LA = local appearance ($p^{\cP}$), PW = 2D pairwise constraints ($\psi^{\cP,\cP}$), CO = 3D primitive constraints ($\xi^{\cP}$), 3D = 3D points ($\varphi^{\cL}$,$\psi^{\cP, \cL}$), Full Model = all potentials including 3D pairwise constraints ($\psi^{\cL,\cL}$). Percentages denote fractions of estimated pixels with highest confidence. }
\label{tab:ablation_semantic_detailed}
\end{table*}

\begin{table*}[t]
\setlength{\tabcolsep}{2.65pt}
\begin{center}
\begin{tabular}{|p{3.0cm}|>{\centering}p{0.7cm}>{\centering}p{0.7cm}>{\centering}p{0.7cm}>{\centering}p{0.7cm}>{\centering}p{0.7cm}|>{\centering}p{0.7cm}|>{\centering}p{0.7cm}|}  \hline
\small Method & \small Bldg & \small Car & \small Trler & \small Crvn & \small Box &\small mIoU & \small Acc \tabularnewline \hline 
{\small LA }&79.0 & 88.4 & 54.4 & 89.3 & 50.3 & 72.3 &86.7 \tabularnewline
{\small LA+3D }&77.6 & 89.6 & 56.3 & \textbf{91.5} & 49.7 & 72.9 &88.7 \tabularnewline
{\small LA+PW }&79.0 & 90.3 & 57.2 & 89.8 & 48.4 & 72.9 &85.8 \tabularnewline
{\small LA+PW+CO }&82.2 & \textbf{90.6} & 87.7 & 89.4 & 58.2 & 81.6 &91.0 \tabularnewline
{\small LA+PW+CO+3D }& \textbf{84.4} & \textbf{90.6} & 93.6 & 91.2 & 58.2 & 83.6 &91.7 \tabularnewline
{\small Full Model }&  \textbf{84.4} & \textbf{90.6} & \textbf{93.8} & 91.2 & \textbf{58.6} & \textbf{83.7} &\textbf{91.8} \tabularnewline
\hline
{\small Full Model (90\%) }&88.8 & 93.4 & 95.3 & 93.4 & 74.3 & 89.0 &94.9 \tabularnewline
{\small Full Model (80\%) }&92.0 & 94.8 & 96.4 & 95.6 & 77.8 & 91.3 &96.6 \tabularnewline
{\small Full Model (70\%) }&\textbf{93.6} & \textbf{95.7} & \textbf{97.2} & \textbf{96.8} & \textbf{79.9} & \textbf{92.7} &\textbf{97.4} \tabularnewline
\hline
\end{tabular}

\end{center}
\caption{{\bf Comparison to Label Transfer Baselines on Instance Segmentation on Static Objects}. We compare our method to 2D label transfer baselines (top) and to 3D to 2D label transfer baselines (bottom) on $120$ consecutive images. }
\label{tab:ablation_instance_detailed}
\end{table*}

\section{Dataset}

\subsection{Statistical Analysis}
\label{app:statistics}
\begin{figure}[t]
\center
\begin{subfigure}{\linewidth}
\includegraphics[width=\linewidth]{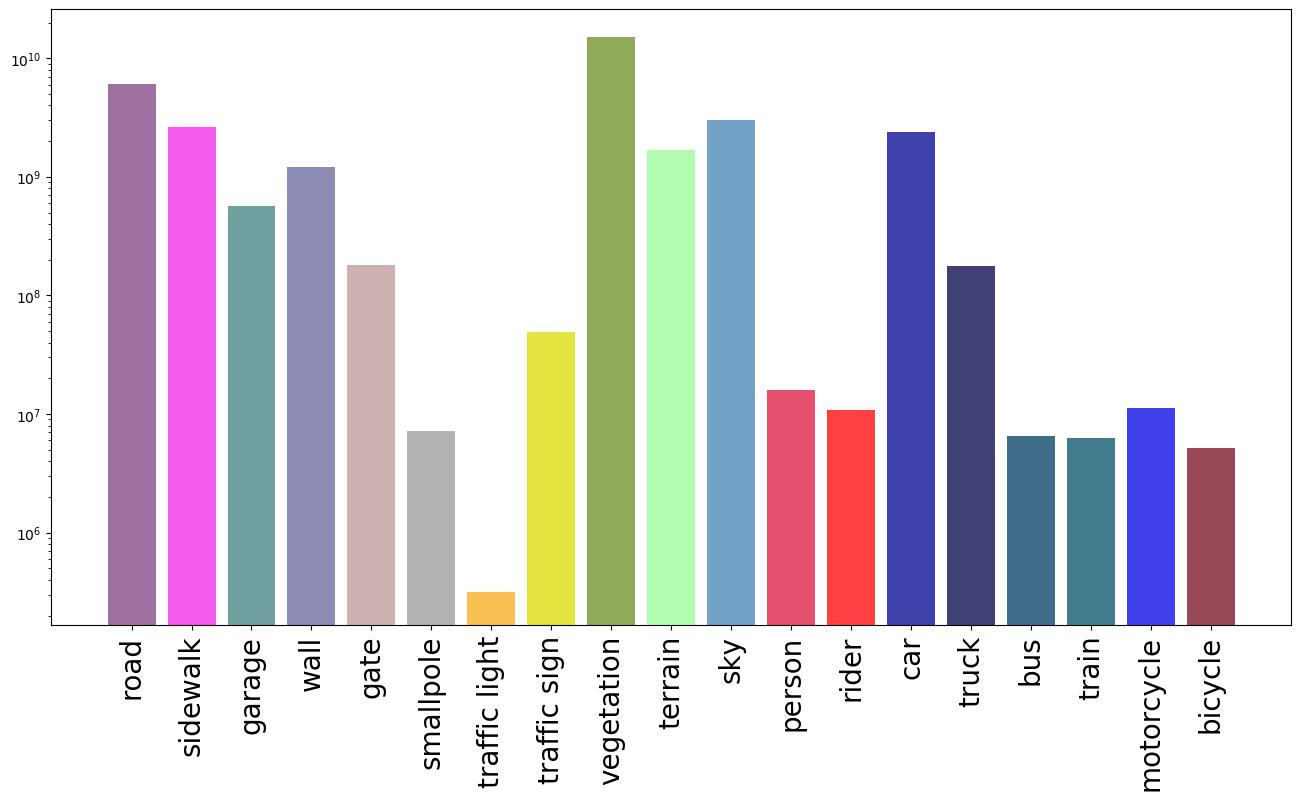}
\caption{Distribution of 2D semantic labels over 78k frames.} 
\label{fig:stats_2d}
\end{subfigure}
\begin{subfigure}{\linewidth}
\includegraphics[width=\linewidth]{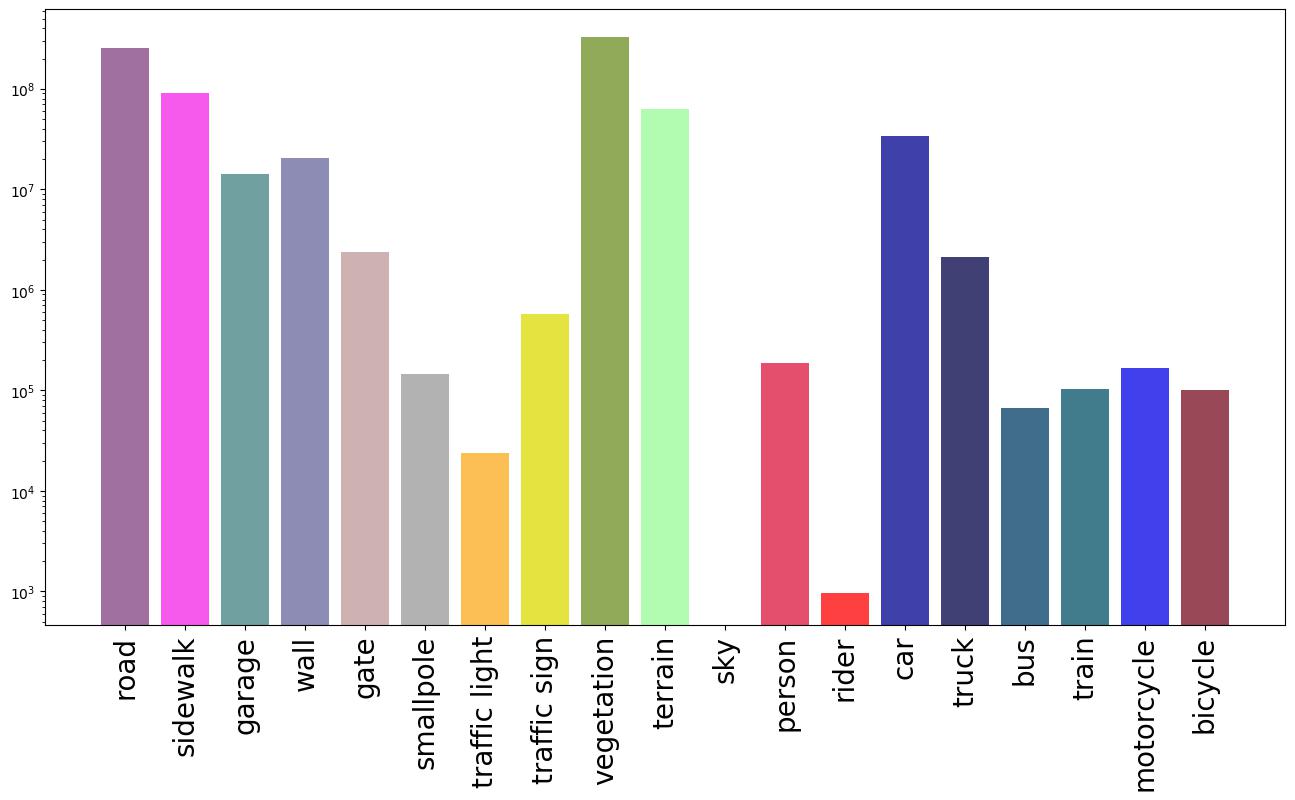}
\caption{Distribution of 3D semantic labels over 1B points.} 
\label{fig:stats_3dpcd}
\end{subfigure}
\begin{subfigure}{\linewidth}
\includegraphics[width=\linewidth]{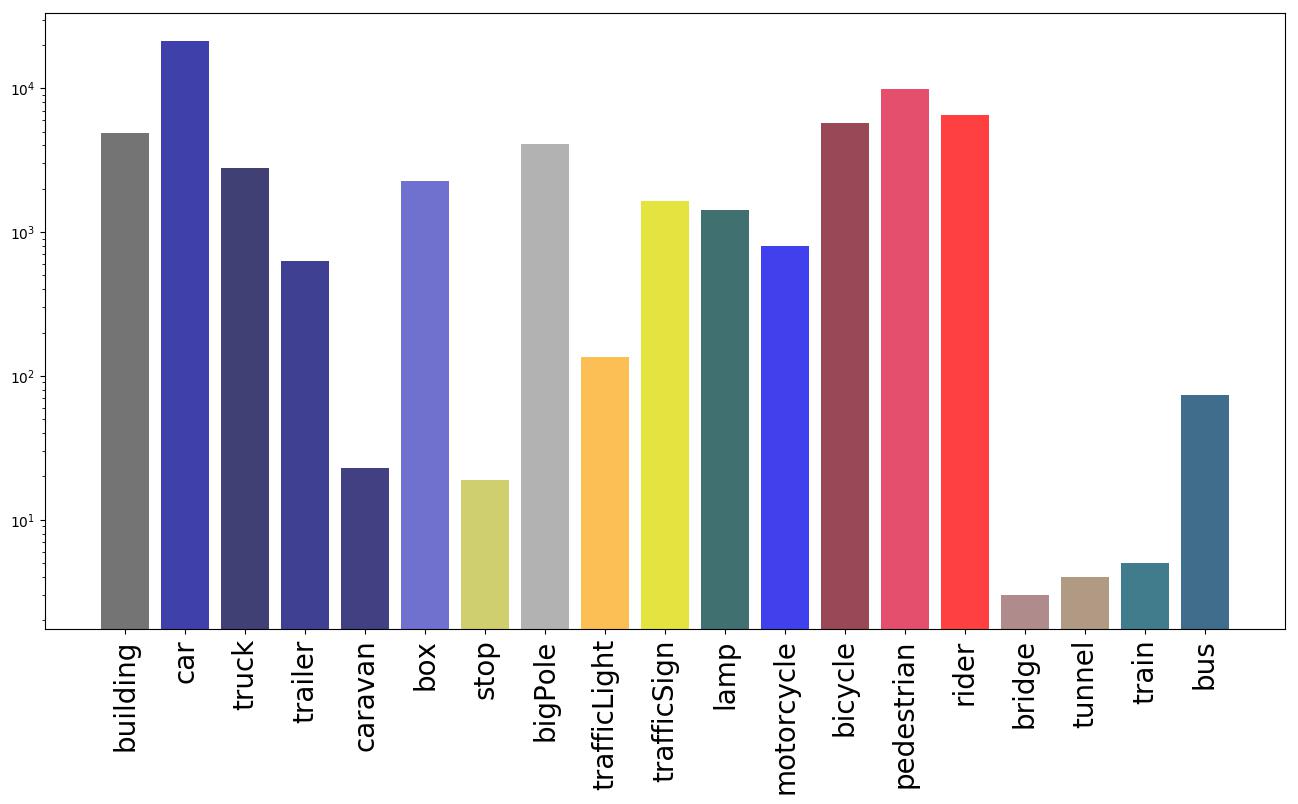}
\caption{Distribution of 3D semantic labels over 68k bounding boxes.} 
\label{fig:stats_3dbox}
\end{subfigure}
\caption{{\bf Dataset Statistics.} (\subref{fig:stats_2d}) and (\subref{fig:stats_3dpcd}) show the histogram of the 19 training semantic classes in 2D and 3D, respectively. (\subref{fig:stats_3dbox}) shows the class distribution of 3D bounding boxes that contains instance IDs.  
}
\label{fig:stats}
\end{figure}

\figref{fig:stats} shows the distribution of the semantic labels in KITTI-360. \figref{fig:stats_2d} and \figref{fig:stats_3dpcd} suggests that the semantic distribution of the 2D pixels and the 3D points are similar (except for the ``Sky'' class). We also show the distribution of our 3D bounding boxes in \figref{fig:stats_3dbox}.

\subsection{Dataset Split}
\label{app:split}
\begin{figure}[t]
\center
\includegraphics[width=\linewidth]{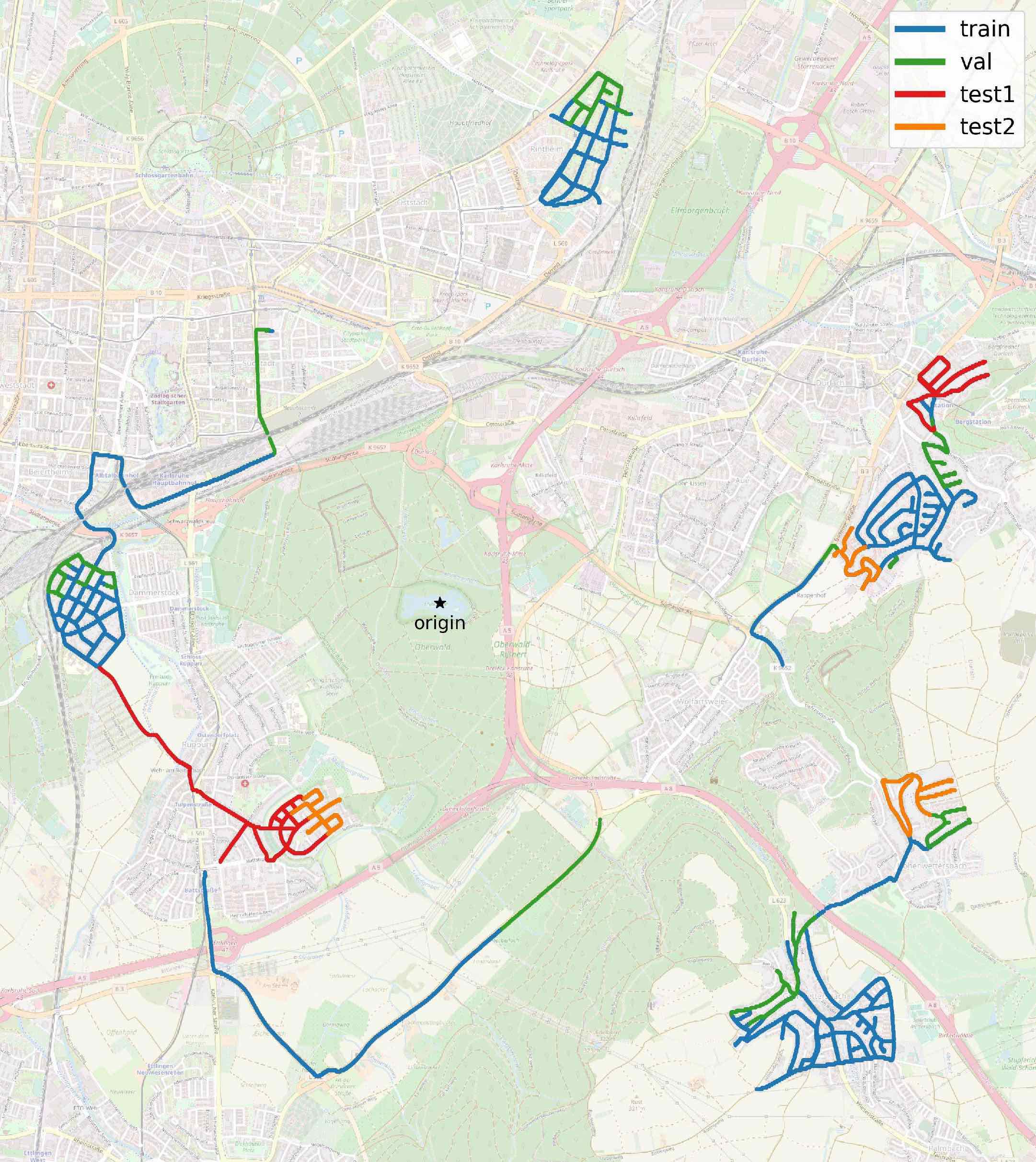}
\caption{{\bf Dataset Split.} We split the dataset into one training set, one validation set, and two test sets with held-out ground truth.
}
\label{fig:split}
\end{figure}

We split KITTI-360 into training and test sets without spatial overlapping as shown in \figref{fig:split}. We maintain an online evaluation server and hold back the labels of the test set. Considering that different tasks involve different label modalities, the test set is further divided into two parts with different information released. Specifically, the first part of the test dataset is used for semantic scene understanding (except for semantic scene completion) and novel view synthesis, where we hold back the 2D semantic/instance segmentation maps and 3D pointwise labels. Note that the accumulated point clouds are released while their labels are removed. The other part is adopted for semantic scene completion and semantic SLAM where the accumulated point clouds are further removed. We release vehicle poses for both test sets.

\section{Semantic Scene Understanding Benchmark}
\subsection{Benchmark of 2D Semantic segmentation}
\label{app:benchmark_2d_semantic}
\subsubsection{Evaluation Metrics}
We evaluate confidence weighted mIoU where both the intersection and the union are weighted (per-pixel) by the confidence of our pseudo-ground truth. More formally, let $\{\text{TP}\}$ and $\{\text{TP, FP, FN}\}$ denote the set of image pixels in the intersection and the union of one class label (or one category label), respectively. The weighted IoU of this class can be defined as follow:
\begin{equation}
\text{IoU} = \frac{ \sum_{i\in{\{\text{TP}\}}}c_{i} }{ \sum_{i\in{\{\text{TP, FP, FN}\}}}c_{i}}
\label{eq:weighted_iou}
\end{equation}
where $c_i \in [0, 1]$ denotes the confidence value at pixel $i$. In the standard evaluation $c_i=1$ for all pixels. The mIoU is then calculated as the mean of the weighted IoU over all class labels or category labels.

While we provide 19 classes for training following Cityscapes~\cite{Cordts2016CVPR}, we omit two classes, ``Train'' and ``Bus'' during evaluation since these two classes are rarely observed in the test region when we split the training and test sets according to the camera poses as shown in \figref{fig:split}.

\subsubsection{Baselines}
We train and evaluate two well-known methods, Fully Convolutional Neural Network (FCN)~\cite{Long2015CVPR} and Pyramid Scene Parsing Network (PSPNet)~\cite{Zhao2017CVPR}. For FCN, we adopt the ResNet-101 model provided by PyTorch as a backbone. The model is pre-trained on a subset of the Microsoft COCO dataset. As for PSPNet, we use the official PyTorch implementation\footnote{\url{https://github.com/hszhao/semseg}} which also uses ResNet-101 as backbone. The model is pre-trained on the ImageNet dataset.

\subsubsection{Additional Results}
We show the IoU of each class in \tabref{tab:benchmark_2d_semantic_app}. We observe that PSPNet consistently outperforms FCN in most of the classes.

\subsection{Benchmark of 2D Instance Segmentation}
\label{app:benchmark_2d_instance}
\subsubsection{Evaluation Metric}
Following~\cite{Lin2014ECCV}, we measure the Average Precision (AP) over 10 IoU thresholds, ranging from 0.5 to 0.95 with a step size of 0.05. We calculate confidence weighted IoU per \textit{instance} using \eqref{eq:weighted_iou}. In this task, we consider 7 classes that contain instance labels, including ``Building'', ``Person'', ``Rider'', ``Car'', ``Truck'', ``Motorcycle'' and ``Bicycle''.
\subsubsection{Baselines}
We evaluated two Mask R-CNN models with different backbones, i.e., ResNet-50 and ResNet-101, based on the official implementation\footnote{\url{https://github.com/facebookresearch/detectron2}}. Both backbones are pre-trained on the ImageNet dataset.
\subsubsection{Additional Results}
\tabref{tab:benchmark_2d_instance_app} shows the AP of each individual class as well as the mean AP. We observe that performance of different backbones is similar in more frequently observed classes (\eg, ``Building'' and ``Car'') while differs in less occurred classes.

\begin{table*}[t!]
	\setlength{\tabcolsep}{0.2pt}
    \begin{subtable}[h]{\textwidth}
		\begin{center}
			\begin{tabular}{|p{2.00cm}|>{\centering}p{0.80cm}>{\centering}p{0.80cm}>{\centering}p{0.80cm}>{\centering}p{0.80cm}>{\centering}p{0.80cm}>{\centering}p{0.80cm}>{\centering}p{0.80cm}>{\centering}p{0.80cm}>{\centering}p{0.80cm}>{\centering}p{0.80cm}>{\centering}p{0.80cm}>{\centering}p{0.80cm}>{\centering}p{0.80cm}>{\centering}p{0.80cm}>{\centering}p{0.80cm}>{\centering}p{0.80cm}>{\centering}p{0.80cm}|>{\centering}p{1.60cm}|}  \hline
Method & Road & Sdwlk & Bldg & Wall & Fence & Pole & Trlgt & Trsgn & Vegt & Terr & Sky & Persn & Rider & Car & Truck & Motor & Bicyc & mIoU$_\text{class}$ \tabularnewline
\hline
FCN~\cite{Long2015CVPR} & 95.6 &84.5 &84.1 &43.4 &38.6 &31.1 &0.0 &38.0 &90.6 &85.7 &91.2 &40.5 &29.3 &94.6 &42.4 &28.4 &0.0 &54.0  \tabularnewline
PSPNet~\cite{Zhao2017CVPR} & 96.6 &87.3 &87.0 &65.0 &55.6 &40.1 &0.0 &43.0 &92.6 &88.4 &91.9 &55.5 &48.3 &95.6 &60.4 &52.1 &44.1 &64.9  \tabularnewline
\hline
\end{tabular}
		\end{center}
		\vspace{-0.3cm}
		\caption{2D Semantic Segmentation}
		\label{tab:benchmark_2d_semantic_app}
		\vspace{0.1cm}
	\end{subtable}\\
    \begin{subtable}[h]{\textwidth}
		\begin{center}
			\begin{tabular}{|p{2.50cm}|>{\centering}p{1.20cm}>{\centering}p{1.20cm}>{\centering}p{1.20cm}>{\centering}p{1.20cm}>{\centering}p{1.20cm}>{\centering}p{1.20cm}>{\centering}p{1.20cm}|>{\centering}p{1.20cm}|}  \hline
Method & Bldg & Persn & Rider & Car & Truck & Motor & Bicyc & AP \tabularnewline
\hline
ResNet50 & 26.5 &27.2 &10.9 &53.2 &6.2 &5.3 &7.3 &19.5  \tabularnewline
ResNet101 & 27.0 &22.9 &15.9 &52.0 &10.2 &12.7 &5.7 &20.9  \tabularnewline
\hline
\end{tabular}
		\end{center}
		\vspace{-0.3cm}
		\caption{2D Instance Segmentation}
		\vspace{0.1cm}
		\label{tab:benchmark_2d_instance_app}
	\end{subtable}\\
    \begin{subtable}[h]{\textwidth}
		\begin{center}
			\begin{tabular}{|p{2.5cm}|>{\centering}p{1.5cm}>{\centering}p{1.5cm}|>{\centering}p{1.5cm}||>{\centering}p{1.5cm}>{\centering}p{1.5cm}|>{\centering}p{1.5cm}|}  \hline
Method & Bldg & Car & AP$_{50}$ & Bldg & Car & AP$_{25}$ \tabularnewline 
\hline
BoxNet~\cite{Qi2019ICCV} & 8.2 & 0.0 & 4.1 & 46.6 & 0.6 & 23.6  \tabularnewline 
VoteNet~\cite{Qi2019ICCV} & 5.7 & 1.1 & 3.4 & 40.3 & 20.9 & 30.6 \tabularnewline 
\hline 
\end{tabular}
		\end{center}
		\vspace{-0.2cm}
		\caption{3D Bounding Box Detection}
		\vspace{0.1cm}
		\label{tab:benchmark_3d_bbox_app}
	\end{subtable}\\
    \begin{subtable}[h]{\textwidth}
		\begin{center}
			\begin{tabular}{|p{2.00cm}|>{\centering}p{0.80cm}>{\centering}p{0.80cm}>{\centering}p{0.80cm}>{\centering}p{0.80cm}>{\centering}p{0.80cm}>{\centering}p{0.80cm}>{\centering}p{0.80cm}>{\centering}p{0.80cm}>{\centering}p{0.80cm}>{\centering}p{0.80cm}>{\centering}p{0.80cm}>{\centering}p{0.80cm}>{\centering}p{0.80cm}>{\centering}p{0.80cm}>{\centering}p{0.80cm}>{\centering}p{0.80cm}>{\centering}p{0.80cm}|>{\centering}p{1.60cm}|}  \hline
Method & Road & Sdwlk & Bldg & Wall & Fence & Pole & Trlgt & Trsgn & Vegt & Terr & Sky & Persn & Rider & Car & Truck & Motor & Bicyc & mIoU$_\text{class}$ \tabularnewline
\hline
PointNet~\cite{Qi2017CVPR} & 54.2 &14.4 &28.1 &2.7 &1.4 &0.3 &0.0 &2.2 &56.5 &16.4 &-- &0.0 &-- &19.9 &0.0 &0.0 &0.0 &13.1  \tabularnewline
PointNet++~\cite{Qi2017NIPS}  & 82.1 &66.3 &62.1 &30.5 &24.9 &38.3 &0.0 &23.4 &71.2 &47.3 &-- &2.0 &-- &84.8 &0.5 &1.6 &0.0 &35.7  \tabularnewline
\hline
\end{tabular}
		\end{center}
		\vspace{-0.2cm}
		\caption{3D Semantic Segmentation}
		\vspace{0.1cm}
		\label{tab:benchmark_3d_semantic_app}
	\end{subtable}\\
    \begin{subtable}[h]{\textwidth}
		\begin{center}
			\begin{tabular}{|p{3.50cm}|>{\centering}p{1.50cm}>{\centering}p{1.50cm}|>{\centering}p{1.50cm}|}  \hline
Method & Bldg & Car & AP \tabularnewline
\hline
PointNet++~\cite{Qi2017NIPS}+~\cite{Ester1996} & 11.5 &35.9 &23.7 \tabularnewline
PointGroup~\cite{JiangCVPR2020} & 9.9 &59.6 &34.8 \tabularnewline
\hline
\end{tabular}
		\end{center}
		\vspace{-0.2cm}
		\caption{3D Instance Segmentation}
		\vspace{0.1cm}
		\label{tab:benchmark_3d_instance_app}
	\end{subtable}\\
    \begin{subtable}[h]{\textwidth}
		\begin{center}
			\begin{tabular}{|p{2.00cm}|>{\centering}p{0.80cm}>{\centering}p{0.80cm}>{\centering}p{0.80cm}>{\centering}p{0.80cm}>{\centering}p{0.80cm}>{\centering}p{0.80cm}>{\centering}p{0.80cm}>{\centering}p{0.80cm}>{\centering}p{0.80cm}>{\centering}p{0.80cm}>{\centering}p{0.80cm}>{\centering}p{0.80cm}>{\centering}p{0.80cm}>{\centering}p{0.80cm}>{\centering}p{0.80cm}>{\centering}p{0.80cm}>{\centering}p{0.80cm}|>{\centering}p{1.60cm}|}  \hline
Method & Road & Sdwlk & Bldg & Wall & Fence & Pole & Trlgt & Trsgn & Vegt & Terr & Sky & Persn & Rider & Car & Truck & Motor & Bicyc & mIoU$_\text{class}$ \tabularnewline
\hline
Enc-Dec & 42.0 &16.9 &16.2 &2.3 &0.1 &4.7 &0.0 &2.6 &25.4 &9.0 &-- &0.0 &-- &16.2 &0.0 &0.0 &0.0 &9.1  \tabularnewline
\hline
\end{tabular}
		\end{center}
		\vspace{-0.2cm}
		\caption{Semantic Scene Completion}
		\vspace{0.1cm}
		\label{tab:benchmark_scene_complt_app}
	\end{subtable}
	\vspace{-0.3cm}
	\caption{{\bf Additional Quantitative Results for Semantic Scene Understanding.} In each table, we show the performance of individual classes with the overall metric in the last column.}
\end{table*}

\subsection{Benchmark of 3D Bounding Box Detection}
\label{app:benchmark_3d_instance}
\subsubsection{Evaluation Metric}
We evaluate AP at a threshold of 0.5 and 0.25 for 3D bounding box detection. As it is particularly challenging for learning-based algorithms to generalize well to other classes with fewer training samples, we measure the mean AP over two classes: ``Building'' and ``Car''.
\subsubsection{Baselines}
We evaluate the state-of-the-art 3D bounding box detection method, VoteNet~\cite{Qi2019ICCV}, and its simplified version, BoxNet~\cite{Qi2019ICCV} as baselines. We adopt the official implementation~\footnote{\url{https://github.com/facebookresearch/votenet}} for both methods. 
\subsubsection{Additional Results}
\tabref{tab:benchmark_3d_bbox_app} shows the AP on each class as well as the mean AP. Both methods achieve reasonable performance at the IoU threshold of 0.25 while struggle at the higher threshold. 

\subsection{Benchmark of 3D Semantic Segmentation}
\label{app:benchmark_3d_semantic}
\subsubsection{Evaluation Metric}
For 3D semantic segmentation, we also evaluate confidence weighted mIoU using \eqref{eq:weighted_iou}. Here, the confidence of each 3D point is obtained by averaging the confidence of 3D points on multiple frames, as introduced in \appref{app:inference_accumulation}. Similar to 2D semantic segmentation, we omit ``Train'' and ``Bus'' during evaluation. Note that there is no ``Sky'' point in 3D, thus it is also discarded. Moreover, since we train and evaluate only in static regions, we ignore ``Rider'' as it only appears as a dynamic object. 
\subsubsection{Baselines}
We train and evaluate two baselines, PointNet~\cite{Qi2017CVPR} and PointNet++~\cite{Qi2017NIPS}. As the original implementations are built on Tensorflow, we adopt a faithful Pytorch reimplementation\footnote{\url{https://github.com/yanx27/Pointnet_Pointnet2_pytorch}} that contains both methods.
\subsubsection{Additional Results}
\tabref{tab:benchmark_3d_semantic_app} shows the detailed results of 3D semantic segmentation. As expected, PointNet++ achieves better performance on all classes compared to PointNet.

\subsection{Benchmark of 3D Instance Segmentation}
\label{app:benchmark_3d_instance}
\subsubsection{Evaluation Metric}
In 3D instance segmentation, we also evaluate AP over 10 IoU thresholds, ranging from 0.5 to 0.95 with a step size of 0.05. The IoU of each 3D instance is weighted (per-point) by the confidence of our pseudo-ground truth. Here, we evaluate on ``Building'' and ``Car'' the same as the 3D box bounding detection benchmark.
\subsubsection{Baselines}
We first consider a na\"ive baseline based on the results we obtained from 3D semantic segmentation using PointNet++~\cite{Qi2017NIPS}. Specifically, we first extract points of the same class label (``Building'' or ``Car'') based on the semantic segmentation results. Next, we group the extracted point cloud using DBSCAN~\cite{Ester1996}\footnote{\url{http://www.open3d.org/docs/0.12.0/python_api/open3d.geometry.PointCloud.html}}, where clusters with less than 500 points are ignored. Each valid cluster is then considered as an instance with a confidence score of $1.0$. The second baseline is a state-of-the-art approach, PointGroup~\cite{JiangCVPR2020}. We follow the official implementation\footnote{\url{https://github.com/dvlab-research/PointGroup}} to train and evaluate this baseline. 
\subsubsection{Additional Results}
\tabref{tab:benchmark_3d_instance_app} shows the detailed results of 3D instance segmentation. Interestingly, the 3D instance segmentation performance of ``Car'' is higher than the 2D baselines in \tabref{tab:benchmark_2d_instance_app}. We hypothesize that unlike in 2D where occlusions strongly impact the results, cars can be more easily separated in 3D.

\subsection{Benchmark of Semantic Scene Completion}
\label{app:benchmark_scene_complt}
\subsubsection{Data Preparation}
\begin{figure}[t]
\center
\begin{subfigure}{0.49\linewidth}
\includegraphics[width=\linewidth]{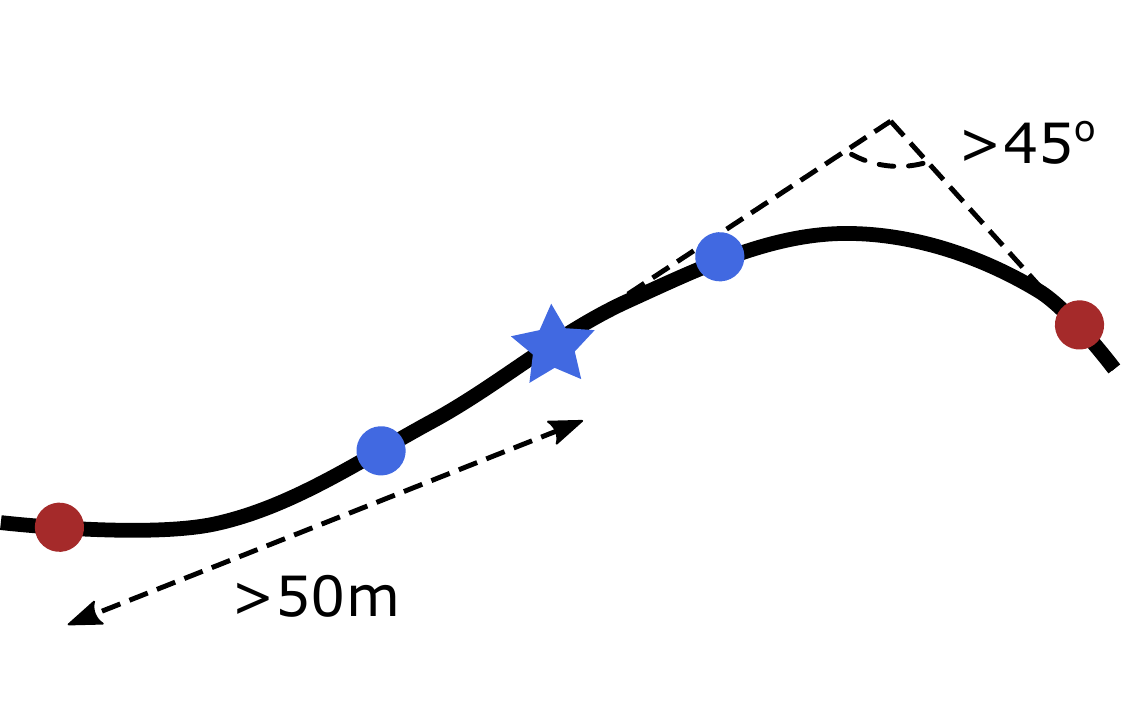}
\caption{Valid neighbor points} 
\label{fig:scene_complt_gt_p1}
\end{subfigure}
\begin{subfigure}{0.49\linewidth}
\includegraphics[width=\linewidth]{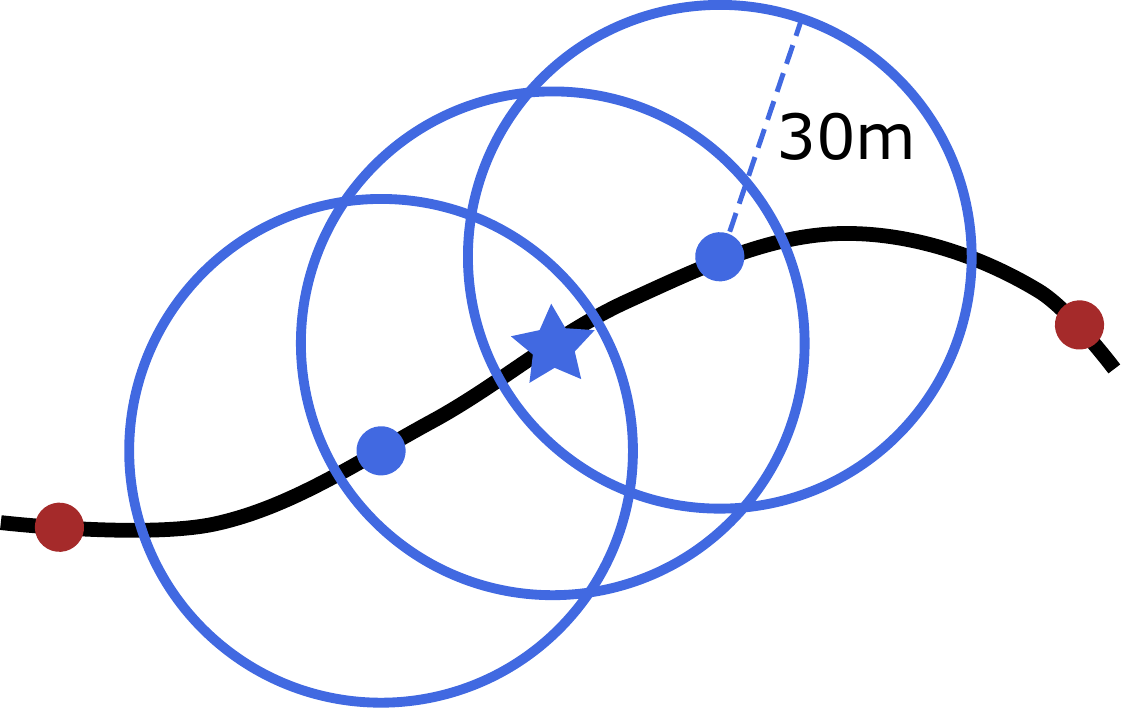}
\caption{Crop point cloud} 
\label{fig:scene_complt_gt_p2}
\end{subfigure}
\caption{{\bf Ground Truth Generation for Scene Completion Benchmark.} The blue star denotes the center of the input laser scan. The blue and red dots denote valid and invalid neighbors along the corridor, respectively.
}
\label{fig:scene_complt_gt}
\end{figure}

The ground truth of the semantic scene completion task is the accumulated point cloud within a corridor of 30m around the vehicle poses of a 100m trajectory (50m in each direction), see \figref{fig:scene_complt_gt} for an illustration. The input to this task is a single LiDAR scan whose center is visualized by the blue star point. We first determine a set of neighboring vehicle poses close to the given center illustrated in  \figref{fig:scene_complt_gt_p1}, and then crop the accumulated point cloud using the union of circles located at those poses as shown in \figref{fig:scene_complt_gt_p2}. To avoid evaluating in significantly occluded regions that typically occur when the vehicle turns a large angle, we also check the orientation of each pose as shown in \figref{fig:scene_complt_gt_p1}. Specifically, if the forward direction of one pose deviates more than $45^\circ$ compared to the heading angle of the given center, it is eliminated from the neighboring poses.

\subsubsection{Evaluation Metric}
In this task we evaluate  geometric completion and semantic estimation, respectively. Geometric completion is evaluated via completeness and accuracy at a threshold of 20cm.
Completeness is calculated as the fraction of ground truth points of which the distances to their closest reconstructed points are below the threshold. Accuracy instead measures the percentage of reconstructed points that are within a distance threshold to the ground truth points. 
As our ground truth reconstruction may not be complete, we prevent punishing reconstructed points by dividing the space into observed and unobserved regions, which are determined by the unobserved volume from a 3D occupancy map obtained using OctoMap~\cite{Hornung2013AR}. A reconstructed point is only evaluated when it falls into the observed region within the union of the neighboring circles shown in \figref{fig:scene_complt_gt_p2}. We further measure the F$_1$ score as the harmonic mean of the completeness and the accuracy. Note that SemanticKITTI~\cite{Behley2019ICCV} also considers a semantic scene completion task, but considers voxel as representation and measures mIoU over voxels for both reconstruction and semantics. We instead avoid discretization and directly evaluate on point clouds using standard metrics to separately assess accuracy and completeness.
\subsubsection{Baselines}
We implement two baselines for this task. For calibration, the first baseline returns the input LiDAR scan as output. The second baseline is a learning-based approach that adopts an encoder-decoder structure. Specifically, the encoder first learns features from the input point cloud. It then merges the point-wise features to voxels such that a 3D U-Net is applied to predict a volumetric reconstruction. The network is trained using a cross-entropy loss where the ground truth point cloud is also discretized into a volume. As our evaluation server requires submission in the form of point clouds, we uniformly and densely sample points from each occupied voxel as the final output. 
\subsubsection{Additional Results}
\tabref{tab:benchmark_scene_complt_app} shows detailed semantic estimation performance of the learning-based baseline. As can be seen, it is challenging to predict the geometry and the semantics jointly. The overall performance of this baseline is worse compared to baselines that directly perform 3D semantic segmentation in \tabref{tab:benchmark_3d_semantic_app}.

\section{Novel View Synthesis Benchmark}
\label{app:benchmark_nvs}

\begin{figure*}[t!]
\centering
\setlength{\tabcolsep}{3.0pt}
\begin{tabular}{P{0.6em}P{35em}P{16em}}
    \rotatebox{90}{GT Image} &
    \includegraphics[width=\linewidth]{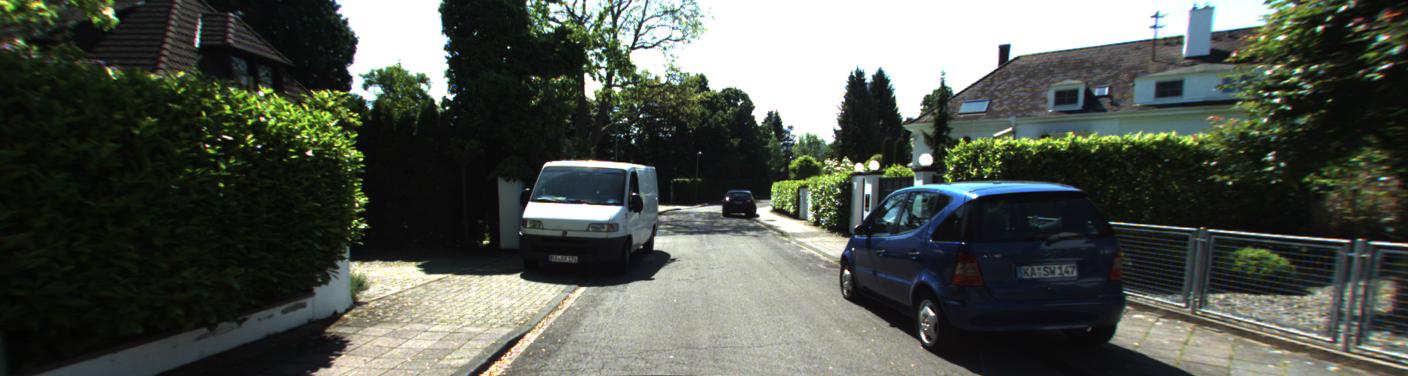}  &
    \includegraphics[width=\linewidth]{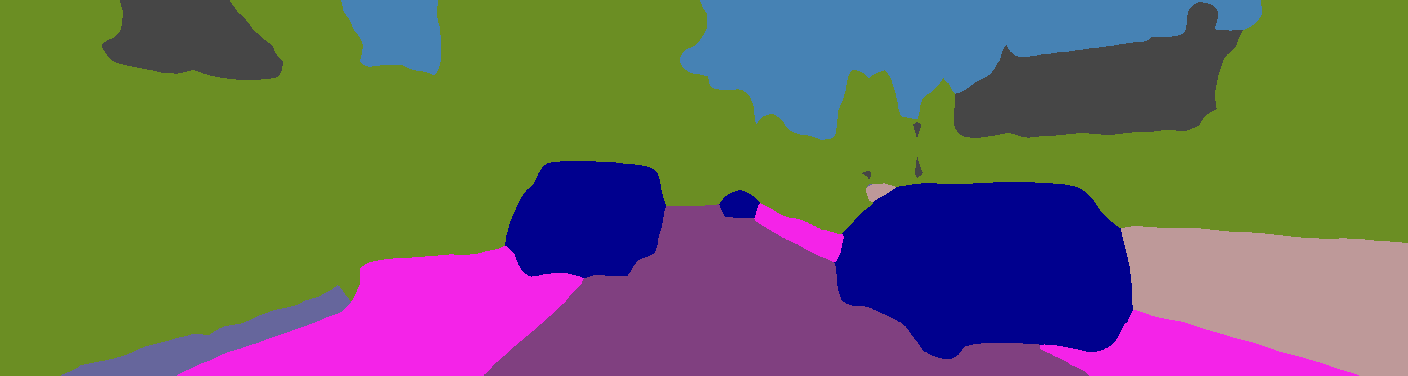} \\
    \rotatebox{90}{PCL} &
    \includegraphics[width=\linewidth]{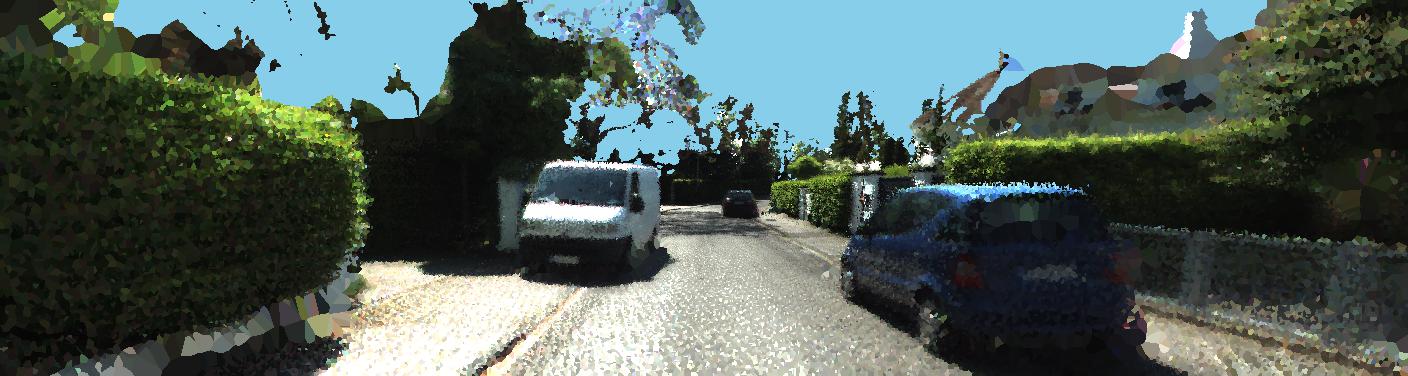} &
    \includegraphics[width=\linewidth]{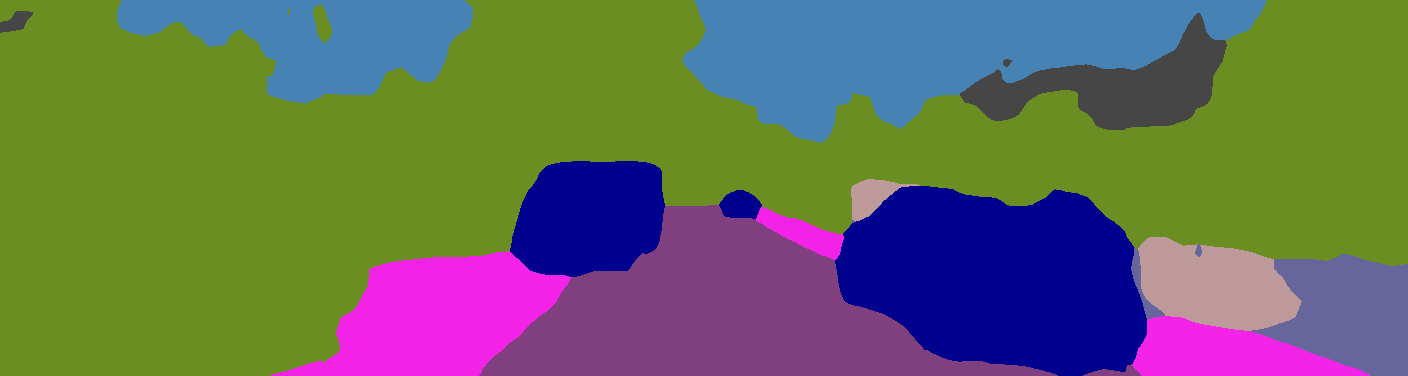}  \\
    \rotatebox{90}{NeRF~\cite{Mildenhall2020ECCV}} &
    \includegraphics[width=\linewidth]{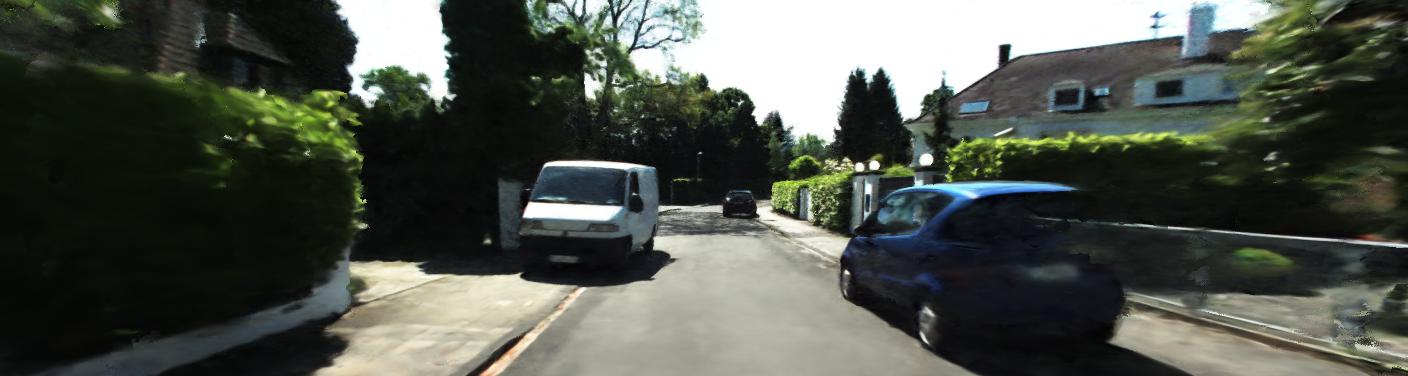} &
    \includegraphics[width=\linewidth]{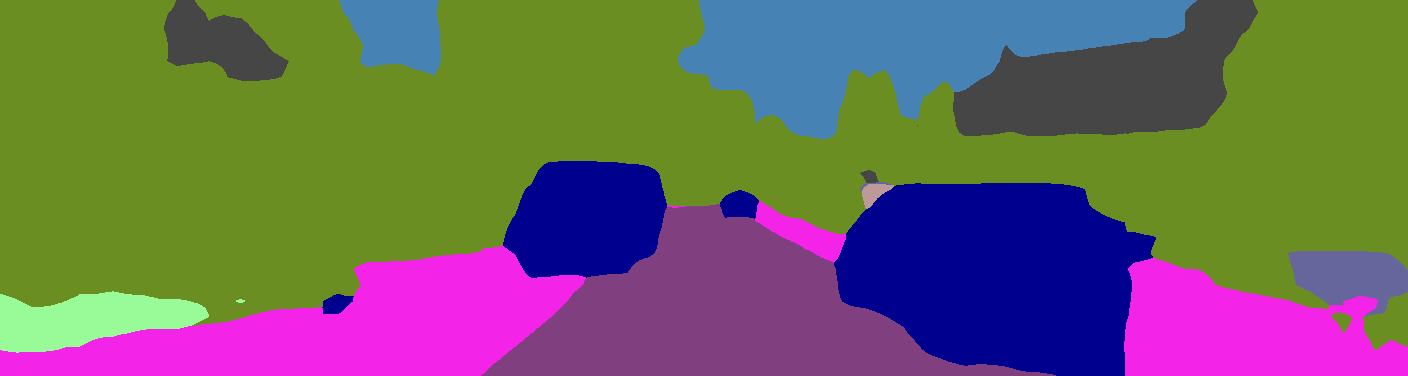} \\
    \rotatebox{90}{mip-NeRF~\cite{Barron2021ICCV}} &
    \includegraphics[width=\linewidth]{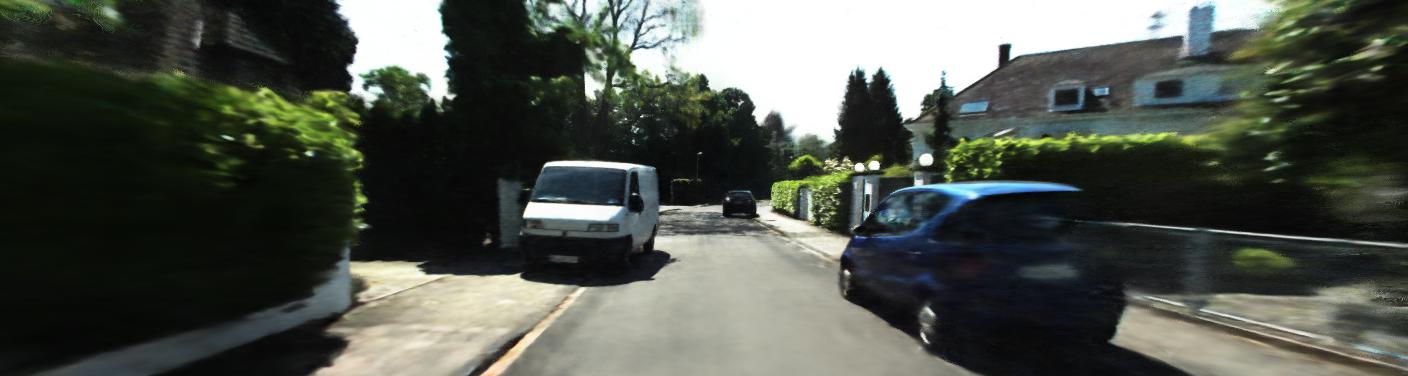} &
    \includegraphics[width=\linewidth]{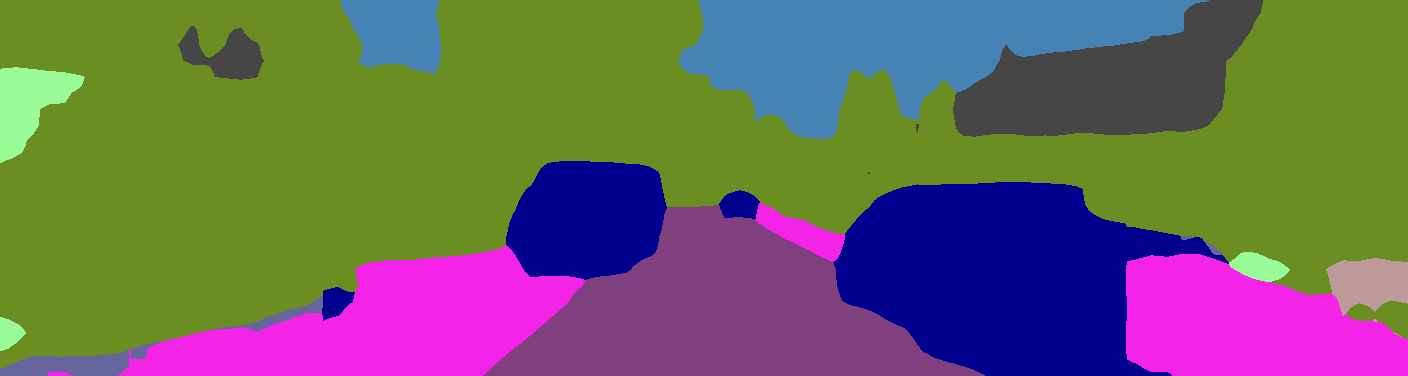} \\
    \rotatebox{90}{DS-NeRF~\cite{Deng2021ARXIV}} &
    \includegraphics[width=\linewidth]{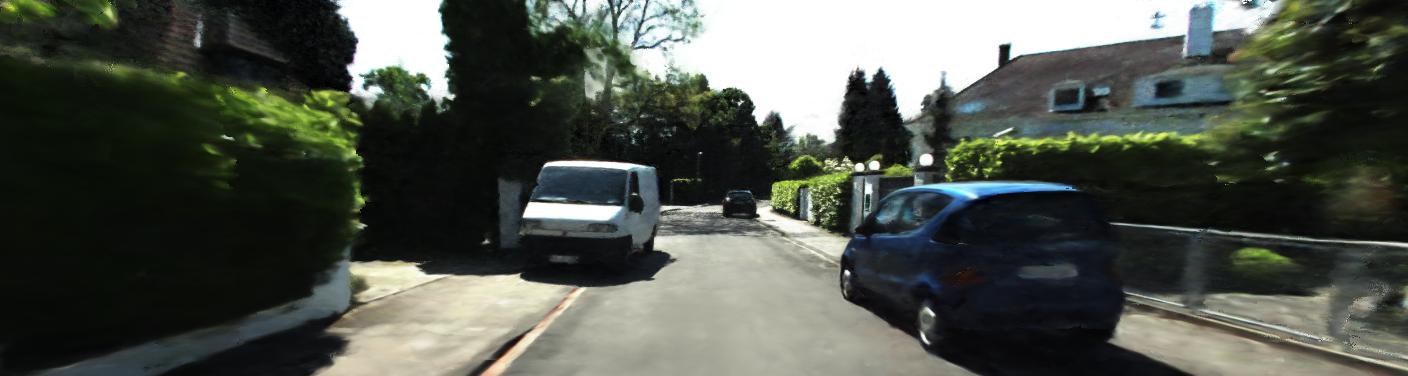} &
    \includegraphics[width=\linewidth]{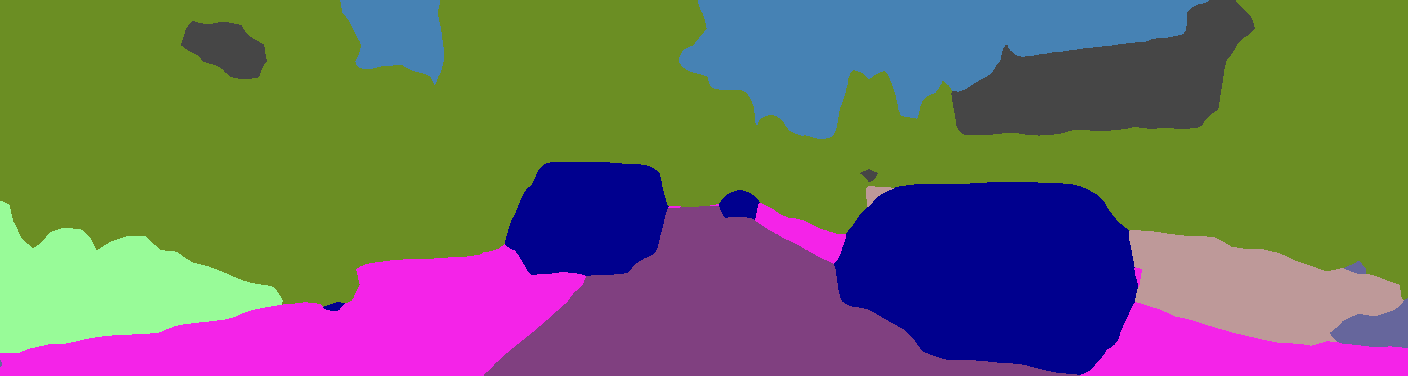} \\ 
    \rotatebox{90}{FVS~\cite{Riegler2020ECCV}} &
    \includegraphics[width=\linewidth]{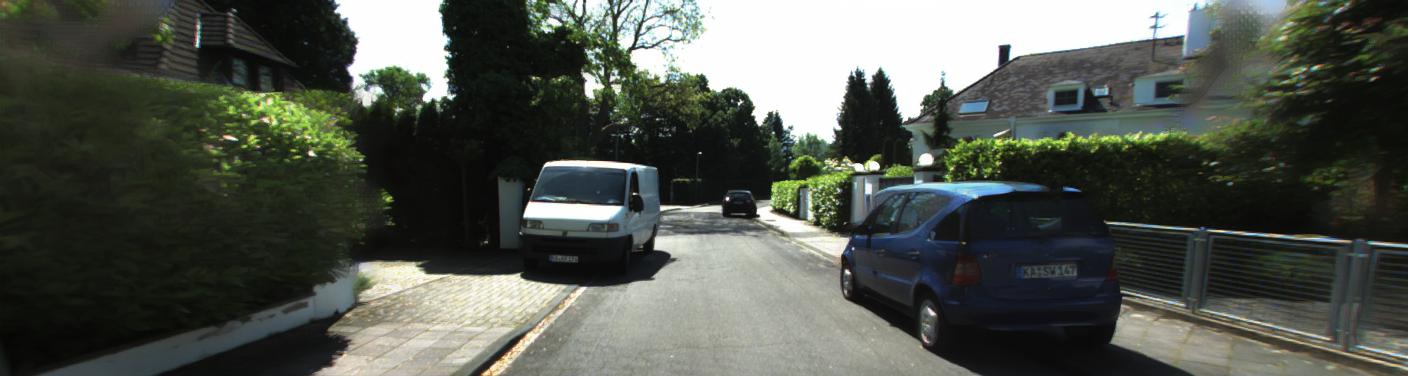} &
    \includegraphics[width=\linewidth]{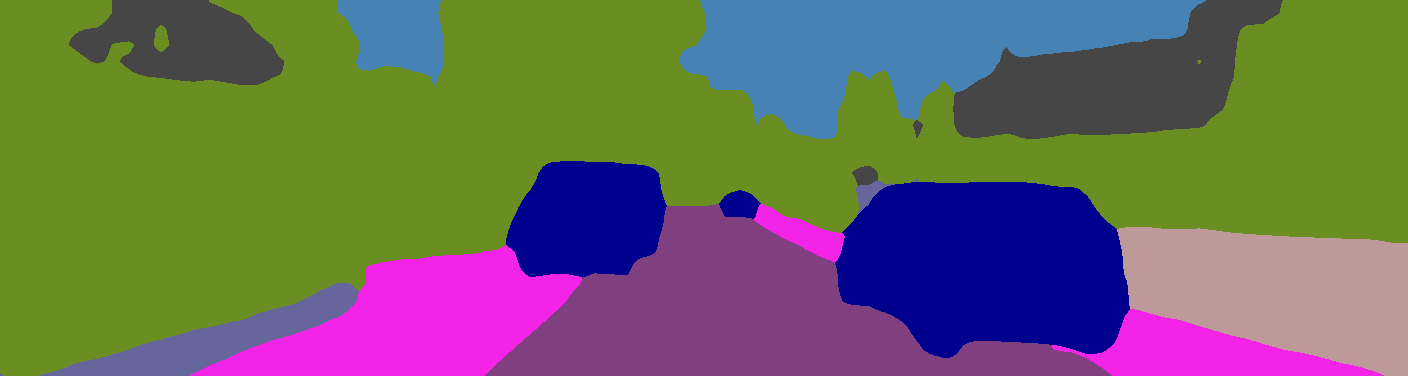} \\
    \rotatebox{90}{PBNR~\cite{Kopanas2021CGF}} &
    \includegraphics[width=\linewidth]{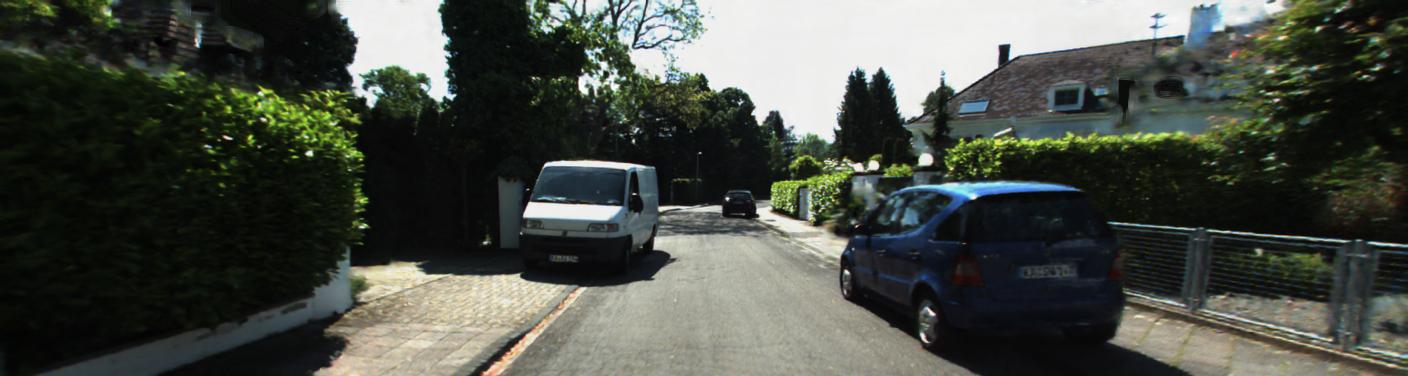} &
    \includegraphics[width=\linewidth]{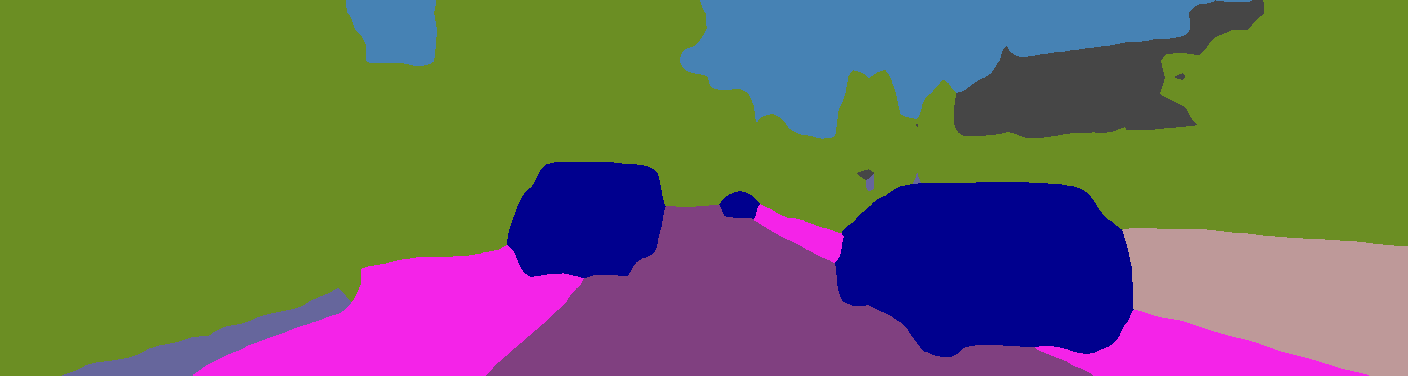} \\
    \end{tabular}
    \caption{{\bf Additional Qualitative Results for Novel View Appearance \& Semantic Synthesis.} The left column row shows the GT image and novel view appearance synthesis results. The right column shows the corresponding semantic segmentation using PSPNet~\cite{Zhao2017CVPR}.}
\label{fig:benchmark_nvs_label_app}
\end{figure*}

\subsection{Benchmark of Novel View Appearance Synthesis}
\subsubsection{Evaluation Metric}
We adopt three standard metrics to evaluate novel view appearance synthesis: peak signal-to-noise ratio (PSNR), and structural similarity index (SSIM), and perceptual metric (LPIPS) \cite{Zhang2018CVPR2}.
\subsubsection{Data Preparation}
We select $5$ static scenes with a driving distance of $\sim 50$ meters each for evaluating NVS at a $50\%$ drop rate. We select one frame every $\sim$ 0.8 meters driving distance (corresponding to the overall average distance between frames) to avoid redundancy when the vehicle is slow. We release $50\%$ of the frames for training and retain $50\%$ for evaluation. Moreover, we select $10$ static scenes with a driving distance of $\sim 50$ meters each for evaluating NVS at a $90\%$ drop rate. On average, we select one frame every 4 meters driving distance in this setting. We release $50\%$ of the frames for training and retain $50\%$ for evaluation.

\subsubsection{Baselines}
We evaluate two sets of baselines for this task. The first baseline (PCL) takes a colored point cloud as input. We project non-occluded points to the test viewpoint and interpolate the missing values to obtain the full image. 
To determine non-occluded points, we reconstruct a mesh using the ball-pivoting method~\cite{Bernardini1999VCG} on the accumulated point cloud. As there is no point in the sky region, we  in-paint the sky using a constant blue color. The sky region is heuristically determined based on the projected 3D points, i.e., a large connected area in the upper half of the image without any 3D projections is considered as the sky.

The second set of baselines takes a set of images as input. For all NeRF-based methods~\cite{Mildenhall2020ECCV,Barron2021ICCV,Deng2021ARXIV}, we train one model on each scene individually, using cascaded sampling with $256$ coarse samples and $256$ fine samples. We adopt the PyTorch reimplementation of NeRF~\footnote{\url{https://github.com/yenchenlin/nerf-pytorch}}, the original implementation of mip-NeRF~\footnote{\url{https://github.com/google/mipnerf}} and DS-NeRF~\footnote{\url{https://github.com/dunbar12138/DSNeRF}}. 
As for Free View Synthesis (FVS)~\cite{Riegler2020ECCV}, we follow its original implementation\footnote{\url{https://github.com/isl-org/FreeViewSynthesis}} and use their released model trained on the Tanks and Temples dataset~\cite{Knapitsch2017SIGGRAPH} which generalizes well. We follow the original implementation\footnote{\url{https://gitlab.inria.fr/sibr/projects/pointbased_neural_rendering}} of PBNR~\cite{Kopanas2021CGF} that optimizes a set of attributes such as reprojected features or depth in each input view.

\subsubsection{Additional Results}
We show additional qualitative results of all methods in \figref{fig:benchmark_nvs_label_app} (left). The PCL baseline exhibits blocky artifacts due to interpolation. The vanilla NeRF shows promising performance but sometimes struggles  due to the sparse input views. While mip-NeRF and DS-NeRF both improve the performance, the thin structures (\eg, fence) are still not well recovered. Interestingly, FVS and PBNR are better at preserving the fine details (\eg,  license plate) but have lower PSNR. This could be due to small misalignments in the image space.

\subsection{Benchmark of Novel View Semantic Synthesis}
\subsubsection{Evaluation Metric}
We evaluate the confidence weighted mIoU using \eqref{eq:weighted_iou}. Similar to the 2D semantic segmentation task, we omit ``Train'' and ``Bus'' during evaluation. We additionally omit ``Truck'', ``Person''', ``Rider'', ``Bicycle'' and ``Traffic Light'' as these classes do not appear in the $5$ static scenes for evaluating NVS.
\subsubsection{Baselines}
As there is no existing research work on this new benchmark, we directly apply PSPNet used in the 2D semantic segmentation task to synthesized images for semantic label prediction.
\subsubsection{Additional Results}
We show confidence weighted IoU on individual classes in \tabref{tab:benchmark_nvs_semantic}. Note that this na\"ive baseline leads to significantly degraded performance on most of the classes. As shown in \figref{fig:benchmark_nvs_label_app} (right), small changes in the image space sometimes lead to sigficant changes in the semantic prediction.

\begin{table*}[t!]
\setlength{\tabcolsep}{0.2pt}
\begin{center}
\begin{tabular}{|p{2.00cm}|>{\centering}p{0.80cm}>{\centering}p{0.80cm}>{\centering}p{0.80cm}>{\centering}p{0.80cm}>{\centering}p{0.80cm}>{\centering}p{0.80cm}>{\centering}p{0.80cm}>{\centering}p{0.80cm}>{\centering}p{0.80cm}>{\centering}p{0.80cm}>{\centering}p{0.80cm}>{\centering}p{0.80cm}|>{\centering}p{1.60cm}|}  \hline
Method & Road & Sdwlk & Bldg & Wall & Fence & Pole & Trsgn & Vegt & Terr & Sky & Car & Motor & mIoU$_\text{class}$ \tabularnewline
\hline
Original & 95.6 &85.6 &75.2 &57.4 &55.0 &42.7 &19.8 &93.2 &91.8 &93.4 &95.3 &58.5 & 72.0  \tabularnewline
PCL & 92.0 &70.4 &39.4 &29.7 &15.3 &2.2 &1.5 &67.2 &81.7 &42.1 &31.3 &0.0  &39.4  \tabularnewline
NeRF~\cite{Mildenhall2020ECCV} & 93.2 &74.2 &58.4 &35.1 &15.2 &16.6 &13.8 &83.8 &64.9 &91.6 &88.1 &1.2 &53.0  \tabularnewline
mip-NeRF~\cite{Barron2021ICCV} & 93.0 &72.9 &55.9 &30.4 &14.8 &12.6 &9.5  &82.0 &63.2 &91.5 &87.9 &0.0 &51.2  \tabularnewline
DS-NeRF~\cite{Deng2021ARXIV} & 93.2 &75.4 &56.5 &35.4 &19.1 &26.5 &16.6 &83.5 &63.9 &91.5 &89.2 &6.5 &54.8  \tabularnewline
FVS~\cite{Riegler2020ECCV} & 95.5 &83.2 &65.0 &54.2 &44.0 &17.8 &33.8 &90.5 &87.3 &91.0 &92.6 &50.2 &67.1  \tabularnewline
PBNR~\cite{Kopanas2021CGF} & 95.3 &82.4 &67.1 &46.6 &44.4 &28.8 &18.9 &90.0 &87.7 &89.1 &88.5 &42.2 &65.1  \tabularnewline
\hline
\end{tabular}

\end{center}
\caption{{\bf Additional Quantitative Results for Novel View Semantic Synthesis.}}
\label{tab:benchmark_nvs_semantic}
\end{table*}

\section{Semantic SLAM Benchmark}
\label{app:benchmark_slam}

\subsection{Localization}
\label{app:benchmark_slam_localization}
\subsubsection{Evaluation Metric}
We adopt the standard Absolute Pose Error (APE) and Relative Pose Error (RPE)\cite{Grupp2017EVO} as metrics for evaluating pose estimation. We align the predicted trajectory to the ground truth using a rigid transformation to evaluate the APE~\cite{Umeyama1991PAMI}. The RPE is evaluated between two frames with a distance of 1 meter. 
\subsubsection{Baselines}
We evaluate ORB-SLAM2~\cite{Artal2017TR}\footnote{\url{https://github.com/raulmur/ORB_SLAM2}} and SUMA++~\cite{Chen2019IROS}\footnote{\url{https://github.com/PRBonn/semantic_suma}} using their official implementations as baselines.
\subsubsection{Additional Results}
\begin{figure*}[t!]
\centering
\begin{subfigure}{.39\linewidth}
	\includegraphics[width=\linewidth]{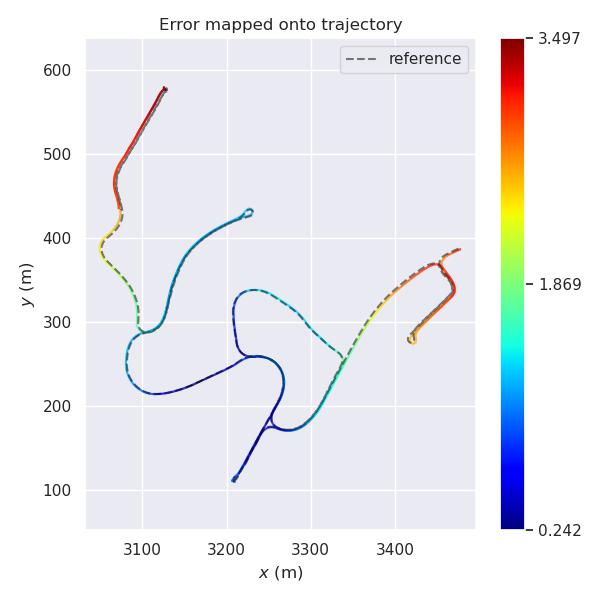} 
	\caption{ORB-SLAM2}
	\label{fig:benchmark_slam_ape_orb}
\end{subfigure}
\quad\quad
\begin{subfigure}{.39\linewidth}
	\includegraphics[width=\linewidth]{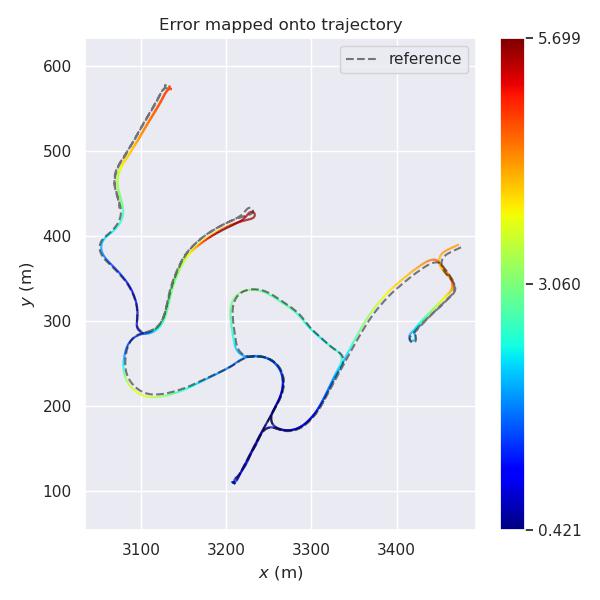} 
	\caption{SUMA++}
	\label{fig:benchmark_slam_ape_suma}
\end{subfigure}
\caption{{\bf Qualitative Results on Localization.} Each figure compares the predicted trajectory to the ground truth. Color indicates the APE in meters.}
\label{fig:benchmark_slam_localization}
\end{figure*}

\figref{fig:benchmark_slam_localization} shows qualitative comparison of predicted trajectories. As can be seen, both methods achieve reasonable performance while SUMA++ has a larger maximum error than ORB-SLAM2.   

\subsection{Geometric \& Semantic Mapping}
\label{app:benchmark_slam_mapping}
\subsubsection{Evaluation Metric}
\label{app:benchmark_slam_metric}
We adopt the same evaluation metrics considered in the semantic scene completion benchmark, as introduced in \appref{app:benchmark_scene_complt}.
When evaluating the quality of reconstruction, we compare ground truth and estimated reconstruction in local windows to minimize the impact of pose drifts. Specifically, we divide the test sequences into a set of local windows, each consisting of $50$ consecutive frames. We first crop the ground truth and the reconstructed point cloud \wrt the region of interest of each window. These two local point clouds are then aligned using the similarity transformation between the corresponding poses~\cite{Umeyama1991PAMI} and compared afterwards. 
Finally, we average the completeness, accuracy, and mIoU metrics over the entire test sequence. Following~\cite{Schops2017CVPR}, we measure completeness and accuracy over discretized voxels such that these metrics are insensitive to the density of the point clouds. 
\subsubsection{Baselines}
To obtain dense semantic reconstruction given the localization results of ORB-SLAM2, we unproject 2D semantic segmentation maps obtained from PSPNet~\cite{Zhao2017CVPR} using depth maps estimated by semi-global matching (SGM)~\cite{Hirschmueller2008PAMI}. We merge the unprojected 3D points using the poses predicted by ORB-SLAM2. As for SUMA++, we use the semantic estimation model pre-trained on KITTI.
\subsubsection{Additional Results}
\begin{figure*}[t!]
\centering
\begin{tabular}{P{0.5em}P{20em}P{20em}}
	\rotatebox{90}{Completeness} & 
	\includegraphics[width=\linewidth]{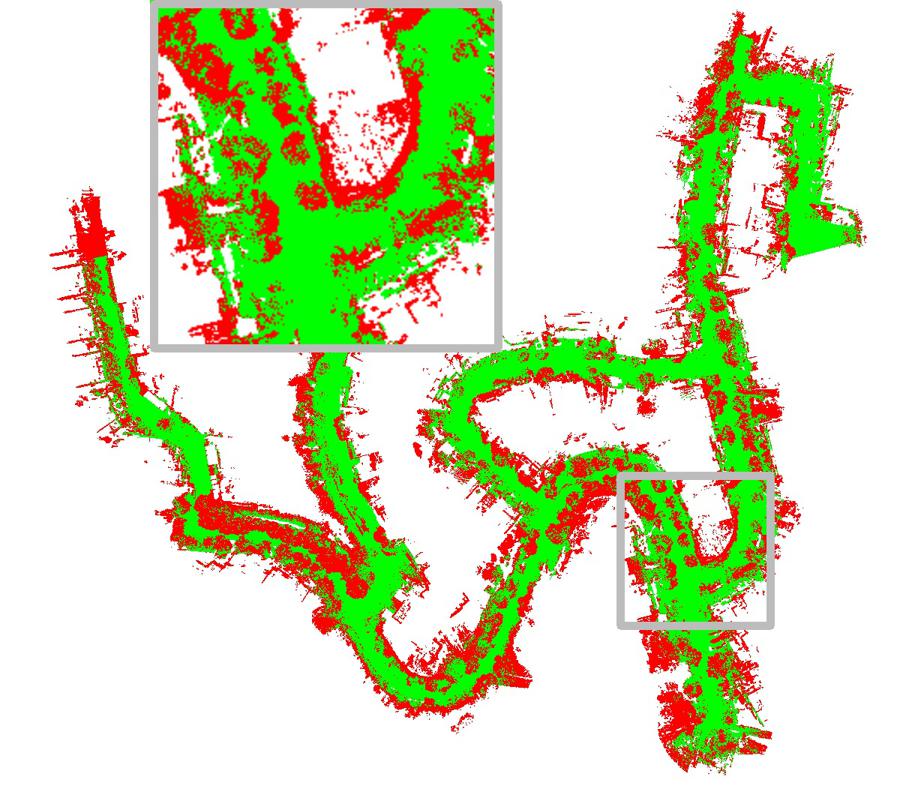} &
	\includegraphics[width=\linewidth]{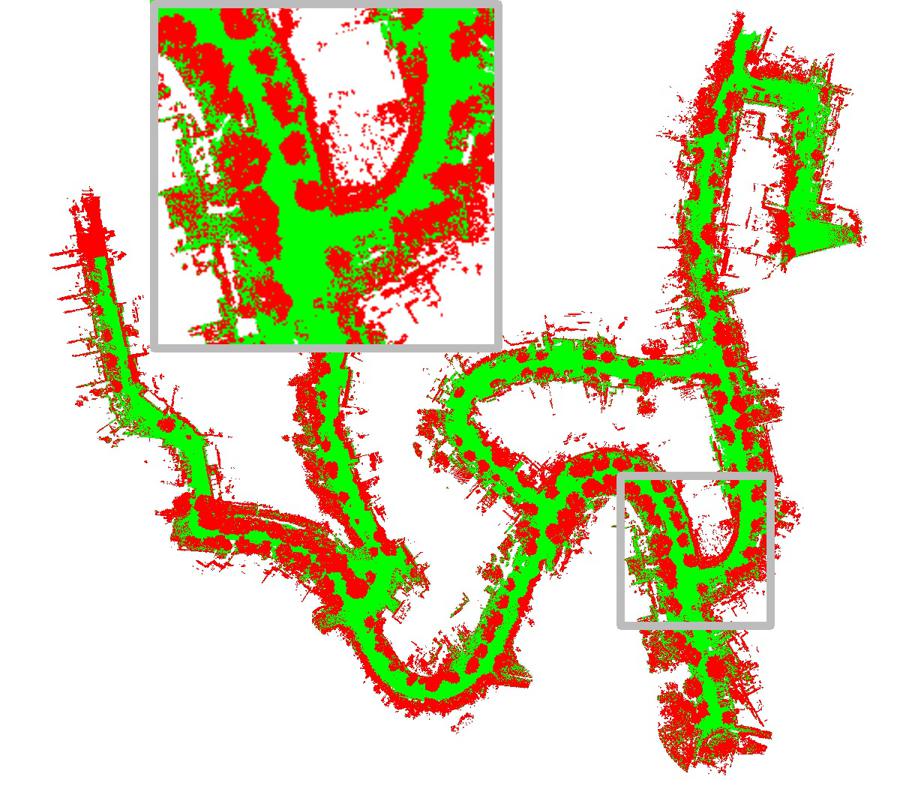} \\
	\rotatebox{90}{Accuracy} & 
	\includegraphics[width=\linewidth]{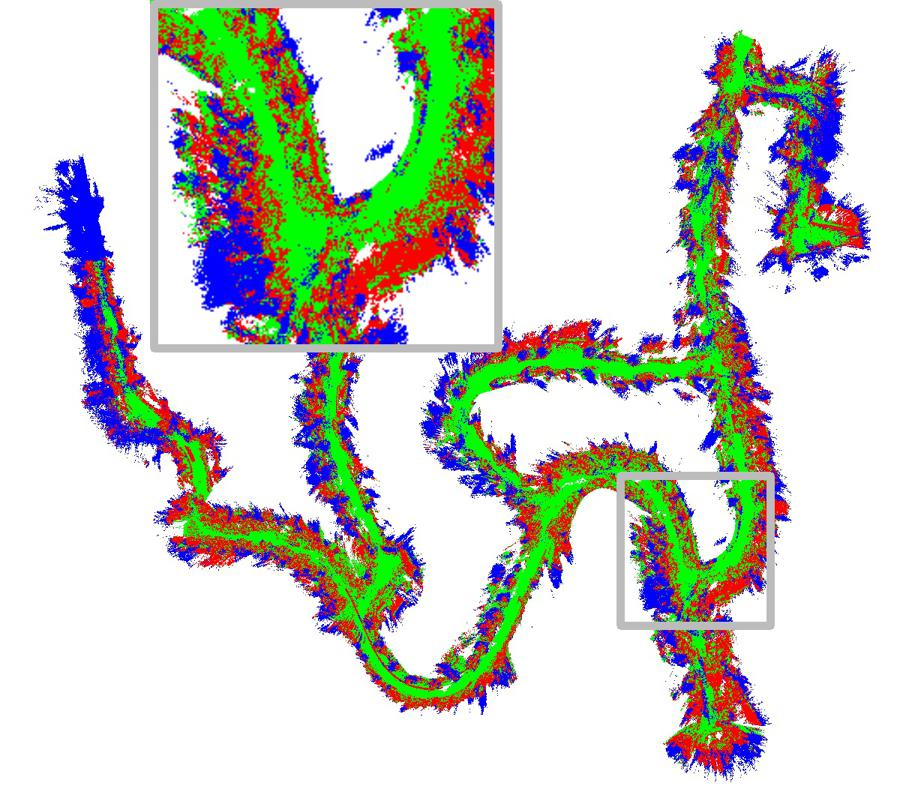} &
	\includegraphics[width=\linewidth]{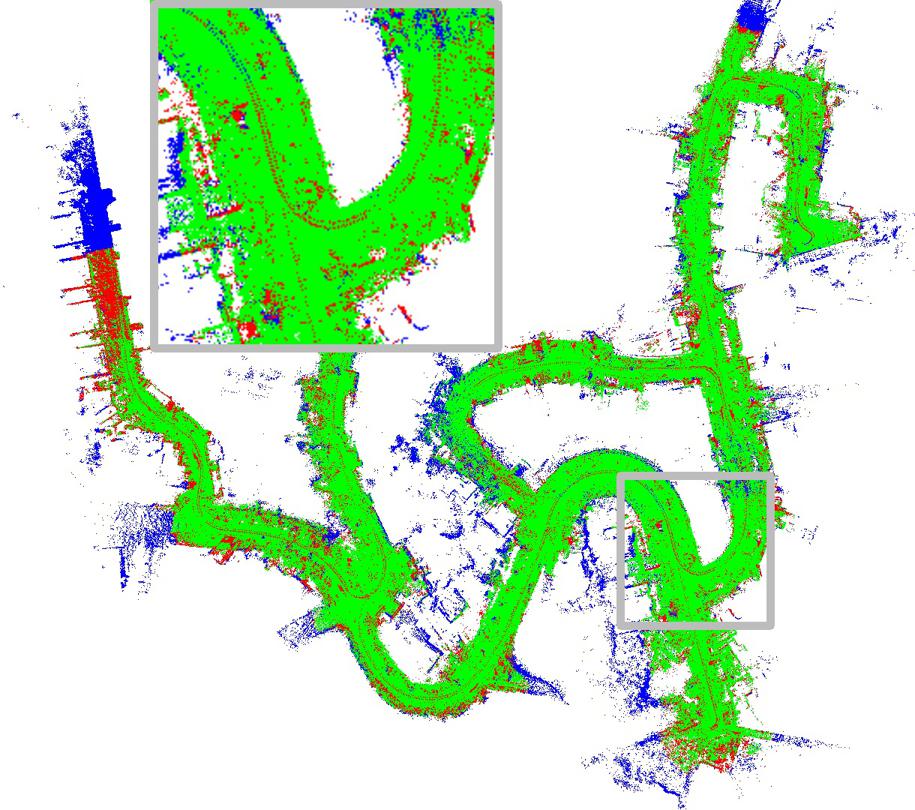} \\
	& (a) ORB-SLAM2 & (b) SUMA ++  
\end{tabular}
\caption{{\bf Qualitative Results on Localization and Mapping.} Completeness and accuracy evaluated at a threshold of 10cm. The first row shows the ground truth point cloud with green denoting complete and red for incomplete. The second row shows the predicted point cloud with green denoting accurate, red for inaccurate, and blue for points in unobserved regions. Note that SUMA++ is more accurate while ORB-SLAM2+SGM is more complete.}
\label{fig:benchmark_slam_reconstruction}
\end{figure*}

We show additional qualitative results of the geometric mapping in \figref{fig:benchmark_slam_reconstruction} in terms of completeness and accuracy at the threshold of 10cm. Consistent with the quantitative results, SUMA++ is more accurate while ORB-SLAM2+SGM is more complete.

\end{appendices}

\end{document}